%% file: paper_journal.tex
\newif\ifconfver
 \confvertrue        

\newif\ifplainver  
 \plainverfalse

\ifplainver
    \confverfalse   
\fi

\ifconfver
     \documentclass[10pt,twocolumn,twoside]{IEEEtran}
\else
    \ifplainver
        \documentclass[11pt]{article}
        \usepackage{fullpage}
    \else
        \documentclass[11pt,draftcls,onecolumn]{IEEEtran}
    \fi
\fi
\usepackage{calc,amsfonts,amssymb,amsmath ,bm,url,theorem,graphicx,cite,shortcuts_OPT,enumitem,booktabs}

\usepackage{colortbl}

\usepackage{algorithm}
\usepackage{algorithmic}
\usepackage{psfrag,subcaption,float,hyperref,bbm,cleveref}
\hypersetup{
  colorlinks   = true, 
  urlcolor     = blue, 
  linkcolor    = blue, 
  citecolor    = blue   
}

\usepackage{tikz}
\usetikzlibrary{
  arrows.meta,
  chains,
  matrix,
  positioning,shapes.geometric}
 \usepackage{circuitikz}

\newtheorem{Lemma}{Lemma}
\newtheorem{Prob}{Problem}
\newtheorem{Prop}{Proposition}

\newtheorem{assumption}{Assumption}
\newtheorem{Remark}{Remark}

\theorembodyfont{\rmfamily}
\newtheorem{Exa}{Example}

\usepackage{pgfplots}
\usetikzlibrary{arrows,shapes,calc,tikzmark,backgrounds,matrix,decorations.markings,backgrounds}
\usepgfplotslibrary{fillbetween}
 
\pgfplotsset{compat=1.3}

\usepackage{relsize}
\tikzset{fontscale/.style = {font=\relsize{#1}}
    }

\definecolor{lavander}{cmyk}{0,0.48,0,0}
\definecolor{violet}{cmyk}{0.79,0.88,0,0}
\definecolor{burntorange}{cmyk}{0,0.52,1,0}

\definecolor{asuorange}{rgb}{1,0.699,0.0625}
\definecolor{asured}{rgb}{0.598,0,0.199}
\definecolor{asuborder}{rgb}{0.953,0.484,0}
\definecolor{asugrey}{rgb}{0.309,0.332,0.340}
\definecolor{asublue}{rgb}{0,0.555,0.836}
\definecolor{asugold}{rgb}{1,0.777,0.008}

\tikzstyle{latent}=[draw,circle, black!70, fill = black!30,
                       text=violet,minimum width=10pt]
\tikzstyle{superpeers}=[draw,circle, asublue!80!white, fill = asublue!50!white,
                       text=violet,minimum width=10pt]
\tikzstyle{susceptible}=[draw,circle, red, fill = red!50,
                       text=violet,minimum width=10pt]

\definecolor{color0}{rgb}{0,0.75,0.75}
\newenvironment{customlegend}[1][]{%
  \begingroup
  \csname pgfplots@init@cleared@structures\endcsname
  \pgfplotsset{#1}%
}{%
  \csname pgfplots@createlegend\endcsname
  \endgroup
}%
\def\addlegendimage{\csname pgfplots@addlegendimage\endcsname}

    \makeatletter
    \def\multilimits@{\bgroup
  \Let@
  \restore@math@cr
  \default@tag
 \baselineskip\fontdimen10 \scriptfont\tw@
 \advance\baselineskip\fontdimen12 \scriptfont\tw@
 \lineskip\thr@@\fontdimen8 \scriptfont\thr@@
 \lineskiplimit\lineskip
 \vbox\bgroup\ialign\bgroup\hfil$\m@th\scriptstyle{##}$\hfil\crcr}
    \def\Sb{_\multilimits@}
    \def\endSb{\crcr\egroup\egroup\egroup}
\makeatother

\usepackage{pgfplots}   
\usepgfplotslibrary{units}
\usetikzlibrary{spy,backgrounds}
\usepgfplotslibrary{groupplots}

\makeatletter
\DeclareRobustCommand*\cal{\@fontswitch\relax\mathcal}
\makeatother

\begin{document}
\title{Online Inference for Mixture Model of Streaming Graph Signals with Non-White Excitation}
\author{Yiran He, Hoi-To Wai\thanks{A preliminary version of this work has been presented at ICASSP 2022, Marina Bay Sands, Singapore \cite{he2022joint}. 
Y.~He and  H.-T.~Wai are with the Department of SEEM, The Chinese University of Hong Kong, Shatin, Hong Kong SAR of China. E-mails: \url{yrhe@se.cuhk.edu.hk}, \url{htwai@se.cuhk.edu.hk}. This work is supported in part by RGC Project \#24203520.}} 

\maketitle

\begin{abstract}
This paper considers a joint multi-graph inference and clustering problem for simultaneous inference of node centrality and association of graph signals with their graphs. We study a mixture model of filtered low pass graph signals with possibly non-white and low-rank excitation. While the mixture model is motivated from practical scenarios, it presents significant challenges to prior graph learning methods. As a remedy, we consider an inference problem focusing on the node centrality of graphs. We design an expectation-maximization (EM) algorithm with a unique low-rank plus sparse prior derived from low pass signal property. We propose a novel online EM algorithm for inference from streaming data. As an example, we extend the online algorithm to detect if the signals are generated from an abnormal graph. We show that the proposed algorithms converge to a stationary point of the maximum-a-posterior (MAP) problem. Numerical experiments support our analysis.\end{abstract}

\begin{IEEEkeywords} 
blind centrality inference, clustering of graph signals, expectation maximization, online graph learning
\end{IEEEkeywords} 
\vspace{-.1cm}

\section{Introduction}
The increasing demands for extracting information from complex systems have motivated the study of graphical models in many disciplines such as social science, biology, and data science.
To analyze graph signals, i.e., observations made on the \emph{nodes}, graph signal processing (GSP) \cite{Sand2013DSP,Shum2013GSP} has emerged as a natural framework for signal processing applications such as
denoising \cite{sandryhaila2014big}, sampling \cite{tanaka2020sampling}, etc. Importantly, studies on graph topology learning using graph signal observations have been reported. Popular methods are proposed based on smoothness \cite{dong2016learning, kalofolias2016learn}, spectral template \cite{segarra2017network}, topological constraints \cite{egilmez2017graph}, causal modeling \cite{mei2016signal}, nonlinear model \cite{shen2019nonlinear, wai2019joint}, partial observations \cite{coutino2020state}; see the overview papers \cite{dong2019learning, mateos2019connecting}. Moreover, a recent direction is to perform end-to-end learning for features of graph topology. The subjects of interest include centrality \cite{roddenberry2020blind,he2022detecting}, communities \cite{wai2019blind,Schaub2020BCD}, network processes \cite{zhu2020estimating, he2021identifying}, etc. Compared to traditional graph learning, the latter approaches are robust to challenging but realistic scenarios such as when the excitation is not white noise, or when the graph signals are not sufficiently smooth. 

Many existing results on graph learning focus on a setting where the goal is to infer a single graph from data. In reality, the data can be more complex and is related to multiple graphs. For example, recent works \cite{hallac2017network,baingana2016tracking,yamada2020time, navarro2020joint, yang2020network,  rey2022joint} studied the time varying graph learning problem when the topology changes slowly. Alternatively, one also considers the scenario where the graph topology differ significantly across samples. For example, a series of resting state brain networks have been identified from brain signals \cite{ricchi2022dynamics}; stock prices recorded at different states of the market may lead to different graph topology \cite{heiberger2014stock}. This model is also relevant to the problem of detecting topology changes in graph signals \cite{kaushik2021network,chepuri2016subgraph,shaked2021identification, isufi2018blind}. 

This paper treats a \emph{joint multi-graph centrality inference and clustering} problem which simultaneously infers the node centrality of multiple graphs and clusters observed signals with respect to the graph that generates them. Our problem is motivated by applications involving multiple graphs with unknown associations between the graphs and observations. 
For example, when observing brain signals, we do not know which state the subject is in; for stock prices observations, the states of the market can be difficult to identify. 
While centrality inference can be performed by prior works \cite{he2022detecting,roddenberry2020blind}, the clustering problem is more challenging as classical algorithms such as spectral clustering \cite{von2007tutorial}, KNN \cite{guo2003knn} do not consider structures in the graph signal observations which is crucial to providing a reliable estimates. 
Recent works have developed algorithms that focus on simultaneous clustering and graph topology learning, e.g., graph Laplacian mixture model \cite{maretic2020graph} and its regularized version \cite{yuan2022gracge}, regularized spectral clustering \cite{karaaslanli2022simultaneous}, $K$-means based method \cite{araghi2019k}. Most of these works are developed from the Gaussian Markov random field model and entail stringent conditions such as requiring the observations to be generated from full-rank, white excitation. 
In comparison, our approach handles a relaxed mixture model of graph signal with possibly low-rank, non-white excitation.  

The current paper also proposes an \emph{online algorithm} for the joint inference problem from {streaming data}. 
Notice that many existing graph learning algorithms require batch data. This is in contrast to the practical environment that involves streaming and even dynamical data collection. 
Furthermore, the online algorithm enjoys a low memory footprint and computation complexity by processing data on-the-fly.
Several online algorithms on graph topology learning have been proposed, e.g., for time varying graph learning \cite{vlaski2018online,shafipour2020online,shumovskaia2021online}, for multi-graph topology learning but with pre-clustered data \cite{saboksayr2021discriminative}.
In contrast, our algorithm is the first to perform multi-graph inference and clustering simultaneously and in an online fashion.
Our key contributions are:
\begin{itemize}[leftmargin=*, itemsep=0.1cm, topsep=0.1cm]
 \item To study graph signals observations based on multiple graphs, we propose a mixture model of graph signals with general non-white excitation. 
 Moreover, the model supports missing data and general observation model such as the logit model for inference from binary data.
 
\item We formulate a joint inference and clustering problem via the MAP framework to infer node centrality and cluster observations according to the graphs. We design a batch EM algorithm under a unique low-rank plus sparse prior. We show that the EM algorithm converges to a stationary point at a sublinear rate. The algorithm supports efficient implementation for inference in the mixture model.

\item We develop a novel online EM algorithm based on the stochastic approximation (SA) scheme for streaming data. The algorithm processes each of the incoming observation on-the-fly and features a low memory footprint while delivering similar performances as the batch EM. Our analysis shows that any fixed point of the algorithm is a stationary point of MAP. We also describe an application of the online algorithm to blind anomaly detection.

\item We perform numerical experiments on synthetic and real data from brain and stock markets. The efficacy of the proposed algorithms support our findings.

\end{itemize}
Compared to the conference version \cite{he2022joint}, this paper considers an extended signal model with missing data and logit observations. We also propose an online algorithm for streaming data and provide an extended set of experiments.

\noindent \textbf{Organization.} This paper is organized as follows. In Sec.~\ref{sec:problem}, we describe the mixture model of graph signals and then formally introduce the joint inference problem. Furthermore, we develop a maximum-a-posterior formulation with low-rank plus sparse prior that adapts to the low pass signal property.
In Sec.~\ref{sec:batchEM}, we propose a batch EM algorithm for Gaussian and logit observations. In Sec.~\ref{sec:onlineEM}, we introduce an online EM algorithm for streaming data and discuss its application to online anomaly detection. Finally, numerical experiments are presented to support our findings in Sec.~\ref{sec:exp}.

\noindent \textbf{Notations.} We use boldfaced character (resp.~boldfaced capital letter) to denote vector (resp.~matrix). For any vector ${\bm x} \in \Re^n$, $\| {\bm x} \|$, $\| {\bm x} \|_1$ denote the Euclidean, $\ell_1$ norm, respectively. For any matrix ${\bm X} \in \Re^{m \times n}$,  we take $[{\bm X}]_{i,j}$ to denote its $(i,j)$th entry. $\| {\bm X} \|$, $\| {\bm X} \|_1$, $\| {\bm X} \|_\star$ denote the spectral norm, 1-norm, nuclear norm, respectively. 
\vspace{-.3cm}

\section{Problem Statement} \label{sec:problem} \vspace{-.2cm}
Consider $\mulG$ undirected graphs $\mG{\mulg} = ({\cal V}, {\cal E}^{(\mulg)},\mA{\mulg})$, $c=1,\dots,C$. They share the same node set ${\cal V} :=\{1,\dots, n\}$ but with different edge sets $\{{\cal E}^{(\mulg)}\}_{\mulg=1}^\mulG\subseteq{\cal V}\times{\cal V}$. Each graph $\mG{\mulg}$ is endowed with a symmetric weighted adjacency matrix $\mA{\mulg} \in \Re^{n\times n}_+$ where $A_{ij}^{(\mulg)} > 0$ if and only if $(i,j)\in{\cal E}^{(\mulg)}$; otherwise, $A_{ij}^{(\mulg)}=0$. Define the eigenvalue decomposition (EVD) $\mA{\mulg}={\bm V}^{(\mulg)}{\bm \Lambda}^{(\mulg)}[{\bm V}^{(\mulg)}]^\top$ where ${\bm V}^{(\mulg)}$ is an orthogonal matrix and ${\bm \Lambda}^{(\mulg)} = \text{Diag}({\bm \lambda}^{(\mulg)})$ contains its  eigenvalues in descending order as: $\mlamb{\mulg}{1} \geq \dots \geq \mlamb{\mulg}{n}$. 
In this paper, we consider graphs that differ from each other in terms of their sets of central nodes.
We adopt the notion of {eigen-centrality} to measure the latter. For graph $\mG{\mulg}$, its centrality vector is given by the top eigenvector ${\bm v}_1^{(\mulg)}$. 
Node $i$ is said to be more \emph{central} if the magnitude of its centrality value is greater.

We observe the graph signals on ${\cal V}$ generated from a process defined on one of the graphs. These graph processes are described via the linear {graph filters} \cite{Sand2013DSP}: for $\mulg = 1, \ldots, \mulG$,
\beq\label{eq:gfilter} \textstyle
{\cal H}(\mA{\mulg}) = \sum_{\tau=0}^{P-1} h_\tau[ \mA{\mulg} ]^\tau \in \Re^{n\times n},
\eeq
where $\{h_\tau\}_{\tau=0}^{P-1}$ are the filter coefficients, $P \in {\cal Z}_+ \cup \{\infty\}$ is the filter's order. 
Each observation is indexed by $t \in {\cal Z}_+$ and is modeled as a graph signal matched with an identifier variable $w_t \in \{1,\dots,\mulG\}$. The latter indicates that the graph signal is generated from $\mG{w_t}$. We describe the observation via a \emph{mixture model of graph signals} with missing data:\vspace{-.05cm}
\begin{subequations}\label{eq:mixmodel}
\begin{align}
{\bm \op}_t & = {\bm \dmp}_t \odot \left\{ \sum_{\mulg=1}^\mulG \mathbbm{1} (w_t=\mulg) {\cal H}(\mA{\mulg}) {\bm x}_t + {\bm \wnoise}_t \right\}. \label{eq:y=Hx} \\
{\bm x}_t & \textstyle = {\bm B} {\bm z}_t = \sum_{j=1}^r {\bm b}_j [{\bm z}_t]_j. \label{eq:x=Bz}
\end{align}
\end{subequations}
In \eqref{eq:y=Hx}, ${\bm \ip}_t \in \Re^n$ is the excitation to the graph filter to be described later, ${\bm \wnoise}_t\sim{\cal N}({\bm 0},\sigma^2{\bm I})$ is a Gaussian observation noise, $\mathbbm{1}(\cdot)$ is the $\{0,1\}$ indicator function, and $\odot$ is the element-wise product. 
The vector ${\cal H}(\mA{\mulg}) {\bm x}_t$ is the output of the graph filter ${\cal H}(\mA{\mulg})$ with the excitation ${\bm x}_t$. 
The graph identifier $w_t$ is a multinomial random variable (r.v.) with probability mass function $\mathbbm{P}(w_t=\mulg)=P_\mulg$. The binary vector ${\bm \dmp}_t\in\{0,1\}^n$ models on which nodes the signal values are missing in the current sample. 
An extension to logit observations will be described in Sec.~\ref{sec:logit}.

\begin{figure}[t]
  \begin{center}\resizebox{.95\linewidth}{!}{\sf
  \begin{tikzpicture}[node distance=10pt]
    \node[fill = red!5] (start)   {{\color{red}${\bm \latp}_t$}};
    \node[draw,right=5pt of start] (B)  {${\bm B}$};
    \node [coordinate, right of=B,xshift=0.3cm](x){};
    \node [coordinate, above of=x,yshift=0.3cm](test1){};
    \node [coordinate, below of=x,yshift=-0.3cm](test2){};
    \node[draw, above of=x, yshift=0.3cm, left of=x, xshift=1.5cm] (filter1)
    {${\gfilter}(\mA{1})$};
    \node [coordinate, right of=filter1,xshift=0.3cm](test5){};
    \node[draw,below of=x, yshift=-0.3cm, left of=x, xshift=1.5cm]
    (filterC)  {${\gfilter}(\mA{\mulG})$};
      \node[ above of=filterC, yshift=0.4cm]
    (filters2)  {$\vdots$};
     \node [coordinate, right of=filterC,xshift=0.3cm](test6){};
    \node[draw, circle,right=100pt of x, fill = white] (plus)  {$+$};
     \node [coordinate, left of=plus,xshift=-0.3cm](test3){};
    \node [coordinate, above of=test3,yshift=0.3cm](test4){};
    \node [coordinate, below of=test3,yshift=-0.3cm](test7){};
    \node[draw,circle, right of=plus, xshift=0.6cm]
    (dproduct)  {$\odot$};
    \node[ above=10pt of dproduct, fill = red!5] (gamma) {{\color{red} ${\bm \dmp}_t$}};
    \node[ below=10pt of plus] (noise) {${\bm \wnoise}_t$};
    \node[ right=10pt of dproduct, fill = red!5] (end) {{\color{red}${\bm \op}_t$}};
    \node[ left of=plus, xshift=-1cm,yshift=0.1cm]
    (wl)  {${w}_t$};
    \node [coordinate, above of=wl,yshift=0.1cm](test8){};
    \node [coordinate, below of=wl,yshift=-0.1cm](test9){};
    \draw[->] (start)  -- (B);
    \draw[-] (B) -- node[below] {${\bm x}_t$} (x);
    \draw[-] (x) --  (test1);
    \draw[-] (x) --  (test2);
    \draw[->] (test1) --  (filter1);
    \draw[->] (test2) --  (filterC);
    \draw[-] (test5) to[cspst]  (test4);
    \draw[-] (test6) to[cspst]  (test7);
    \draw[-] (test4) --  (test3);
    \draw[-] (test7) --  (test3);
    \draw[->] (test3) -- (plus);
    \draw[->] (noise) -- (plus);
    \draw[->] (wl) -- (test8);
    \draw[->] (wl) -- (test9);
    \draw[->] (dproduct) -- (end);
    \draw[->] (gamma) -- (dproduct);
    \draw[->] (plus) --
    (dproduct);
  \end{tikzpicture}}
  \end{center}
  \caption{Generation process for the mixture model of graph signals \eqref{eq:mixmodel}. Black (resp.~{\color{red} red}) color denotes unknown (resp.~known) variables [cf.~Problem~\ref{prob:jg}]. 
  }\vspace{-.3cm} \label{fig:datamodel}
  \end{figure}
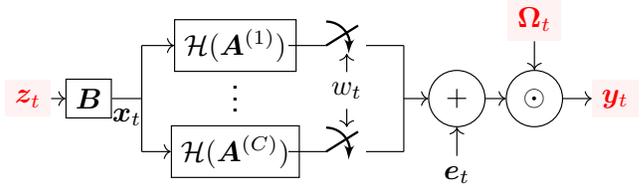

In \eqref{eq:x=Bz}, we further model that the excitation signal ${\bm \ip}_t$ lies in a general $r$-dimensional subspace ${\rm span} \{ {\bm B} \}$ with ${\bm B} \in \RR^{n \times r}$, $r \leq n$. The setting is in line with real world observations as data tends to be low-rank \cite{udell2019big}. The vector ${\bm \latp}_t \in \Re^r$ is an excitation parameter whose element represents an observable source of stimuli on the graph process. Each column ${\bm b}_j$ is the influence profile from the $j$th source ${\bm \latp}_{t,j}$ on the node set ${\cal V}$. 
For instance, ${\bm B}$ can be modeled as a sparse matrix in this regard.
Fig.~\ref{fig:datamodel} summarizes the generation process\footnote{We remark that it is easy to extend \eqref{eq:mixmodel} to the setting that every graph filter has different filter coefficients, every graph is associated with a different excitation subspace matrix ${\bm B}^{(\mulg)}$, etc.} of \eqref{eq:mixmodel}. 

The fact that ${\bm B}$ can be non-identity has made it challenging, if not impossible, to perform inference on \eqref{eq:mixmodel} such as reconstructing the graph topology $\mA{\mulg}$ from the filtered graph signals. Note that even in the single graph setting, prior methods \cite{dong2016learning, segarra2017network, egilmez2017graph, kalofolias2016learn} require the graph filter to be excited by white noise, i.e., with ${\bm B} = {\bm I}$. 
As a remedy inspired by \cite{wai2019blind,roddenberry2020blind,he2022detecting,Schaub2020BCD}, we aim to perform partial inference on \eqref{eq:mixmodel} via the \emph{joint multi-graph centrality inference and clustering} problem:\vspace{-.1cm}
\begin{Prob} \label{prob:jg}
Given the data tuples $\{ {\sf DP}_t \}_{t} := \{ {\bm \op}_t, {\bm \latp}_t, {\bm \dmp}_t \}_{t}$ from \eqref{eq:mixmodel}, 
estimate {\sf (A)} eigen-centrality vector ${\bm v}_1^{(\mulg)}$ for each graph, and {\sf (B)} identifier variable $w_t \in \{1,\dots,\mulG \}$ for each sample  (subject to permutation ambiguity).\vspace{-.2cm}
\end{Prob}
\noindent We consider two settings of data availability. In the first setting, the data tuples are available in a \emph{complete batch}, i.e., one observes $\{ {\sf DP}_t \}_{t=1}^m$ where $m$ denotes the total number of samples. In the second setting, the data tuples are revealed in a \emph{streaming fashion}. At time $t$, we only observe a sample ${\sf DP}_t$ that is generated randomly according to \eqref{eq:mixmodel}.
To avoid degeneracy, we assume that different graphs are equipped with different sets of central nodes such that ${\bm v}_{1}^{(\mulg)} \neq {\bm v}_{1}^{(\mulg')}, \mulg\neq\mulg'$. We concentrate on graphs with small groups of central nodes of high-intra and low-inter connectivity. Such graphs typically admit a core–periphery structure which can be characterized by the eigengap condition $\mlamb{\mulg}{1} \gg \mlamb{\mulg}{2}$ \cite{cucuringu2016detection}. 

Tackling the joint inference problem is challenging due to the large number of unknowns in the model \eqref{eq:mixmodel}. For instance, even with $\mulG = 1$, inferring the eigen-centrality vector ${\bm v}_1^{(1)}$ from \eqref{eq:mixmodel} is difficult since the graph filter ${\cal H}( \mA{1} )$, the excitation subspace ${\bm B}$, etc., are unknown.\vspace{-.1cm}

\begin{Remark}
The requirement for excitation parameters $\{ {\bm \latp}_t \}_t$ to be known may appear restrictive. However, we note in several applications, estimate of these parameters can be obtained as side information. For example, stock networks are excited by the market's interest level on various topics which can be estimated by the popularity of keywords on Google Trend. The excitation can also be endogenous such that $\{ {\bm \latp}_t \}_t$ is approximated by observations on a subset of nodes. See Sec.~\ref{sec:realexp} for two example applications using real data.\vspace{-.3cm}
\end{Remark}

\subsection{MAP Estimation with Reparametrization} \label{sec:MAP} \vspace{-.1cm}
This sub-section proposes a reparameterization technique to leverage the signal structure for finding a robust solution to Problem~\ref{prob:jg}. 
We then formulate the maximum-a-priori (MAP) problem which will be the focus for the rest of this paper. 

In the absence of knowledge on graph filters, Problem~\ref{prob:jg} will be ill-defined due to difficulty in extracting ${\bm v}_1^{(\mulg)}$.
Taking inspirations from \cite{he2022detecting, roddenberry2020blind}, we consider a low pass assumption \cite{Sand2013DSP, ramakrishna2020user} on the underlying graph filters:\vspace{-.1cm}
\begin{assumption} \label{assump:lowpass} 
The graph filter ${\cal H}(\mA{\mulg})$, is 1-low pass with:
\beq \textstyle
  \eta^{(\mulg)} := \max_{j=2,\dots,n}|h(\mlamb{\mulg}{j})|/|h(\mlamb{\mulg}{1})| < 1,
\eeq
for $\mulg=1,\dots,\mulG$, 
where the polynomial $h(\lambda) := \sum_{\tau=0}^{P-1}h_\tau \lambda^{\tau}$ is the frequency response of the graph filter ${\cal H}(\cdot)$. \vspace{-.1cm}
\end{assumption}
\noindent The low pass ratio $\eta^{(\mulg)}$ characterizes the strength of ${\cal H}(\mA{\mulg})$. With a smaller $\meta{\mulg}$, the filter ${\cal H}(\mA{\mulg})$ attenuates the signal components beyond the \emph{cutoff frequency} $\mlamb{\mulg}{1}$ more. If $\meta{\mulg}\approx 1$, then $\gfilter(\mA{\mulg})$ is considered as \emph{weak low pass}; if $\meta{\mulg}\ll 1$, then $\gfilter(\mA{\mulg})$ is considered as \emph{strong low pass}. 
\Cref{assump:lowpass} is common in modeling network processes. Examples include, but are not limited to, opinion dynamics in social networks, stock dynamics, power systems, etc., see \cite{ramakrishna2020user}. 

\Cref{assump:lowpass} implies that the top eigenvector of ${\cal H}(\mA{\mulg})$ corresponds to the centrality vector ${\bm v}_1^{(\mulg)}$. Together with the condition that ${\bm v}_{\sf 1}^{(\mulg)} \neq {\bm v}_{\sf 1}^{(\mulg')}$, one may tackle \Cref{prob:jg} through separating the $m( \gg \mulG)$ observations into $\mulG$ clusters using na\"{i}ve spectral clustering. 
Particularly, assume that ${\bm \dmp}_t = {\bf 1}$, the $(t, t')$th element of the correlation matrix of observations is
\beq 
\pscal{ {\bm y}_t }{ {\bm y}_{t'} } \approx \pscal{ {\cal H}( \mA{ w_t } ) {\bm B} {\bm z}_{t} }{ {\cal H}( \mA{ w_{t'} } ) {\bm B} {\bm z}_{t'} }
\eeq 
Since the top eigenvectors of ${\cal H}( \mA{ w_t } )$, ${\cal H}( \mA{ w_{t'} } )$ differ only if $w_t \neq w_{t'}$, the $m \times m$ correlation matrix shall exhibit a block structure aligned with the graph identifiers $\{ w_t \}_{t=1}^m$.

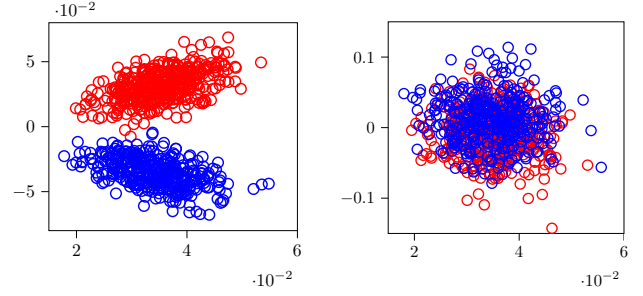
\begin{figure}[t]
\centering
\resizebox{.48\linewidth}{!}{\input{Mixture_slp}}
\resizebox{.475\linewidth}{!}{\input{Mixture_wlp}}
\caption{Toy Example illustrating the data from two core periphery graphs. (Left) Strong low pass filter ${\cal H}_{\sf s}({\bm A}) = ( {\bm I} - \frac{1}{40} {\bm A})^{-1}$; (Right) Weak low pass filter ${\cal H}_{\sf w}({\bm A}) = ( {\bm I} - \frac{1}{80} {\bm A})^{-1}$.}\label{fig:reliaz}\vspace{-.4cm}
\end{figure}

To this end, an intuitive idea is to apply spectral clustering (SC) on the correlation matrix $\widehat{\bm C}_Y = [ {\bm y}_1, \ldots, {\bm y}_m ]^\top [ {\bm y}_1, \ldots, {\bm y}_m ]$ to cluster the graph signals. However, as demonstrated below, the result is sensitive to the low pass filter modeling the graph process:\vspace{-.15cm}
\begin{Exa} \label{ex:bad}
We generate $\mulG=2$ core-periphery graphs with $n=100$ nodes, each with 10 distinct central nodes that are fully connected, and $m=400C$ graph signals are generated according to \eqref{eq:mixmodel}. Fig.~\ref{fig:reliaz} shows the scatter plot whose coordinates of the $m = 400\mulG$ points are taken to be the top/second eigen-vectors of $\widehat{\bm C}_Y$ and colored according to the true graph identifiers $\{ w_t \}_{ t=1 }^m$. For the stronger low pass filter [Fig.~\ref{fig:reliaz} (left)], the data points are found to be clearly clustered. 
For the weaker low pass filter [Fig.~\ref{fig:reliaz} (right)], the data points cannot be clustered. 
In the latter case, applying na\"{i}ve SC would result in erroneously clustered observations.\vspace{-.2cm}
\end{Exa}

The na\"{i}ve spectral clustering essentially utilizes difference in the subspaces ${\rm span} \{ {\cal H}( \mA{ \mulg } ) {\bm B} \}$, $\mulg=1,\ldots,\mulG$ to discern samples from different graphs. While such strategy is successful when ${\cal H}( \mA{ \mulg } )$ is strong low pass, it may not work when the filter is weak low pass; see Remark~\ref{rem:weak} for further justifications. Nevertheless, the above example shows that inferring the natural parameters $\{ {\cal H}( \mA{ \mulg } ) {\bm B} \}_{\mulg=1}^{\mulG}$ can be insufficient for a robust solution to \Cref{prob:jg}.

Particularly, the above example demonstrates that it is necessary to jointly consider the signal structure while clustering the graph signals. Our idea is to model and extract the hidden component(s) in ${\gfilter} (\mA{\mulg}) {\bm \sm}$ that are indicative of the eigen-centrality vector, which thus provides the graph identifiers necessary for clustering. Observe the decomposition:
\beq \label{eq:HB_dec}
{\gfilter}(\mA{\mulg}){\bm \sm}  = ( {\gfilter}(\mA{\mulg}) - \rho {\bm I} ) {\bm \sm} + \rho {\bm \sm} \equiv {\bm L}_{\mulg} + \emsm,
\eeq
for any $\rho \geq 0$.
The component ${\bm L}_{\mulg} = ( {\cal H}(\mA{\mulg}) - \rho{\bm I} ) {\bm B}$ depends on the shifted graph filter ${\cal H}(\mA{\mulg}) - \rho{\bm I}$. It is shown \cite[Observation 1]{wai2019blind} that there exists $\rho > 0$ where the shifted graph filter enjoys a strictly lower low pass ratio, denoted as $\mteta{\mulg}$, than the original ratio $\eta^{(\mulg)}$. 
For example, with ${\cal H}(\mA{\mulg}) = ({\bm I} - \alpha \mA{\mulg} )^{-1}$, it can be shown that the shifted graph filter with $\rho = 1$ has the low pass ratio of $\mteta{\mulg} \leq \frac{\lambda_2^{(\mulg)}}{\lambda_1^{(\mulg)}} \eta^{(\mulg)} \ll \meta{\mulg}$, 
provided that $\mlamb{\mulg}{1} \gg \mlamb{\mulg}{2}$ which can be satisfied for graphs with core-periphery structure \cite{cucuringu2016detection}. Consequently, the matrix ${\bm L}_\mulg$ will be approximately rank-one. 

Below, we show that the low-rank components ${\bm L}_\mulg$ are distinct for the different graphs that they are associated with. 
Assume without loss of generality (w.l.o.g.) that $(\widetilde{\bm v}^{(\mulg)}_1)^\top {\bm v}^{(\mulg)}_1 \geq 0$, the following lemma is adapted from \cite[Corollary 1]{he2022detecting}:\vspace{-.2cm}
\begin{Lemma} \label{cor:cen}
For each $\mulg = 1, \ldots, \mulG$, if $({\bm v}^{(\mulg)}_1)^\top {\bm B} {\bm q}^{(\mulg)}_1 \neq 0$, then\vspace{-.1cm}
\beq \label{eq:lcbd}
\| \widetilde{\bm v}^{(\mulg)}_1 - {\bm v}_1^{(\mulg)} \| \leq \sqrt{2} \widetilde{\eta}^{(\mulg)} \,
\frac{|| ({\bm V}^{(\mulg)}_{n-1})^\top {\bm B} {\bm q}^{(\mulg)}_1 ||}{| ({\bm v}^{(\mulg)}_1)^\top {\bm B} {\bm q}^{(\mulg)}_1 |},\vspace{-.1cm}
\eeq
where ${\bm V}^{(\mulg)}_{n-1}$ is the last $n-1$ eigenvectors of $\mA{\mulg}$, and $\widetilde{\bm v}^{(\mulg)}_1, {\bm q}^{(\mulg)}_1$ are the top left, right singular vector of ${\bm L}_\mulg$.\vspace{-.15cm}
\end{Lemma}
\noindent The right hand side of \eqref{eq:lcbd} is bounded by ${\cal O} (\mteta{\mulg})$ with $\mteta{\mulg} \ll 1$. Together with the observation that ${\bm L}_{\mulg}$ is approximately rank one, we obtain ${\bm L}_{\mulg} \propto {\bm v}_1^{(\mulg)} ({\bm q}_1^{(\mulg)})^\top$. 
With the condition ${\bm v}_1^{(\mulg)} \neq {\bm v}_1^{(\mulg')}$, we observe that ${\bm L}_{\mulg}$ provides an effective indicator to distinguish the samples with different graph identifiers.

Establishing that ${\bm L}_{\mulg}$ is low rank may not be sufficient for its recovery in \eqref{eq:HB_dec}, where extra structure has to be leveraged for the residual term $\rho {\bm B}$ as inspired by \cite{agarwal2012noisy}. Fortunately, since ${\bm B}$ models the influences from external sources on the graph(s), we note from the applications described in \cite{he2022detecting, wai2019blind} that ${\bm B}$ admits certain low-dimensional structure. For example, ${\bm B}$ can be sparse, the number of non-zero row/column vectors of ${\bm B}$ can be small, etc. 
As such, herein we model ${\bm B}$ to be a sparse matrix which includes the special case of ${\bm B} = {\bm I}$.
We observe that the matrix product ${\gfilter}(\mA{\mulg}){\bm \sm}$ admits a `low-rank plus sparse' structure under the said premises. 
\vspace{.1cm}

\noindent \textbf{MAP Estimation.}
The above motivates us to explicitly account for the implicit components ${\bm L}_{\mulg}, \emsm$ during the inference process through a careful re-parameterization. Denote $\Theta := \big\{ \{ {\bm L}_\mulg \}_{\mulg=1}^\mulG, \emsm, \{ P_\mulg \}_{\mulg=1}^\mulG \big\}$ as the set of parameters. We yield the following \emph{structured} MAP estimation problem:
\beq\label{eq:MAP} 
\begin{array}{rl} 
\ds \max_{\Theta \in \mathfrak{T}} & \ds {\cal L}( \Theta ) := 
\EE \left[ \log p({\bm Y} | \Theta, {\bm Z} ,{\bm \dmp} ) \right]
+ \log p(\Theta) ,
\end{array}
\eeq
where $\mathfrak{T} = \{ \Theta : P_\mulg \geq 0,~\sum_{\mulg=1}^\mulG P_c = 1 \}$ and $p(\Theta)$ models the prior on $\Theta$ with the `low-rank plus sparse' structure of ${\bm L}_\mulg, \emsm$. A natural choice for the prior distribution is 
\beq \label{eq:prior} \textstyle
p( \Theta ) \propto \exp \big( - \lambda_S \| \emsm \|_1 - \lambda_L \sum_{\mulg=1}^\mulG \| {\bm L}_{\mulg} \|_\star \big), 
\eeq
where 
$\lambda_S, \lambda_L \geq 0$ are regularization parameters.
Furthermore, the expectation $\EE [ \cdot ]$ is defined w.r.t.~the observation law for ${\bm Y}, {\bm Z}, {\bm \dmp}$ and
the log-likelihood function is given by:
\begin{align}
& \log p({\bm \op} |\Theta,{\bm \latp},{\bm \dmp} ) = \notag \\
& \textstyle \log \left( \sum_{\mulg=1}^\mulG P_\mulg e^{ -\frac{1}{2\sigma^2}{||{\bm \op} -{\bm \dmp} \odot({\bm L}_\mulg+\emsm){\bm\latp} ||^2} } \right) + {\rm constant}, \label{eq:logli} 
\end{align} 
which is non-concave due to the nonlinear coupling between $P_c$, ${\bm L}_c$, $\emsm$, making
direct optimization of \eqref{eq:MAP} intractable. 

We notice that EM algorithms for classical models such as Gaussian Mixture Model (GMM) cannot be directly applied to \eqref{eq:MAP}. The reason is that \eqref{eq:MAP} entails regularization terms for the `low-rank plus sparse' structure. Additionally, the missing data and excitation parameter ${\bm z}_t$ have to be incorporated into the inference process. 
In the next section, we concentrate on developing effective algorithms for \eqref{eq:MAP} via the EM paradigm.\vspace{-.1cm}

\begin{Remark}\label{rem:weak}
The poor performance of na\"{i}ve SC with weak low pass filter can be explained by \cite[Lemma~2]{he2022detecting}. Under mild conditions, the latter lemma shows:
\beq
{ \| {\bm L}_\mulg \| } / {  \| \emsm \| } \lesssim ({ 1 - \eta^{(\mulg)} }) / ({ 1 + \eta^{(\mulg)} }),
\eeq
where ${\bm L}_{\mulg}, \emsm$ are defined in \eqref{eq:HB_dec}. Now, if $\meta{\mulg} \approx 1$, we have ${\cal H}( \mA{\mulg} ) {\bm B} \approx \emsm$ for any $\mulg = 1,\ldots,\mulG$. Consequently, the correlation matrix ${\bm Y}^\top {\bm Y}$ does not have the anticipated block structure that is necessary for successful clustering.\vspace{-.3cm}
\end{Remark}

\section{Batch EM Algorithm}\label{sec:batchEM}\vspace{-.2cm}
This section develops a customized EM algorithm that is efficient to implement and enjoys desirable theoretical convergence properties.
Particularly, we focus on \emph{batch} data where $m$ samples $\{ {\sf DP}_{t} \}_{t=1}^m$ are available all at once.
To begin, let us fix $\widetilde{\Theta} =\big\{\{\widetilde{\bm L}_\mulg\}_{\mulg=1}^\mulG,\widetilde{\bm B}_\rho,\{\widetilde{P}_\mulg\}_{\mulg=1}^\mulG\big\}$ and denote the conditional probability mass function for the latent r.v.~$w_t$ as $q(\cdot| \widetilde{\Theta}, {\sf DP}_t )$. The Jensen's inequality implies the following lower bound on the log-likelihood term in \eqref{eq:MAP}:
\begin{align}
    & \log p({\bm \op}_t | \Theta, {\bm \latp}_t, {\bm \dmp}_t ) 
    \label{eq:Jensen}
    \\
    &=\log \left( \sum_{\mulg=1}^\mulG p({\bm \op}_t, w_t = \mulg |\Theta, {\bm \latp}_t, {\bm \dmp}_t) \frac{q(w_t=\mulg| \widetilde{\Theta}, {\sf DP}_t )}{q(w_t = \mulg | \widetilde{\Theta}, {\sf DP}_t )} \right) \nonumber
    \\
    & \geq \EE_{ w_t\sim q( \cdot | \widetilde{\Theta},{\sf DP}_t ) } \big[ \log p({\bm \op}_t, w_t |\Theta, {\bm \latp}_t, {\bm \dmp}_t)\big] + \zeta( \widetilde{\Theta} ), \nonumber 
\end{align}
where $\zeta( \widetilde{\Theta} )$ is a function that only depends on the fixed $\widetilde{\Theta}$.
Taking the batch data setting into consideration, the above led us to the batch surrogate optimization problem:
\begin{align}
\ds \max_{\Theta}~
& \textstyle \frac{1}{m} \sum_{t=1}^m \mathbbm{E}_{ w_t\sim q( \cdot | \widetilde{\Theta},{\sf DP}_t ) }[\log p({\bm\op}_t,w_t |\Theta,{\bm\latp}_t,{\bm \dmp}_t)] \notag \\
& \textstyle - \lambda_S \| \emsm \|_1 - \lambda_L \sum_{\mulg=1}^{\mulG} \| {\bm L}_{\mulg} \|_\star \label{eq:newop} \\[.1cm]
\text{s.t.}~ & \textstyle \sum_{\mulg=1}^{\mulG} P_{\mulg} = 1, P_{\mulg} \geq 0,~\mulg=1,\ldots,\mulG. \notag
\end{align}

Let us take a closer look at the first term in the objective function of \eqref{eq:newop}. For $t = 1, \ldots, m$, the Bayes' rule implies
\beq \notag
\begin{split} 
& \EE_{ w_t\sim q( \cdot | \widetilde{\Theta},{\sf DP}_t ) }[\log p({\bm \op}_t,w_t |\Theta,{\bm \latp}_t,{\bm \dmp}_t)] \\
& = \EE_{w_t}[ \mathbbm{1}(w_t=\mulg) \log \mathbbm{P}(w_t = \mulg) p({\bm \op}_t |\Theta,{\bm \latp}_t,{\bm \dmp}_t, w_t = \mulg )] \\
& = \EE_{w_t}  \big[ \mathbbm{1}(w_t=\mulg) \{\log(P_\mulg) + \log p({\bm \op}_t|{\emsm} ,{\bm L}_\mulg, {\bm \latp}_t, {\bm\dmp}_t)\} \big],
\end{split}
\eeq
where we used $\mathbbm{P}( w_t = c ) = P_c$ and the dependence on $q( \cdot | \widetilde{\Theta},{\sf DP}_t )$ were omitted for brevity. Moreover, 
\begin{align}
& \log p({\bm \op}_t|{\emsm},{\bm L}_{\mulg},{\bm \latp}_t,{\bm \dmp}_t) =
\label{eq:logp} \\
& \frac{ \langle {\bm \dmp}_t \odot({\bm L}_{\mulg}+{\emsm}){\bm \latp}_t|{\bm \op}_t \rangle}{\sigma^2} - \frac{||{\bm \dmp}_t \odot({\bm L}_{\mulg}+{\emsm}){\bm \latp}_t||^2}{2 \sigma^2} + \widetilde{\zeta}_t ( \widetilde{\Theta} ). \notag
\end{align}
where $\widetilde{\zeta}_t (\widetilde{\Theta} ) := \zeta( \widetilde{\Theta} ) + \frac{1}{2\sigma^2} \| {\bm y}_t \|^2$.
The above expressions can be simplified as
\[ 
\begin{split}
& \langle {\bm \dmp}_t \odot({\bm L}_{\mulg}+{\emsm}){\bm \latp}_t|{\bm \op}_t \rangle 
= {\rm Tr}( ({\bm L}_{\mulg}+{\emsm})^\top \, {\bm \dmp}_t \odot {\bm \op}_t {\bm \latp}_t^\top  ), \\[.1cm]
& ||{\bm \dmp}_t \odot({\bm L}_{\mulg}+{\emsm}){\bm \latp}_t||^2 \\
& \textstyle = \sum_{i=1}^n {\rm Tr} \big( ( {\bm L}_{\mulg}+{\emsm} )^\top {\bm e}_i {\bm e}_i^\top ( {\bm L}_{\mulg}+{\emsm} ) \, \Omega_{t,i}  {\bm z}_t {\bm z}_t^\top \big).
\end{split}
\]

Define the conditional probability for the event that the $t$th data tuple is associated to the $\mulg$th graph: 
\begin{align}
& p_c(\widetilde{\Theta} , {\sf DP}_t) := \EE_{ w_t \sim q( \cdot | \widetilde{\Theta},{\sf DP}_t ) } \big[ \mathbbm{1} (w_t=\mulg) \big] \label{eq:w} \\
& = 
\frac{\widetilde{P}_\mulg\exp(\frac{-1}{2\sigma^2}||({\bm \op}_t -{\bm \dmp}_{t}\odot( \widetilde{\bm B}_{\rho} + \widetilde{\bm L}_\mulg ) {\bm \latp}_t)||^2 )}{\sum_{\mulg'=1}^\mulG \widetilde{P}_{\mulg'} \exp( \frac{-1}{2\sigma^2} || ({\bm \op}_t - {\bm \dmp}_{t} \odot( \widetilde{\bm B}_{\rho} + \widetilde{\bm L}_{\mulg'} ) {\bm \latp}_t) ||^2 )} . \notag
\end{align} 
and the sufficient statistics:
\allowdisplaybreaks
\beq \label{eq:suffstat}
\begin{aligned}
& \textstyle \overline{P}_\mulg^{\widetilde{\Theta}} = \frac{1}{m} \sum_{t=1}^m p_c(\widetilde{\Theta} , {\sf DP}_t) , \\
& \textstyle \overline{{\bm \ops}{\bm \latps}}_\mulg^{\widetilde{\Theta}} = \frac{1}{m} \sum_{t=1}^m p_c(\widetilde{\Theta} , {\sf DP}_t) \, {\bm \dmp}_t \odot {\bm \op}_t {\bm \latp}_t^\top , \\
& \textstyle \overline{{\bm \latps}{\bm \latps}}_{\mulg,i}^{\widetilde{\Theta}} = \frac{1}{m} \sum_{t=1}^m p_c(\widetilde{\Theta} , {\sf DP}_t) \, \dmp_{t,i} {\bm \latp}_t {\bm \latp}_t^\top . 
\end{aligned}
\eeq 
For any $\widetilde{\Theta} \in \mathfrak{T}$, the lower bound surrogate objective function of \eqref{eq:newop} can be written as
\beq \label{eq:surrogate}
\begin{aligned}
& \textstyle \widetilde{\mathcal{L}}(\Theta \, | \, {\widetilde{\Theta}}  )= \widetilde{\zeta} ( \widetilde{\Theta} ) -\lambda_S||\emsm||_1 - \sum_{\mulg=1}^\mulG \lambda_L ||{\bm L}_\mulg||_\ast \\
&+ \sum_{\mulg=1}^\mulG \Bigg\{\overline{P}_\mulg^{\widetilde{\Theta}} \log(P_\mulg)+\frac{1}{\sigma^2} {\rm Tr} \big( ({\bm L}_{\mulg}+\emsm)^\top \, \overline{{\bm \ops}{\bm \latps}}_\mulg^{\widetilde{\Theta}} \big) \\
&-\frac{1}{2\sigma^2}\sum_{i=1}^n\Tr\big( ({\bm L}_\mulg+\emsm)^\top{\bm e}_i {\bm e}_i^\top({\bm L}_\mulg+\emsm) \overline{{\bm \latps}{\bm \latps}}_{\mulg,i}^{\widetilde{\Theta}} \big) \Bigg\},
\end{aligned}
\eeq 
which is a concave function in $\Theta$ and we have defined $\widetilde{\zeta} ( \widetilde{\Theta} ) = (1/m )\sum_{t=1}^m \widetilde{\zeta}_t ( \widetilde{\Theta} )$. For any $\Theta, \widetilde{\Theta} \in \mathfrak{T}$, notice that it holds
\beq \label{eq:Llower}\textstyle
{\cal L}( {\Theta} ) \geq \widetilde{\mathcal{L}}(\Theta \, | \, {\widetilde{\Theta}} ),~~{\cal L}( {\Theta} ) = \widetilde{\mathcal{L}}(\Theta \, | \, {{\Theta}}  ).
\eeq 
The above derivations led us to a regularized (batch) EM algorithm. In particular, we initialize by fixing ${\Theta}^0$ and evaluate the sufficient statistics using \eqref{eq:suffstat}. Then, we alternate between the {\sf M}-step and the {\sf E}-step --- in the {\sf M}-step, we optimize w.r.t.~$\Theta$ for the surrogate problem \eqref{eq:newop}; in the {\sf E}-step, we update the sufficient statistics using \eqref{eq:suffstat} through the new $\Theta$. The overall algorithm is summarized in \Cref{alg:em}. 

\algsetup{indent=.5em}
\begin{algorithm}[t]
\caption{Batch EM for Partial Inference on \eqref{eq:mixmodel}}\label{alg:em}
\begin{algorithmic}[1]
\STATE \textbf{Input}: graph signals ${\bm Y}$, excitation parameters ${\bm Z}$, missing-information vectors $\{{\bm \dmp}_t\}_{t=1}^m$, no.~of graphs $\mulG$.
\IF{$\Theta^0 := \big\{ \{ {\bm L}_\mulg^0 \}_{\mulg=1}^\mulG, \emsm^0, \{ P_\mulg^0 \}_{\mulg=1}^\mulG \big\}$ is available}
\STATE Evaluate \eqref{eq:w}, \eqref{eq:suffstat}. 
\ELSIF{$p_{\mulg} (\Theta^0, {\sf DP}_t), t = 1,\ldots,m$ are available}
\STATE Evaluate \eqref{eq:suffstat}. \label{line:init_a}
\ENDIF
\FOR{$k=1,2,\dots,K_{\sf max}$}
\STATE {\sf M-step}: solve the concave maximization
\beq \label{eq:gen_Mstep}
\begin{array}{rl}
\ds \Theta^{k} \in \argmax_{\Theta} & \widetilde{\mathcal{L}} (\Theta \, | \, \Theta^{k-1} ) \\
\text{s.t.} & \sum_{\mulg=1}^{\mulG} P_{\mulg} = 1, P_{\mulg} \geq 0,~\forall~\mulg, 
\end{array} 
\eeq
see \eqref{eq:rpca} for efficient implementation in Gaussian case.
\STATE {\sf E-step}: evaluate $p_\mulg (\Theta^k, {\sf DP}_t)$ for all $t$ using \eqref{eq:w}, and the sufficient statistics using \eqref{eq:suffstat}.
\ENDFOR
\STATE \textbf{Output}: converged parameters $\Theta^{K_{\sf max}}$ and conditional probabilities $p_\mulg ( \Theta^{K_{\sf max}}, {\sf DP}_t )$ for all $t$.
\end{algorithmic}
\end{algorithm}

Denote ${\tt D}( \Theta | \widetilde{\Theta} ) := {\cal L}( {\Theta} ) - \widetilde{\mathcal{L}}(\Theta \, | \, {\widetilde{\Theta}}) \geq 0$ as the difference function between ${\cal L}(\Theta)$ and the surrogate.
We observe:
\begin{Prop} \label{prop:conv}
Consider the sequence $\{ \Theta^k \}_{k \geq 0}$ generated by \Cref{alg:em}. The following holds:
\begin{enumerate}[leftmargin=*]
\item The regularized log-likelihood value is non-decreasing:
\beq\label{eq:nondecr}
{\cal L}( \Theta^{k+1} ) \geq {\cal L}( \Theta^k ),~\forall~k \geq 0.
\eeq 
\item If the gradient w.r.t.~$\Theta$ for the difference function ${\tt D}( \Theta | \widetilde{\Theta} )$ is $L$-Lipschitz continuous, then for any $K_{\sf max} \geq 1$, 
\beq 
\min_{ k=1,\ldots,K_{\sf max} } \| \grd_{\Theta} {\tt D}( \Theta^k | \Theta^{k-1} )  \|^2 = {\cal O} (L / K_{\sf max}). 
\eeq 
In addition, the directional derivative:
\beq 
{\cal L}'( \widetilde{\Theta} ; {\Theta} - \widetilde{\Theta} ) := \lim_{ \delta \to 0 } \frac{ {\cal L}( \widetilde{\Theta} + \delta ( {\Theta} - \widetilde{\Theta} ) ) -  {\cal L}( \widetilde{\Theta} ) }{ \delta }
\eeq 
exists for any $\Theta, \widetilde{\Theta} \in \mathfrak{T}$. Thus, 
\beq \label{eq:stationary}
\hspace{-.1cm}\min_{ k=1,\ldots,K_{\sf max} } \sup_{ \Theta \in \mathfrak{T} } \frac{{\cal L}'( \Theta^k ; \Theta - \Theta^k )}{ \| \Theta^k - \Theta \| } = {\cal O} \left(\sqrt{ \frac{L}{K_{\sf max}} } \right).
\eeq 
\end{enumerate}
\end{Prop}
Note that if ${\cal L}'( \overline{\Theta} ; {\Theta} - \overline{\Theta} ) \leq 0$ for all $\Theta \in \mathfrak{T}$, then $\overline{\Theta}$ is a stationary point to the MAP problem \eqref{eq:MAP}. As such, \Cref{alg:em} finds a stationary point to \eqref{eq:MAP} at a sublinear rate. A key challenge in our analysis is that the MAP problem \eqref{eq:MAP} is non-smooth due to the sparse/low-rank priors in \eqref{eq:prior}. We achieve the proof through extending \cite{mairal2015incremental, karimi2022minimization}, see Appendix~\ref{app:EM}. 
\vspace{.1cm}

\noindent \textbf{Implementation Details.} We comment on the {\sf M}-step. First, the maximizer for $\{P_\mulg\}_{\mulg=1}^\mulG$ is given by: 
\beq \label{eq:pcupdate}
P_c^\star = \overline{P}_\mulg^{\widetilde{\Theta}} ( {\textstyle \sum_{\mulg'=1}^\mulG \overline{P}_{\mulg'}^{\widetilde{\Theta}} } )^{-1}   ,~~\mulg = 1,\dots, \mulG.
\eeq
Second, the parameters $\{\emsm,\{{\bm L}_\mulg\}_{\mulg=1}^\mulG\}$ can be obtained through solving the regularized least square problem:
\begin{align}\label{eq:rpca} 
 &\min_{ \{ {\bm L}_\mulg \}_{\mulg=1}^\mulG, \emsm }~  \lambda_L \sum_{\mulg=1}^\mulG || {\bm L}_\mulg ||_\ast + \lambda_S || \emsm ||_1 + \frac{1}{2 \sigma^2} \Bigg\{ \\
 & \quad \sum_{\mulg=1}^\mulG \sum_{i=1}^n {|| {\bm e}_i^\top ({\bm L}_\mulg + \emsm ) \overline{\bm Z}_{\mulg,i}^{\widetilde{\Theta}} - {\bm e}_i^\top \overline{{\bm \ops} {\bm \latps}}_\mulg^{\widetilde{\Theta}} (\overline{\bm Z}_{\mulg,i}^{\widetilde{\Theta}})^{-1} ||^2_F} \Bigg\},\notag
\end{align}
where $\overline{\bm Z}_{\mulg,i}^{\widetilde{\Theta}}$ is the matrix square root of ${\overline{{\bm \latps}{\bm \latps}}_{\mulg,i}^{\widetilde{\Theta}}}$. Note \eqref{eq:rpca} is a convex robust PCA problem \cite{agarwal2012noisy} which can be efficiently solved by available software such as \cite{agarwal2012noisy}. 
\Cref{alg:em} also supports initialization in the absence of $\Theta^0$. In fact, it suffices to initialize the algorithm through evaluating the sufficient statistics in the {\sf E}-step. For the latter, we estimate the conditional probability that the $t$th data tuple is associated to the $c$th graph, e.g. by applying the na\"{i}ve SC. The sufficient statistics can then be found using  \eqref{eq:suffstat}. 

Finally, we demonstrate how to tackle \Cref{prob:jg} using the outputs from \Cref{alg:em}. The operations are straightforward: (i) the eigen-centrality can be estimated by applying SVD on the matrices $\{{\bm L}_\mulg^{K_{\sf max}} \}_{\mulg=1}^\mulG$ and extract the top left singular vectors;
(ii) the graph identifiers are estimated by taking 
\beq \textstyle
\hat{w}_t = \argmax_{\mulg=1,\dots,\mulG}~ p_c( \Theta^{K_{\sf max}}, {\sf DP}_t ),
\eeq 
for all $t=1,\ldots,m$.
\vspace{-.4cm}

\subsection{Extension to Logit Model}\label{sec:logit}\vspace{-.2cm}
We conclude this section by extending Algorithm~\ref{alg:em} to tackling Problem~\ref{prob:jg} with \emph{binary} graph signals observations. For example, this applies if the latter consists of vote data. 
Consider the case without missing data, i.e., ${\bm \dmp}_t = {\bf 1}$ and focus on a logit observation model. The observed data $\{ {\sf DP}_t \}_{t=1}^m = \{ {\bm y}_t, {\bm z}_t \}_{t=1}^m$ satisfy: 
\beq\label{eq:logitmodel}
\mathbbm{P}(\op_{t,j}=Y) =\frac{\exp((\widetilde{\op}_{t,j} + b)Y)}{1+\exp(\widetilde{\op}_{t,j} + b)},~~Y \in \{0,1\},
\eeq
for $j = 1,\ldots,n$,
where $b < 0$ is the bias parameter of the logit model and $\widetilde{\op}_{t,j}$ is the $j$th element of the vector:
\beq \textstyle
\widetilde{\bm y}_t = \sum_{\mulg = 1}^{\mulG} \mathbbm{1} (w_t=\mulg) {\cal H}(\mA{\mulg}) {\bm B} {\bm z}_t.
\eeq 
Similar to Sec.~\ref{sec:MAP}, we further adopt the parameterization with ${\cal H}(\mA{\mulg}) {\bm B} = {\bm L}_c + \emsm$.

The EM algorithm on the above model can be developed similarly as Algorithm~\ref{alg:em}. In particular, the derivations up to \eqref{eq:Jensen} remain valid. Now, denote the conditional probability of the graph identifier $w_t$ [cf.~\eqref{eq:w}] for the $t$th data tuple as: 
\begin{align}\label{eq:logitw}
& p_\mulg^{\sf lg} (\widetilde{\Theta},{\sf DP}_t) 
= \frac{\widetilde{P}_\mulg\Pi_{j=1}^n\frac{\exp(\op_{t,j}\widetilde{\nu}_{t,j,\mulg})}{1+\exp(\widetilde{\nu}_{t,j,\mulg} )}}
{\sum_{\mulg'=1}^\mulG \widetilde{P}_{\mulg'}\Pi_{j=1}^n\frac{\exp(\op_{t,j}\widetilde{\nu}_{t,j,\mulg'})}{1+\exp(\widetilde{\nu}_{t,j,\mulg'})}},
\end{align}
where $\widetilde{\nu}_{t,j,\mulg} := b+ {\bm e}_j^\top (\widetilde{\bm L}_\mulg + \widetilde{\bm B}_{\rho} ) {\bm \latp}_t$. We observe that the following surrogate objective function lower bounds the MAP objective function with the logit model \eqref{eq:logitmodel}:
\beq\label{eq:logitsurrogate}
 \begin{aligned}
 & \widetilde{\mathcal{L}}_{\sf logit}(\Theta \, | \, {\widetilde{\Theta}}  )= {\rm constant} -\lambda_S||\emsm||_1 - \sum_{\mulg=1}^\mulG \lambda_L ||{\bm L}_\mulg||_\ast\\
& + \frac{1}{m} \sum_{t=1}^\mulG \sum_{t=1}^m p_\mulg^{\sf lg} ( \widetilde{\Theta} , {\sf DP}_t ) \Bigg\{ \log(P_\mulg) + \sum_{j=1}^n 
\op_{t,j} \nu_{t,j,\mulg} \Bigg\} \\
& - \frac{1}{m} \sum_{\mulg=1}^\mulG \sum_{t=1}^m p_\mulg^{\sf lg} (\widetilde{\Theta} , {\sf DP}_t ) \sum_{j=1}^n   \log(1+\exp(\nu_{t,j,\mulg})) ,
\end{aligned}
\eeq
where $\nu_{t,j,\mulg} := b+ {\bm e}_j^\top ({\bm L}_\mulg+\emsm) {\bm \latp}_t$ is a linear function of the decision variables ${\bm L}_c, \emsm$. 

We observe that \eqref{eq:logitsurrogate} is a concave function in $\Theta$. To develop the EM algorithm, the {\sf M}-step in Algorithm~\ref{alg:em} can now be replaced by maximizing \eqref{eq:logitsurrogate} w.r.t.~$\Theta$ when $\Theta^{k-1}$ is given. On the other hand, {\sf E}-step only involves evaluating $p_\mulg^{\sf lg} (\widetilde{\Theta} , {\sf DP}_t )$ according to \eqref{eq:logitw}. 
Compared to the case with Gaussian observation, the {\sf M}-step involves \eqref{eq:logitsurrogate} which is a finite-sum problem that can be difficult to optimize when $m \gg 1$. This is caused by the nonlinear log-likelihood function associated with the logit model \eqref{eq:logitmodel}.\vspace{-.2cm}

\section{Online EM Algorithm}\label{sec:onlineEM}\vspace{-.1cm}
This section considers tackling Problem~\ref{prob:jg} under \emph{streaming data}. We focus on an online learning process where the data tuple is revealed sequentially. Particularly, at time $t$, we only observe the data tuple ${\sf DP}_t = \{ {\bm \op}_t, {\bm \latp}_t, {\bm \dmp}_t \}$ that is generated from the model \eqref{eq:mixmodel} in an i.i.d.~fashion. 

We aim to design an \emph{online} algorithm for the MAP problem \eqref{eq:MAP} with stochastic log-likelihood objective. Consider the following surrogate problem at the $t$th iteration:
\begin{align}
\ds \max_{\Theta}~
& \ds \EE_{\sf DP, W \sim q( \cdot | {\Theta}^{t-1},{\sf DP} ) } \big[\log p({\bm Y} , W |\Theta,{\bm Z},{\bm \dmp} ) \big] - \Psi( \Theta) \notag
\end{align}
\beq \label{eq:newop_stoc}
\begin{aligned}
 \text{with}~~\Psi(\Theta) & := \textstyle \lambda_S \| \emsm \|_1 + \lambda_L \sum_{\mulg=1}^{\mulG} \| {\bm L}_{\mulg} \|_\star \\
 & \textstyle + \epsilon \sum_{\mulg=1}^{\mulG-1} \log(P_\mulg) + \epsilon \log( 1-\sum_{\mulg=1}^{\mulG-1}P_\mulg ),
\end{aligned}
\eeq
where the expectation $\EE_{\sf DP}[\cdot]$ is taken w.r.t.~the random generative model for the data tuple ${\sf DP} = \{ {\bm Y}, {\bm Z}, {\bm \dmp} \}$, and $\lambda_S, \lambda_L, \epsilon > 0$ are regularization parameters.
Compared to \eqref{eq:newop}, the additional regularizer on $\{ P_c \}_{c=1}^{C}$ enforces the latter to be in the interior of the simplex set.  

The surrogate objective function of \eqref{eq:newop_stoc} admits a similar form as \eqref{eq:surrogate}, which is derived as (constants are omitted)
\begin{align}
& \widetilde{\cal L}_{\sf ol} ( \Theta ; \overline{P}_\mulg^{{\Theta}^{t-1}}, \overline{{\bm \ops}{\bm \latps}}_\mulg^{{\Theta}^{t-1}}, \overline{{\bm \latps}{\bm \latps}}_{\mulg,i}^{{\Theta}^{t-1}}, \forall c,i ) := - \Psi(\Theta) \label{eq:surrogate_2stoc} \\
 & \textstyle + \sum_{\mulg=1}^\mulG \big\{\overline{P}_\mulg^{{\Theta}^{t-1}} \log(P_\mulg)+\frac{1}{\sigma^2} {\rm Tr} \big( ({\bm L}_{\mulg}+\emsm)^\top \, \overline{{\bm \ops}{\bm \latps}}_\mulg^{{\Theta}^{t-1}} \big) \notag \\
 & \textstyle -\frac{1}{2\sigma^2}\sum_{i=1}^n\Tr\big( ({\bm L}_\mulg+\emsm)^\top{\bm e}_i {\bm e}_i^\top({\bm L}_\mulg+\emsm) \overline{{\bm \latps}{\bm \latps}}_{\mulg,i}^{{\Theta}^{t-1}} \big) \big\}, \notag 
\end{align}
where we have defined the \emph{population sufficient statistics} as:
\begin{align}
& \textstyle \overline{P}_\mulg^{{\Theta}^{t-1}} = \EE_{\sf DP}[ p_c({\Theta}^{t-1} , {\sf DP}) ] , \notag \\
& \overline{{\bm \latps}{\bm \latps}}_{\mulg,i}^{{\Theta}^{t-1}} = \EE_{\sf DP}[ p_c({\Theta}^{t-1} , {\sf DP}) \, \dmp_{i} {\bm Z} {\bm Z}^\top ], \label{eq:suffstat_stoc}  \\
& \textstyle \overline{{\bm \ops}{\bm \latps}}_\mulg^{{\Theta}^{t-1}} = \EE_{\sf DP}[ p_c({\Theta}^{t-1} , {\sf DP} ) \, {\bm \dmp} \odot {\bm Y} {\bm Z}^\top ], \notag
\end{align}
such that $p_c({\Theta}^{t-1} , {\sf DP} )$ was defined in \eqref{eq:w}. Note that \eqref{eq:surrogate_2stoc}, \eqref{eq:suffstat_stoc} generalize \eqref{eq:suffstat}, \eqref{eq:surrogate} to observations drawn from any distribution. To see this, \eqref{eq:suffstat} can be recovered from \eqref{eq:suffstat_stoc} as the special case with empirical distribution. 

Following the development of the batch EM algorithm, we wish to maximize \eqref{eq:surrogate_2stoc} w.r.t.~$\Theta$ at the {\sf M}-step. 
However, unlike \eqref{eq:suffstat}, computing \eqref{eq:suffstat_stoc} is challenging as we are observing the data tuple on-the-fly.
To this end, we adopt the stochastic approximation (SA) scheme \cite{robbins1951stochastic} from \cite{cappe2009line, karimi2019non} on the space of \emph{sufficient statistics} to dynamically track \eqref{eq:suffstat_stoc}.\vspace{.1cm}

\noindent \textbf{SA Scheme for \eqref{eq:suffstat_stoc}.} Let $\overline{P}_\mulg^{t-1}, \overline{{\bm \ops}{\bm \latps}}_\mulg^{t-1}, \overline{{\bm \latps}{\bm \latps}}_{\mulg,i}^{t-1}$ be the estimate for the sufficient statistics at iteration $t-1$, we consider the following SA scheme to estimate \eqref{eq:suffstat_stoc}:
\begin{align}\label{eq:SAsuffstat}
& \textstyle \overline{P}_\mulg^{t} = \overline{P}_\mulg^{t-1} + \beta_t( p_\mulg( \Theta^{t-1} , {\sf DP}_t) - \overline{P}_\mulg^{t-1}) , \\
& \textstyle \overline{{\bm \ops}{\bm \latps}}_\mulg^{t} = \overline{{\bm \ops}{\bm \latps}}_\mulg^{t-1}+\beta_t( p_\mulg( \Theta^{t-1} , {\sf DP}_t) {{\bm \dmp}_t \odot}  {\bm \op}_t{\bm \latp}_t^\top-\overline{{\bm \ops}{\bm \latps}}_\mulg^{t-1}), \notag \\
& \textstyle \overline{{\bm \latps}{\bm \latps}}_{\mulg,i}^{t} = \overline{{\bm \latps}{\bm \latps}}_{\mulg,i}^{t-1} + \beta_t( p_\mulg( \Theta^{t-1} , {\sf DP}_t) \dmp_{t,i}{\bm \latp}_t{\bm \latp}_t^\top-\overline{{\bm \latps}{\bm \latps}}_{\mulg,i}^{t-1}) ,\notag
\end{align}
for $i=1, \ldots, n$, $c=1,\ldots,\mulG$, where $\beta_t \in ( 0, 1]$ is the step size. 
Notice that the SA scheme only uses the current data ${\sf DP}_t$ available in the streaming data setting, where it replaces the {\sf E}-step in the EM algorithm. The above estimates are then used in \eqref{eq:surrogate_2stoc} to construct the surrogate $\widetilde{\cal L}_{\sf ol}( \Theta;  \overline{P}_\mulg^{t}, \overline{{\bm \ops}{\bm \latps}}_\mulg^{t}, \overline{{\bm \latps}{\bm \latps}}_{\mulg,i}^{t} , \forall c,i )$, whose maximization leads to the {\sf M}-step.

To understand \eqref{eq:SAsuffstat}, let us focus on $\overline{P}_\mulg^{t}$ for the illustration purpose. Herein, the \emph{mean field} of SA update is given by the expected value of the drift term conditioned on iterates up to the $t-1$th iteration. The latter is
\beq \notag
\EE_{t-1} [ p_\mulg( \Theta^{t-1} , {\sf DP}_t) - \overline{P}_\mulg^{t-1} ] = \EE_{\sf DP} [ p_\mulg( \Theta^{t-1} , {\sf DP} )] - \overline{P}_\mulg^{t-1}.
\eeq  
Substituting into \eqref{eq:suffstat_stoc} shows that in expectation, $\overline{P}_\mulg^{t}$ is a convex combination of $\overline{P}_\mulg^{t-1}$ and $\overline{P}_\mulg^{\Theta^{t-1}}$. 
In other words, the recursion \eqref{eq:SAsuffstat} drives the sufficient statistics estimates towards \eqref{eq:suffstat_stoc}. 

\begin{algorithm}[t]
\caption{Online EM for Partial Inference on \eqref{eq:mixmodel}}\label{alg:olem}
\begin{algorithmic}[1]
\STATE \textbf{Input}: no. of graphs $\mulG$, initial parameters $\Theta^{0}$ and sufficient statistics $\{ \overline{P}_\mulg^{0},\overline{ {\bm \ops} {\bm \latps}}_\mulg^{0},\{\overline{{\bm \latps}{\bm \latps}}_{\mulg,i}^{0}\}_{i=1}^n \}_{\mulg=1}^\mulG$.
\FOR{$t=1,2,\dots$}
\STATE \label{line:sample} {\sf Sample}: ${\sf DP}_t = \{ {\bm \op}_t$, ${\bm \latp}_t$, ${\bm \dmp}_t \}$ according to \eqref{eq:mixmodel}.
\STATE \label{line:outlier} \texttt{// Optional: Anomaly Detection //}
$ 
\begin{cases}
\text{go to line~\ref{line:estep}}, & \text{if \eqref{eq:detector_p} outputs}~{\cal H}_0, \\
\text{go to line~\ref{line:sample}}, & \text{if \eqref{eq:detector_p} outputs}~{\cal H}_1. 
\end{cases}\vspace{.2cm}
$
\STATE \label{line:estep} {\sf E-step}: update the sufficient statistics 
via \eqref{eq:SAsuffstat}.
\STATE \label{line:mstep_o} {\sf M-step}: maximize the surrogate function through solving
\beq
\Theta^{t} \in \argmax_{\Theta}~ \widetilde{\cal L}_{\sf ol}( \Theta;  \overline{P}_\mulg^{t}, \overline{{\bm \ops}{\bm \latps}}_\mulg^{t}, \overline{{\bm \latps}{\bm \latps}}_{\mulg,i}^{t} , \forall c,i ).\notag
\eeq
\ENDFOR
\end{algorithmic}
\end{algorithm}

Equipped with the above derivations, we summarize the online EM algorithm in Algorithm~\ref{alg:olem}. Note that the algorithm is fully online as it does not store the history of $\{ {\sf DP}_t \}_{t \geq 0}$. Instead, information from the latter is absorbed by the sufficient statistics estimates in each iteration. Lastly, though Algorithm~\ref{alg:olem} bears similarities to \cite{cappe2009line, karimi2019non}, our algorithm incorporates a set of non-smooth regularizers, i.e., $\| \emsm \|_1$, $\| {\bm L}_\mulg \|_\star$ that are motivated by the graph signal model.

Lastly, let us comment on the fixed point for the recursions \eqref{eq:suffstat_stoc}. Note that a fixed point $(\overline{P}_\mulg, \overline{ {\bm \ops} {\bm \latps}}_\mulg, \overline{{\bm \latps}{\bm \latps}}_{\mulg,i})$ for the recursion satisfies for any $\mulg = 1, \ldots, \mulG$ that
\beq \label{eq:satpoint}
\begin{split}
& \EE_{\sf DP} [ p_\mulg( \overline{\Theta} , {\sf DP} )] - \overline{P}_\mulg = 0, \\
& \EE_{\sf DP} [ p_\mulg( \overline{\Theta} , {\sf DP} ) \bm{\dmp} \odot {\bm \ops} {\bm \latps}^\top ] - \overline{ {\bm \ops} {\bm \latps}}_\mulg = 0, \\
& \EE_{\sf DP} [ p_\mulg( \overline{\Theta} , {\sf DP} ) \Omega_i {\bm \latps} {\bm \latps}^\top ] - \overline{{\bm \latps}{\bm \latps}}_{\mulg,i} = 0,~\forall~i, \\
\end{split}
\eeq 
where $\overline{\Theta} \in \argmax_{\Theta} \widetilde{\cal L}_{\sf ol}( \Theta;  \overline{P}_\mulg, \overline{{\bm \ops}{\bm \latps}}_\mulg, \overline{{\bm \latps}{\bm \latps}}_{\mulg,i} )$.
Observe:
\begin{Prop} \label{prop:fixedpt}
Let $\Gamma$ be the set of stationary solutions of the MAP problem with modified regularizer [cf.~\eqref{eq:MAP}, \eqref{eq:newop_stoc}]:
\beq \label{eq:MAPa}
\max_{ \Theta }~\EE_{\sf DP}[ \log p ( {\bm Y} | \Theta, {\bm Z}, \bm{\Omega} ) ] - \Psi(\Theta),
\eeq 
If the sufficient statistics $(\overline{P}_\mulg, \overline{ {\bm \ops} {\bm \latps}}_\mulg, \overline{{\bm \latps}{\bm \latps}}_{\mulg,i}, \forall \mulg,i)$ satisfies \eqref{eq:satpoint}, then $\overline{\Theta} \in \Gamma$. Conversely, assume in addition, the maximizer of $\widetilde{\cal L}_{\sf ol}( \Theta; {P}_\mulg, { {\bm \ops} {\bm \latps}}_\mulg, {{\bm \latps}{\bm \latps}}_{\mulg,i})$ is unique for any sufficient statistics [cf.~line~\ref{line:mstep_o} of \Cref{alg:olem}]. Then if $\overline{\Theta} \in \Gamma$, the tuple $(\overline{P}_\mulg, \overline{ {\bm \ops} {\bm \latps}}_\mulg, \overline{{\bm \latps}{\bm \latps}}_{\mulg,i}, \forall \mulg,i)$ satisfies \eqref{eq:satpoint}.
\end{Prop}
\noindent The proof, which extends \cite{cappe2009line, nguyen2022online} to the regularized MAP setting in \eqref{eq:MAPa}, is relegated to Appendix~\ref{app:fixedpt}.

The above proposition shows that if the SA recursion \eqref{eq:SAsuffstat} converges to a fixed point, then such fixed point must lead to the parameter $\overline{\Theta}$ stationary to the MAP problem \eqref{eq:MAPa}. The convergence of \eqref{eq:SAsuffstat} to a fixed point typically requires
\beq \textstyle 
\sum_{t=1}^\infty \beta_t = \infty,~\sum_{t=1}^\infty \beta_t^2 < \infty,
\eeq 
and additional conditions such as Lipschitz continuity of the population sufficient statistics map \eqref{eq:suffstat_stoc}. In the interest of space, the readers are referred to \cite{kushner2003stochastic, karimi2019non} for details.
We remark that the stochastic gradient EM algorithm in \cite{balakrishnan2017statistical} is an alternative to \Cref{alg:olem}. However, \cite{balakrishnan2017statistical} applies stochastic gradient in the parameter ($\Theta$) space, which can be less computationally efficient. \vspace{-.2cm}

\subsection{Online Joint Inference \& Anomaly Detection} \label{sec:anom} \vspace{-.1cm}
We conclude by discussing an application of \Cref{alg:olem} to \emph{online} {joint inference and anomaly detection} of graph signals that are not generated from one of the candidate graphs, $G^{(\mulg)}$, $\mulg = 1,\ldots, \mulG$, in \eqref{eq:mixmodel}. 
Detecting if graph signals are originated from an `abnormal' graph is an important task for, e.g., power systems, pathological signal detection, see \cite{isufi2018blind, kaushik2021network}.
While prior works require the normal graph topology to be known a-priori, our goal is to simultaneously perform graph inference through estimating central nodes and detect these abnormal graph signals. In this setting, an \emph{online} algorithm is preferred as we aim to detect anomalies as soon as possible. 

At time $t \geq 1$, we consider the graph signal ${\bm \op}_t$ (and the latent variable ${\bm \latp}_t$) satisfying ${\bm \op}_t = \bm{\dmp}_t \odot {\cal H}( {\bm A}_t ) {\bm B} {\bm z}_t + {\bm e}_t$ akin to  \eqref{eq:mixmodel}. Herein, ${\bm A}_t$ denotes the adjacency matrix of a graph $G_t$ that ${\bm \op}_t$ is originated from. 
Accordingly, ${\bm \op}_t$ is said to be a \emph{normal signal} if $G_t \in \{ G^{(\mulg)} \}_{\mulg=1}^{\mulG}$; conversely, the signal is said to be \emph{abnormal} if it is generated from an outlier graph $G_t \notin \{ G^{(\mulg)} \}_{\mulg=1}^{\mulG}$. We define the binary hypothesis classes:
\beq \label{eq:class_det}
\begin{aligned}
    \texttt{(normal)}~~&{\cal H}_0: G_t \in \{\mG{\mulg}\}_{\mulg=1}^\mulG ,
    \\ 
    \texttt{(abnormal)}~~&{\cal H}_1: G_t \notin \{\mG{\mulg}\}_{\mulg=1}^\mulG.
\end{aligned}
\eeq 
We shall work with cases where ${\cal H}_1$ occurs with a lower probability than ${\cal H}_0$ to allow for successful graph inference. Moreover, under ${\cal H}_1$, the outlier graph is sufficiently different from the normal graphs in terms of its eigencentrality.  

Under H\ref{assump:lowpass}, we expect the distance $\min_{ \mulg=1,...,\mulG }\| {\bm y}_t - \bm{\Omega}_t \odot {\cal H}( {\bm A}^{(\mulg)} ) {\bm B} {\bm \latp}_t \|^2$ to be small under ${\cal H}_0$ and large under ${\cal H}_1$. While determining such distance would require knowledge of the normal graphs, we utilize the online EM algorithm and replace the latter using up-to-date estimates. This leads to the online detector: let $\kappa>0$ be a user-defined threshold,
\beq\label{eq:detector_p}
    \min_{\mulg=1,\dots,\mulG} ||{\bm \op}_t - {\bm \dmp}_t \odot ( {\bm L}^{t-1}_\mulg + {\bm B}^{t-1}_\rho ) {\bm \latp}_t ||^2 \mathop{\lessgtr}_{{\cal H}_1
    }^{{\cal H}_0} \kappa .
\eeq
Note that ${\bm L}^{t-1}_\mulg + {\bm B}^{t-1}_\rho \approx {\cal H}( {\bm A}^{(\mulg)} ) {\bm B}$ and the estimation quality improves as \Cref{alg:olem} gathers more data samples. We expect the detection performance to improve as $t$ grows. Finally, we incorporate the outlier rejection mechanism into \Cref{alg:olem} by a slight modification; see line~\ref{line:outlier}.\vspace{-.1cm}

\begin{Remark}
Compared to \cite{isufi2018blind, kaushik2021network}, our approach requires additional information on the excitation parameter ${\bm \latp}_t$. On the other hand, our approach is capable of simultaneous graph learning and abnormal graph detection.
\end{Remark}\vspace{-.4cm} 

\section{Numerical Experiments}\label{sec:exp}
In this section, we compare the performance of our EM algorithms on tackling \Cref{prob:jg} for synthetic and real data with state-of-the-art algorithms. \vspace{-.2cm}

\subsection{Experiments on Synthetic Data}  \label{sec:synexp}\vspace{-.2cm}
We describe the setup used throughout for synthetic data. We generate $\mulG$ \emph{core-periphery ({\sf CP}) graphs} with $n=100$ nodes. For $\mulg = 1, \dots, \mulG$, the node set ${\cal V} = \{1,...,n\}$ is partitioned into a core set ${\cal V}_{\sf o}^{(\mulg)}$ with size $|{\cal V}_{\sf o}^{(\mulg)}|=10$ and a non-core set ${\cal V}_{\sf p}^{(\mulg)} = {\cal V} \setminus {\cal V}_{\sf o}^{(\mulg)}$.  Each node in ${\cal V}_{\sf o}^{(\mulg)}$ is chosen uniformly at random such that ${\cal V}_{\sf o}^{(\mulg)}\neq {\cal V}_{\sf o}^{(\mulg')}$ if $\mulg \neq \mulg'$.
For any $i,j \in {\cal V}$, an edge is assigned with probability $1$ if $i,j \in {\cal V}_{\sf o}^{(\mulg)}$; with probability $0.2$ if $i \in {\cal V}_{\sf o}^{(\mulg)}, j \in {\cal V}_{\sf p}^{(\mulg)}$; and with probability $0.05$ if $i,j \in {\cal V}_{\sf p}^{(\mulg)}$.
Each observed signal ${\bm \op}_t$ is generated through the mixture model \eqref{eq:mixmodel} with the noise variance of $\sigma^2 = 10^{-2}$.
The graph identifier $w_t$ is drawn uniformly from $\{1,\dots,\mulG\}$. The missing information vector $\bm{\dmp}_t$ composes of Bernoulli r.v.s~with $\EE[ [\bm{\dmp}_t]_i ] = \gamma \in [0,1]$. 
For the excitation, the matrix ${\bm B}\in\Re^{n\times r}$ is generated as $B_{ij} = M_{ij} \widetilde{B}_{ij}$, where $M_{ij}, \widetilde{B}_{ij}$ are independent r.v.s, $M_{ij} \in \{0,1\}$ is Bernoulli  with $\EE[ M_{ij} ] = 0.1$, and $\widetilde{B}_{ij} \sim {\cal U}([0.1,1])$. The latent parameter matrix ${\bm z}_t \in\Re^{r}$ is generated by $[{\bm z}_t]_{i} = N_{it}\widetilde{Z}_{it}$, where $N_{it}$, $\widetilde{Z}_{it}$ are independent r.v.s, $N_{it} \in \{0,1\}$ is Bernoulli with $\EE[N_{it}] = 0.6$, and $Z_{it} \sim {\cal U}([0.1,1])$. 
Unless otherwise specified, the excitation rank will be set at $r=40$.

We evaluate the performance of ({\sf A}) central nodes detection and ({\sf B}) graph signals clustering. For {\sf (A)}, we compare the mismatch between the ground truth ${\cal V}_{\sf o}^{(\mulg)}, \mulg=1,\ldots,\mulG$ and the detected central nodes via the average error rate:
\beq \textstyle
{\sf Error~rate} = 1.0-(1/\mulG) \sum_{\mulg=1}^\mulG \EE \big[ \frac{1}{10} | {\cal V}_{\sf o}^{(\mulg)} \cap \widehat{\cal V}_{\sf o}^{(\mulg)} | \big],
\eeq 
where $\widehat{\cal V}_{\sf o}^{(\mulg)}$ is the top-10 central nodes detected in graph $G^{(\mulg)}$ with the algorithm.
For {\sf (B)}, we compute the normalized mutual information (NMI) \cite{vinh2010information}  between the detected graph identifiers $\{\hat{w}_t\}_{t=1}^m$ and the ground truth identifiers $\{w_t\}_{t=1}^m$. A large NMI value indicates a high clustering accuracy.
\vspace{.1cm}

\noindent \textbf{Batch Algorithms.}~We initialize \Cref{alg:em} by assigning the conditional probabilities using the SC method. Let $ {\bm U} \in \RR^{m \times C}$ be the collection of top-$\mulG$ eigenvectors of data correlation matrix ${\bm \ops}^\top {\bm \ops}$ and $\bar{\bm u}_\mulg$ is the centroid vector of the $\mulg$th cluster computed from SC. We set:
\beq\label{eq:initial}
p_\mulg(\Theta^0,{\sf DP}_t)= \frac{\exp(-||{\bm U}_{t,:}-\bar{\bm u}_\mulg||^2)}{\sum_{\mulg'=1}^\mulG\exp(-||{\bm U}_{t,:}-\bar{\bm u}_{\mulg'}||^2)},
\eeq
We remark that SC gives a good \emph{initialization} to \Cref{alg:em} despite that the method alone may not perform well on signals originated from weak low pass filters; see \Cref{ex:bad}.

The first experiment considers a batch data setting with $m=200\mulG$ signal tuples $\{ {\sf DP}_t \}_{t=1}^m$ generated from \eqref{eq:mixmodel}. The graph filters applied are ${\gfilter}_{\sf s} (\mA{w_t}) = ( {\bm I} - \frac{1}{40} \mA{w_t} )^{-1}$, ${\gfilter}_{\sf w} (\mA{w_t}) = ( {\bm I} - \frac{1}{80} \mA{w_t} )^{-1}$, where ${\cal H}_{\sf w}(\cdot)$ is a weaker low pass filter than ${\cal H}_{\sf s}(\cdot)$. We set $K_{\sf max} = 100$, $\lambda_L = 0.01$ and $\lambda_S = 0.001$.
We benchmark \Cref{alg:em} against GLMM \cite{maretic2020graph}, SC, SpecTemp \cite{segarra2017network} and the method by Kalofolias \cite{kalofolias2016learn}. To infer central nodes using the SC method, we first apply \eqref{eq:initial} to initialize the {\sf E}-step and perform only one iteration of the {\sf M}-step in \Cref{alg:em} to estimate the low rank and sparse matrices. On the other hand, a three-step procedure is simulated for \cite{segarra2017network, kalofolias2016learn}. We first cluster data into $\mulG$ groups with the graph identifiers from \Cref{alg:em}; then, we apply \cite{segarra2017network, kalofolias2016learn} on the individual data groups to learn the corresponding graphs and compute the eigen-centrality vectors subsequently. 

\begin{figure}[t]
\centering
{\sf 
\begin{subfigure}{.95\linewidth}
\centering
 \resizebox{.99\linewidth}{!}{\input{legend_batchEm}}\\[.3cm]
\resizebox{!}{.4\linewidth}{\input{BatchEM_wlf_vsC_errorrate}}
\resizebox{!}{.4\linewidth}{\input{BatchEM_slf_vsC_errorrate}}
\caption{\footnotesize Error rate of top-10 central nodes detection.}
\end{subfigure}
\begin{subfigure}{.95\linewidth}
\centering
{
\resizebox{!}{.4\linewidth}{\input{BatchEM_wlf_vsC_NMI}}
\resizebox{!}{.4\linewidth}{\input{BatchEM_slf_vsC_NMI}}
} 
\caption{\footnotesize Clustering accuracy.}
\end{subfigure}
\begin{subfigure}{.95\linewidth}
\centering
\resizebox{!}{.5\linewidth}{\input{BatchEM_objf}}
\resizebox{!}{.5\linewidth}{\input{BatchEM_slp_objf}}
\caption{\footnotesize MAP objective value for Alg.~\ref{alg:em} in one trial.}
\end{subfigure}

}
\caption{Performance of the batch EM algorithm [cf.~\Cref{alg:em}] against the number of graphs $\mulG$ under (Left) ${\cal H}_{\sf w}(\mA{w_t})$ and (Right) ${\cal H}_{\sf s}(\mA{w_t})$.}\label{fig:batchEM}\vspace{-.3cm}
\end{figure}
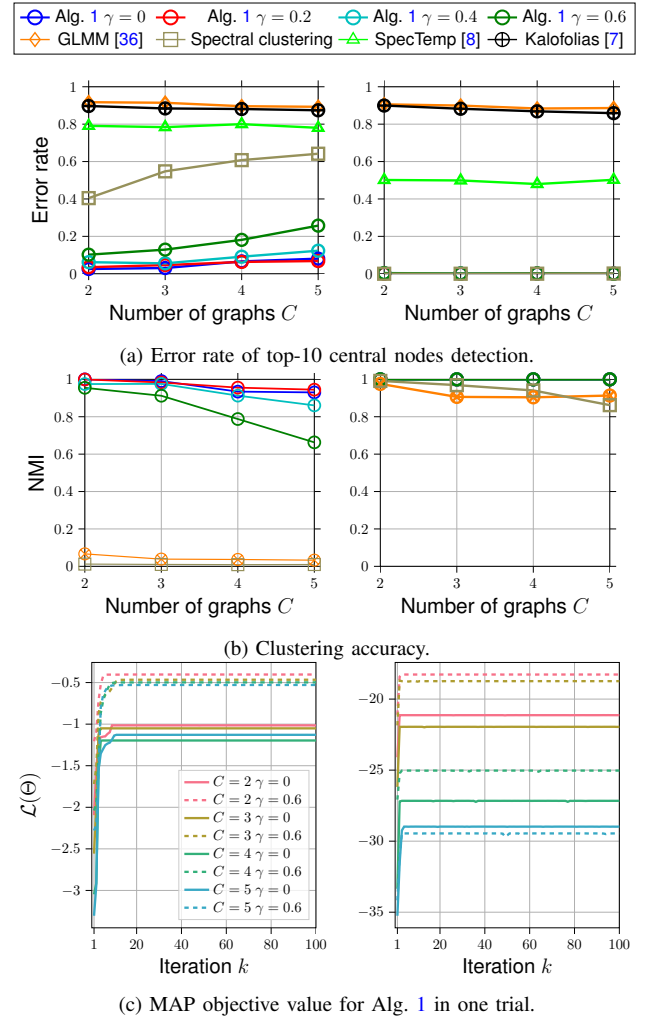

Fig.~\ref{fig:batchEM} compares the performance of algorithms against the number of graphs $\mulG$ with respect to the clustering accuracy (measured by NMI) and centrality detection error rate from 100 Monte-Carlo trials. First, observe that \Cref{alg:em} achieves significantly better performance than the benchmarks, even when a portion of observations are missing (with $\gamma > 0$). Second, under the weak low pass filter ${\cal H}_{\sf w}(\cdot)$ (Fig.~\ref{fig:batchEM}, Left), the performances of tested algorithms' worsen with the number of graphs $\mulG$; while the effect of $\mulG$ is less significant with the strong low pass filter ${\cal H}_{\sf s}(\cdot)$ (Fig.~\ref{fig:batchEM}, Right). 
Our results indicate that \Cref{alg:em} is robust to smoothness (i.e., low pass property) of graph signals and low-rank excitation. 
\vspace{.1cm}

\begin{figure}[t]
\centering
{\sf 
\begin{subfigure}{.95\linewidth}
\centering
\resizebox{!}{.39\linewidth}{\input{onlineEM_vsm_Errorrate}}
\resizebox{!}{.41\linewidth}{\input{onlineEM_vsm_Errorarate_slp}}
\caption{\footnotesize Error rate of top-10 central nodes detection.}
\end{subfigure}
\begin{subfigure}{.95\linewidth}
\centering
{
\resizebox{!}{.4\linewidth}{\input{onlineEM_vsm_NMI}}
\resizebox{!}{.4\linewidth}{\input{onlineEM_vsm_NMI_slp}}
} 
\caption{\footnotesize Clustering accuracy.}
\end{subfigure}
\begin{subfigure}{.95\linewidth}
\centering
\resizebox{!}{.52\linewidth}{\input{onlineEM_vsm_Fvalue}}
\resizebox{!}{.52\linewidth}{\input{onlineEM_vsm_Fvalue_slp}}
\caption{\footnotesize MAP objective value for Alg.~\ref{alg:olem} in one trial.}
\end{subfigure}
}
\caption{Performance of the online EM algorithm [cf.~\Cref{alg:olem}] against time $t$. (Left) ${\cal H}_{\sf w}(\mA{w_t})$ and (Right) ${\cal H}_{\sf s}(\mA{w_t})$. }\label{fig:onlineEM}\vspace{-.2cm}
\end{figure}
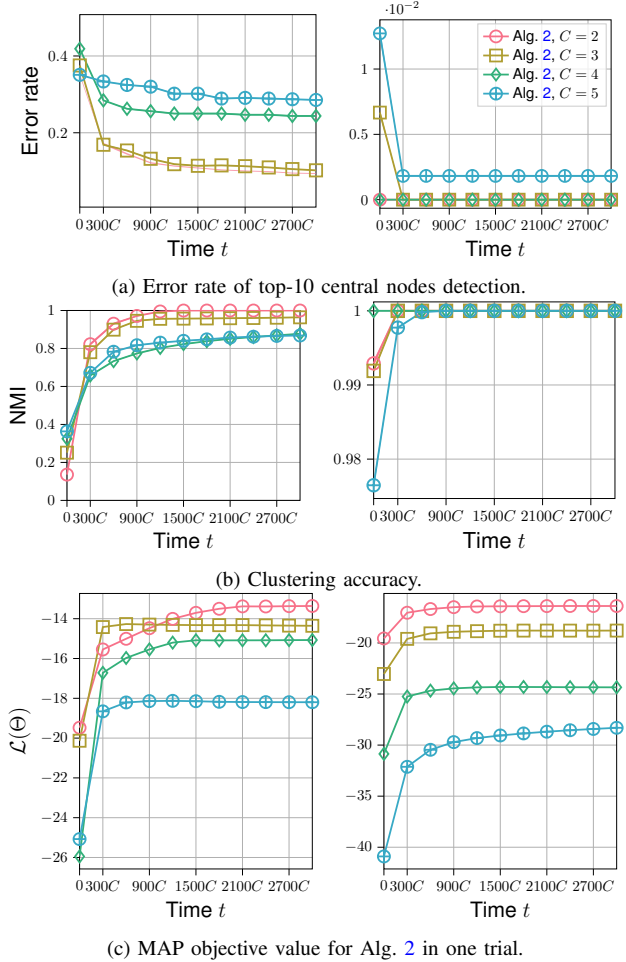

\noindent \textbf{Online Algorithm.}~~The next experiment considers the streaming data setting where at time $t$, only the $t$th signal tuple ${\sf DP}_t$ is available and we focus on applying \Cref{alg:olem} to continuously estimate centrality of graphs and cluster the graph signals. The data tuples are generated from \eqref{eq:mixmodel} in the same way as in the batch data setting with no missing data, i.e., $\gamma = 0$, and we simulate the same pair of strong (${\cal H}_{\sf s}(\cdot)$) and weak (${\cal H}_{\sf w}(\cdot)$) graph filters. Note that we maintain the full dataset with $m=3000\mulG$ samples for benchmark. For \Cref{alg:olem}, we set $\lambda_L = 0.01, \lambda_S = 0.001$ and initialize the algorithm through applying \Cref{alg:em} on $m_{\sf init} = 50 \mulG$ signal tuples. The step size is selected as $\beta_t = \frac{0.5}{t+m_{\sf init}}$.
We are not aware of existing works which perform simultaneous graph learning and graph signal clustering. 

Fig.~\ref{fig:onlineEM} plots the trajectories of clustering accuracy through evaluating NMI on the \emph{full} dataset, centrality detection error rate, and the MAP objective value evaluated over the full dataset with 10 Monte-Carlo runs with the 90\% confidence intervals. 
As observed, the performance of \Cref{alg:olem} improves with time as the algorithm obtains more samples, which allows the algorithm to construct better estimate to the sufficient statistics in \eqref{eq:SAsuffstat} via the SA scheme.  
Comparing between Fig.~\ref{fig:onlineEM} (Left) and (Right), the terminal performance is affected by the strength of low pass graph filter as well as the model order, i.e., number of candidate graphs. The latter observation is similar to that in the batch data setting.  
\vspace{.1cm}

\noindent \textbf{Logit model.}~~Before concluding this subsection, let us also consider an application of \Cref{alg:em} to the logit model with batch data; cf.~Sec.~\ref{sec:logit}. We consider a set of $\mulG$ graphs built on the simple \emph{star graph} with $n=20$ nodes, each with a different central node, and additional edges are assigned with probability $0.02$ between the non-central nodes. 
In each of 30 Monte-Carlo trials, we generate $m = 80\mulG$ data tuples according to \eqref{eq:logitmodel}. The excitation matrix ${\bm B}$ follows a similar generation process as before but with $\EE[ M_{ij} ] = 0.3$. The excitation rank is $r=16$ and the tested graph filter is ${\gfilter} (\mA{w_t}) = ( {\bm I} - \frac{1}{30} \mA{w_t} )^{-1}$. 
For the logit model, we set the bias parameter $b$ as the negative average value of all signals.
Lastly, \Cref{alg:em} is implemented in MATLAB with the CVX package \cite{cvx} for solving \eqref{eq:logitsurrogate}. Note the benchmark algorithms are implemented through directly treating the binary observations as real-valued graph signals.

Table \ref{table:logittest} compares \Cref{alg:em} with benchmark algorithms on tackling \Cref{prob:jg} in terms of the clustering accuracy (NMI) and error rate in detecting the central node of each graph. We observe that \Cref{alg:em} can accurately separate the observations into $\mulG$ groups and detect the most central nodes inside $\mulG$ graphs while the benchmarks have  failed in almost all 30 trials under the logit model.\vspace{-.3cm}

\begin{table}[t]
    \setlength{\tabcolsep}{2.5pt}
    \centering
\begin{tabular}{l l l l l l}
    \toprule 
    \bfseries ~ & Alg.~\ref{alg:em} & SC & GLMM\cite{maretic2020graph} & SpecTemp\cite{segarra2017network} & Kalofolias\cite{kalofolias2016learn}\\ \midrule
     {\bf Error rate} & \cellcolor{black!7} 0.15 & 0.73 & 1.00 & 0.71 & 0.91 \\
     {\bf NMI} & \cellcolor{black!7} 0.93 & 0.18 & $10^{-3}$ & {\tt N/A} & {\tt N/A} \\
    \bottomrule
    \end{tabular}
    \caption{Tackling \Cref{prob:jg} with Logit (binary) observations. The number of graphs is set as $\mulG=2$.}\vspace{.0cm}
    \label{table:logittest} 
\end{table}


\subsection{Application: Anomaly Detection}\vspace{-.2cm}
This section considers applying \Cref{alg:olem} to online anomaly graph detection application as described in Sec.~\ref{sec:anom}.
In the following simulation results, we consider two groups of graphs with the same size $n=100$ and the tested graph filter is ${\cal H}_{\sf w}(\mA{w_t})$ from Sec.~\ref{sec:synexp}. The first group generates \emph{normal graph signals} from $\mulG=2$ different {\sf CP} graphs $\{ \mG{\mulg} \}_{\mulg=1}^\mulG$. The second group generates \emph{abnormal graph signals} via a {\sf CP} graph $\mG{\mulG+1}$ with a different core nodes set ${\cal V}_o^{(\mulG+1)}$ from $\{ {\cal V}_o^{(\mulg)} \}_{\mulg=1}^\mulG$. In our simulation, the abnormal graph signals are observed in two modes, either briefly in order or randomly. To initialize \Cref{alg:olem}, we use a set of $m_{\sf init}$ normal graph signals with the batch \Cref{alg:em}. 

\begin{figure}[t]
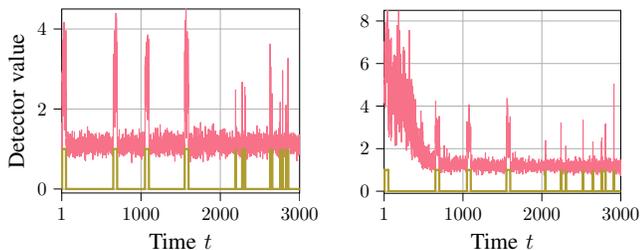

\centering
{
\resizebox{!}{.37\linewidth}{\input{online_randomdetector}}
\resizebox{!}{.37\linewidth}{\input{online_poordetector}}
} 
\caption{Anomaly detection under ${\cal H}_{\sf w}(\mA{w_t})$ with (Left) $m_{\sf init} = 200\mulG$, and (Right) $m_{\sf init} = 20\mulG$. Red line is the detector value of \eqref{eq:detector_p} against time $t$. Yellow line indicates when the abnormal signal is observed. If its value is $1$, an abnormal signal is observed.
} \label{fig:abdetector}\vspace{-.3cm}
\end{figure}

We first compare the detector value of \eqref{eq:detector_p} against time $t$ in Fig.~\ref{fig:abdetector}. The left panel considers the case with $m_{\sf init} = 200\mulG$ where \Cref{alg:olem} is initialized with a large batch of normal signals; while the right panel considers the case with $m_{\sf init} = 20 \mulG$ where the initialization for \Cref{alg:olem} can be inaccurate. 
With a large batch initialization (left panel), we observe that the detector value \eqref{eq:detector_p} records a significant spike over the time intervals with abnormal signals. With small batch initialization (right panel), the detector is less sensitive to the abnormal signals at the beginning. However, as time goes by and \Cref{alg:olem} processes enough samples, \eqref{eq:detector_p} produces clear spikes when an abnormal graph signal is recorded.

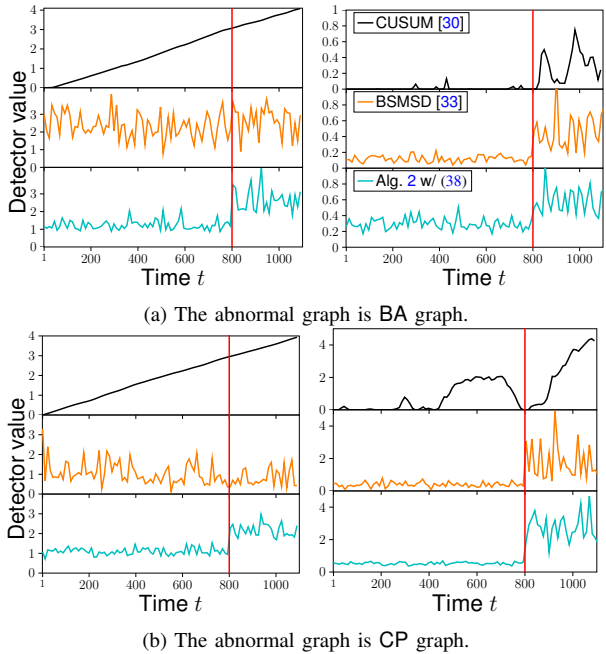
\begin{figure}[t]
\centering
{\sf 
\begin{subfigure}{\linewidth}
\centering
\resizebox{!}{.43\linewidth}{\input{onlineEMdetector_baw}}
\resizebox{!}{.43\linewidth}{\input{onlineEMdetector_babl}}
\caption{\footnotesize The abnormal graph is {\sf BA} graph.}
\end{subfigure}
\begin{subfigure}{\linewidth}
\centering
{
\resizebox{!}{.43\linewidth}{\input{onlineEMdetector_cpw}}
\resizebox{!}{.43\linewidth}{\input{onlineEMdetector_cpbl}}
} 
\caption{\footnotesize The abnormal graph is {\sf CP} graph.}
\end{subfigure}}
\caption{Change point detection under (Left) $ {\gfilter}_{\sf w} (\mA{w_t})$ and  (Right)   ${\gfilter}^{'}_{\sf bl} (\mA{w_t}) $. The detector values are normalized for better illustration. The time intervals $\{1, \ldots, 400\}, \{401, \ldots, 800\}$ refer to a 'normal region' respectively with graph $G^{(1)}, G^{(2)}$, and $\{ 801, \ldots 1100 \}$ refer to a 'abnormal region' such that $t=800$ is the change point. 
} \label{fig:cpdetector}\vspace{-.3cm}
\end{figure}

We next study the change point detection problem which is a special case of anomaly graph detection. Our aim is to detect the time instance when the underlying graph is switched to an abnormal one. For instance, the problem is relevant for detecting events such as transmission line failure in power systems.
We consider two types of abnormal graphs: {\sf CP} graphs with different sets of central nodes as described previously, Barabasi–Albert ({\sf BA}) graphs where each added node is connected to random $n_{\sf BA} = 10$ existing nodes with probability proportional to their degrees. The first 800 samples are generated from normal graphs, while the remaining 300 samples are generated from abnormal graph. 

We compare the detector values \eqref{eq:detector_p} against $t$ with  CUSUM\cite{kaushik2021network}, BSMSD \cite{isufi2018blind}. Both existing algorithms consider the case with only one normal graph. To extend them into multiple graph settings, for the observed signal ${\bm \op}_t$ we compute the CUSUM/BSMSD detector values based on each normal graph $\mG{\mulg}$ whose topology are assumed known, then we take the minimum of the $\mulG$ detector values. We expect a spike in the combined detector value at the change point, i.e., the time instance when ${\bm \op}_t$ is generated from the abnormal graph.
We consider two graph filter designs in our experiment. The first one in Fig.~\ref{fig:cpdetector} (Left) takes the weak low pass filter ${\gfilter}_{\sf w} (\cdot) $ defined previously. The second one in Fig.~\ref{fig:cpdetector} (Right) adopts the low pass filter defined in \cite{isufi2018blind}. In the latter case, we set ${\bm B} = {\bm I}$ and the latent parameters are generated as ${\bm \latp}_t \sim {\cal N}( {\bm 0}, {\bm I} )$. The eigenvalues of the tested graph filter 
${\gfilter}^{'}_{\sf bl} (\cdot)$ 
is ${\bm \Sigma}^{(w_t)}$ with ${\Sigma}^{(w_t)}_{ii} = \exp(-i/10),~i=1,\dots,n$. 

Fig.~\ref{fig:cpdetector} presents the detector values against time $t$ to compare the performance of three tested detectors. 
We observe that all detectors are able to detect the change point in the setting with graph filter ${\gfilter}^{'}_{\sf bl} (\cdot)$, as seen from the pronounced spikes in detector values. \Cref{alg:olem} has a comparable sensitivity to existing works despite the algorithm does not know the graph topology a-priori. 
On the other hand, under the weak low pass filter ${\cal H}_{\sf w}(\cdot)$, \Cref{alg:olem} still provides reliable detection on the abnormal signals. The other two detectors do not show any detectable pattern in 'normal region' and 'abnormal region'.\vspace{-.2cm} 



\subsection{Experiments on Real Data} \label{sec:realexp}\vspace{-.2cm}
In this subsection, we apply the proposed algorithms on two datasets of graph signals. The first dataset ({\sf Stock}) is the daily returns of S\&P 100 stocks in May 2018 to Aug 2019 with $n=99$ stocks and $m=300$ samples, collected from \url{https://www.alphavantage.co/}\footnote{The {\sf Stock} dataset is pre-processed by subtracting the daily returns by the minimum return value across all samples. Note that the transformed daily return values are non-negative.}. To estimate the excitation parameters ${\bm \latps}$, we model the latter as the state of the world and consider the \emph{interest levels} over time of $k=5$ keywords 'trade war', 'sales tax', 'Iran', 'oil crisis', 'election' obtained from Google Trend (\url{https://trends.google.com}). As the stock graph may be time varying, our aim is to cluster the $m=300$ graph signals into $\mulG=2$ groups and detect central stocks on the graph associated with each group of signals. We use \Cref{alg:em} with $\lambda_L = 1.8 \times 10^{-3}$, $\lambda_S = 1.09 \times 10^{-4}$.

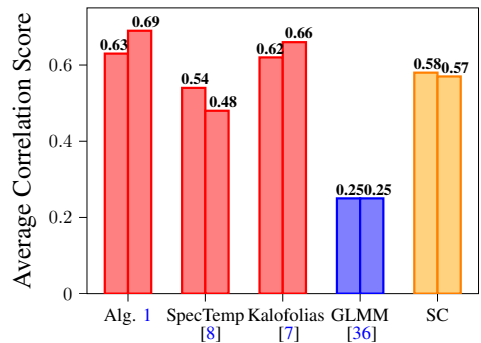
\begin{figure}[t]
    \centering
    \resizebox{!}{.53\linewidth}{\input{stock}}
    \captionof{figure}{Average correlation scores with the top-10 detected central stocks (in two graphs) for {\sf Stocks} dataset. {\red Red}/{\blue Blue}/{\orange Orange} color indicates that ${\bm \ops}^{(\mulg)}$ are found by first clustering with \Cref{alg:em}/GLMM\cite{maretic2020graph}/SC, then the central nodes are detected with respective algorithms.}
    \label{fig:stock}
  \end{figure}

For the $\mulg$th clustered group of samples, 
denote ${\bm g}^{(\mulg)}$ as the corresponding S\&P100 index and ${\bm \ops}^{(\mulg)}$ as the individual stocks' daily returns. To measure the quality of central stocks detected, we evaluate the \emph{normalized correlation} between the stock's daily returns and S\&P100 index through
\beq\label{eq:correlation}
 {\sf corr}^{(\mulg)}_i = || {\bm \ops}^{(\mulg)}_{i,:} ||^{-1} || {\bm g}^{(\mulg)} ||^{-1} \langle {\bm \ops}^{(\mulg)}_{i,:}, {\bm g}^{(\mulg)} \rangle \in [0,1],
\eeq
where $i = 1,\dots,n$. A higher correlation score indicates the corresponding stock $i$ is a better representative of all stocks, which may imply a more central node.
Fig.~\ref{fig:stock} shows the average correlation scores of top-10 detected central stocks from the corresponding groups of samples clustered by the tested algorithms.
Observe that \Cref{alg:em} delivers higher average correlation scores than other algorithms.  
We also observe from Table~\ref{table:stock} that \Cref{alg:em} detects two different groups of central stocks while other algorithms detect pairs of graphs with repeated groups of central stocks.  

The second dataset ({\sf Brain}) collects the functional magnetic resonance imaging (fMRI) data of 50 subjects from the Human Connectome Project with $m = 240,000$ samples, where the subjects were in resting state (RS). We use the preprocessed RS fMRI data by \cite{ricchi2022dynamics} and consider the Automated Anatomical Labeling Atlas 90 \cite{tzourio2002automated} for labels of brain regions. The $k=5$ regions with largest summed absolute values of all samples are selected to form the excitation parameters ${\bm \latps}$. They are left \& right medial superior frontal gyrus (label 23, 24), left \& right cuneus (label 45, 46) and right inferior occipital gyrus (label 54). The remaining $n=85$ brain regions are regarded as nodes in the unknown brain graphs. To initialize \Cref{alg:olem}, we take $m_{\sf init} = 4800$ randomly selected samples and apply the batch algorithm.
We set the step size as $\beta_t = \frac{1}{t+m_{\sf init}}$, and the MAP problem parameters are $\lambda_L = 0.45$, $\lambda_S = 0.045$.

\begin{figure*}[t]
\begin{center}
    \includegraphics[width=.875\linewidth]{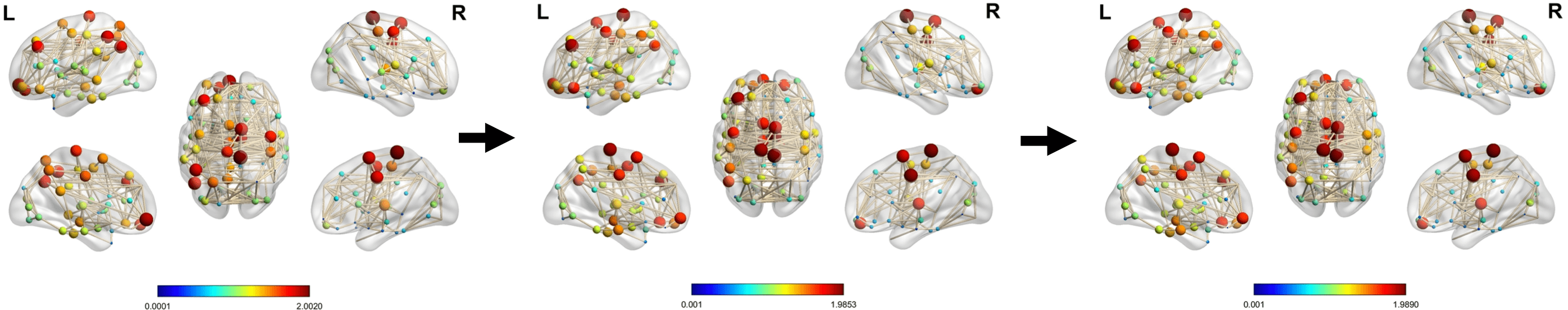}\\
    \includegraphics[width=.875\linewidth]{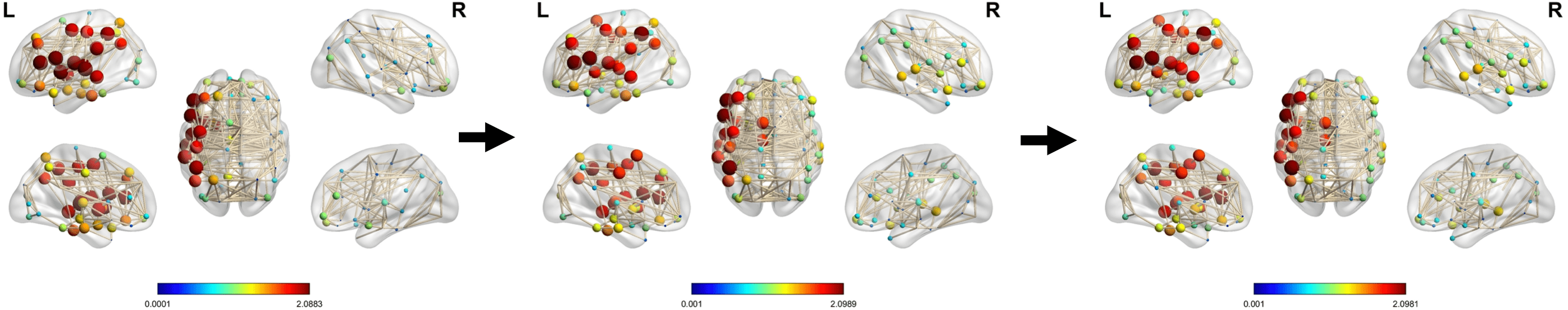}\\
    \includegraphics[width=.875\linewidth]{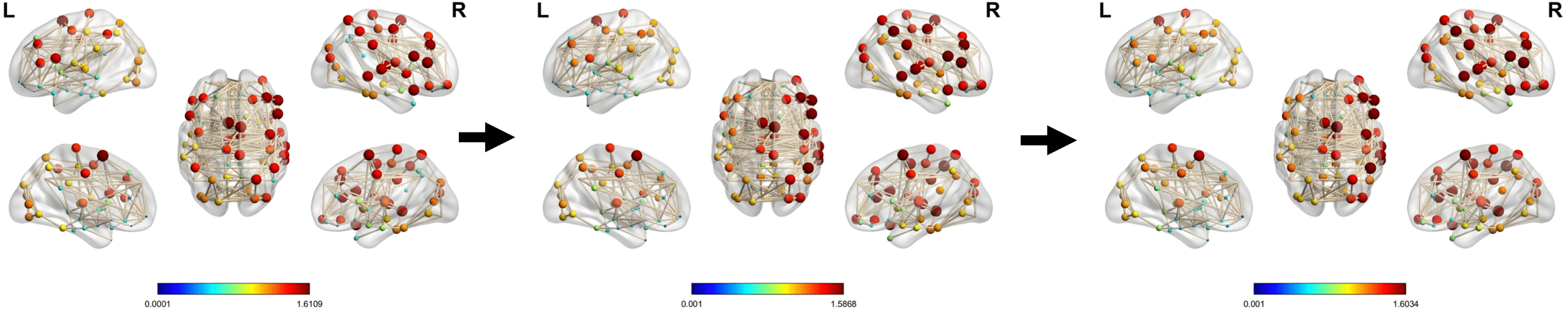}    
\end{center}
\caption{Estimated centrality vectors by \Cref{alg:olem} at time $t=4800/134,160/240,000$ (Left/Middle/Right). (Top) $\mG{1}$, (Middle) $\mG{2}$ and (Bottom) $\mG{3}$. More central brain regions are presented with darker color and larger node size. 
At time $t=240,000$, $P_1 = 0.3582$, $P_2 = 0.1562$ and $P_3 = 0.4856$.
} \label{fig:brain}\vspace{-.2cm}
\end{figure*}

The brain connectivity graph can vary for different resting state brain networks \cite{ricchi2022dynamics}.
Fig.~\ref{fig:brain} presents the estimated centrality of $\mulG=3$ graphs against time while running \Cref{alg:olem}. The output of the algorithm stabilizes as more samples are observed, indicating a consistent estimation. Furthermore, the graphs $\mG{2}$ and $\mG{3}$ show central nodes that are concentrated in the left brain and right brain, respectively. Table~\ref{table:brain} lists the labels of estimated central brain regions. For $\mG{1}$, label 70 is Paracentral lobule \cite{grotta2021stroke}. It controls the movement and sensation in the lower body. For $\mG{2}$ and $\mG{3}$, label 12 and label 13 belong to the inferior frontal gyrus region \cite{greenlee2007functional}, which is associated with speech and language processing.\vspace{-.2cm} 

\section{Conclusions}\vspace{-.2cm}
We study a joint graph inference problem on the challenging mixture model of filtered graph signals under general (non-white) excitation, weak low pass graph filters and missing data. We design an efficient algorithm based on EM and develop the latter's online extension for streaming data. The online algorithm is further applied to abnormal graph signals detection. Efficacy of the proposed algorithms are verified with convergence analysis and numerical experiments. \vspace{-.2cm}

\begin{table}[t]
    \setlength{\tabcolsep}{2.5pt}
    \centering
\begin{tabular}{l l l l l l}
    \toprule 
    \bfseries ~ & Alg.~\ref{alg:em}  & SpecTemp\cite{segarra2017network} & Kalofolias\cite{kalofolias2016learn} & GLMM\cite{maretic2020graph} & SC\\ \midrule
     {\bf Max} & \cellcolor{black!7} 0.81(V) & 0.76(ACN) & 0.76(ACN)  & 0.4 (MDLZ) & 0.78(ACN)\\
     {\bf Min} & \cellcolor{black!7} 0.34(KO) & 0.16(SO) & 0.34(KO) & 0.07(DUK) & 0.35(MCD) \\
     {\bf Rep.$^\dagger$} & \cellcolor{black!7} 0 & 0.2 & 0.5 & 1 & 0.7  \\
    \bottomrule
    \end{tabular}
    \footnotesize{$^\dagger$Fraction of repeated central stocks detected $\frac{1}{10}| \widehat{\cal V}_o^{(1)} \cap \widehat{\cal V}_o^{(2)} |$.
    }
    \caption{Correlation score of top-10 detected central stocks of two graphs for {\sf Stocks} dataset.}\vspace{-.2cm}
    \label{table:stock} 
\end{table}

    \begin{table}[t]
    \centering
     \begin{tabular}{l l l l l l l l l l l}
    \toprule
    \bfseries  & \multicolumn{10}{c}{\bfseries Brain Regions with Top-10 estimated centrality value}\\ \midrule
    $\mG{1}$ & 70 & 34 & 69  & 20 & 7 & 1 & 19 & 15 & 78 & 6\\
    \hline
     $\mG{2}$ & 13 & 61 & 11 & 17 & 7 & 85 & 63 & 81 & 1 & 33  \\
    \hline 
    $\mG{3}$ & 12 & 62 & 14 & 20 & 8 & 86 & 82 & 64 & 2 & 19\\
    \bottomrule
    \end{tabular}
    \caption{Labels of estimated central brain regions sorted left to right for {\sf Brain} dataset.}\vspace{-.2cm}
    \label{table:brain}
    \end{table}

\appendices 
\section{Proof of Proposition~\ref{prop:conv}}\label{app:EM}\vspace{-.2cm}
The first part follows directly from the steps of EM algorithm as the latter are majorization-minimization iterations for the regularized log-likelihood. For any $k \geq 0$,
\beq \textstyle
{\cal L}( {\Theta}^{k+1} ) -{\cal L} ( {\Theta^k)} \geq \widetilde{\mathcal{L}}(\Theta^{k+1} \, | \, {\Theta}^k ) - \widetilde{\mathcal{L}}(\Theta^k \, | \, {{\Theta}}^k  ) \geq 0,
\eeq
where the last inequality holds since $\Theta^{k+1}$ is a maximizer to the surrogate $\widetilde{\cal L}( \Theta^{k+1} | \Theta^k )$ \eqref{eq:surrogate} and $\widetilde{\cal L}( \Theta^k | \Theta^k ) = {\cal L}( \Theta^k )$. 

For the second part of the proposition, we observe that as
\beq \notag
{\tt D}( \Theta^k | \Theta^{k-1} ) = {\cal L} ( \Theta^k ) - \widetilde{\cal L}( \Theta^k | \Theta^{k-1} ) \leq {\cal L} ( \Theta^k ) - {\cal L}( \Theta^{k-1} ),
\eeq 
where the inequality follows from $\widetilde{\cal L}( \Theta^k | \Theta^{k-1} ) \geq \widetilde{\cal L}( \Theta^{k-1} | \Theta^{k-1} ) = {\cal L}( \Theta^{k-1} )$.
Since the MAP problem has bounded objective value ${\cal L}( \Theta ) \leq {\cal L}^\star$. It holds
\beq \textstyle \label{eq:telescope}
\sum_{k=1}^{K_{\sf max}} {\tt D}( \Theta^k | \Theta^{k-1} ) \leq {\cal L}^\star - {\cal L}(\Theta^0),
\eeq 
for any $K_{\sf max} \geq 1$.
Now, as the gradient w.r.t.~$\Theta$ of ${\tt D}( \Theta | \widetilde{\Theta} )$ is $L$-Lipschitz, it implies
\beq 
\begin{split}
0 & \leq {\tt D} \left( \Theta^k - {\textstyle \frac{1}{L}} \grd {\tt D} ( \Theta^k | \Theta^{k-1} ) \,|\, \Theta^{k-1} \right) \\
& \leq {\tt D}( \Theta^{k} | \Theta^{k-1} ) - \frac{1}{2L} \| \grd {\tt D} ( \Theta^k | \Theta^{k-1} ) \|^2, \\
& \Longrightarrow \frac{1}{2L} \| \grd {\tt D} ( \Theta^k | \Theta^{k-1} ) \|^2 \leq {\tt D}( \Theta^{k} | \Theta^{k-1} ).
\end{split}
\eeq
Summing up both sides from $k=1$ to $k= K_{\sf max}$ and using \eqref{eq:telescope} yields
\beq \label{eq:grdD}
\min_{ k=1,\ldots, K_{\sf max} } 
\| \grd {\tt D} ( \Theta^k | \Theta^{k-1} ) \|^2 \leq 
\frac{2L}{K_{\sf max}} \big({\cal L}^\star - {\cal L}(\Theta^0)\big).
\eeq 
Furthermore, the directional derivative satisfies
\beq \notag
\begin{split}
& {\cal L}'( \Theta^k ; \Theta - \Theta^k ) \\
& = \pscal{ \grd {\tt D}( \Theta^k | \Theta^{k-1} ) }{ \Theta - \Theta^k } + \widetilde{\cal L}' ( \Theta^k ; \Theta - \Theta^k | \Theta^{k-1} ).
\end{split}
\eeq 
As $\Theta^k$ maximizes the concave function $\widetilde{\cal L}( \Theta | \Theta^{k-1} )$, it holds
\beq 
\widetilde{\cal L}' ( \Theta^k ;  \Theta - \Theta^k | \Theta^{k-1} ) \leq 0,~\forall~\Theta \in \mathfrak{T}.
\eeq 
By Cauchy-Schwarz inequality, this implies that 
\beq 
\sup_{ \Theta \in \mathfrak{T}} \frac{ {\cal L}'( \Theta^k ; \Theta - \Theta^k ) }{ \| \Theta - \Theta^k \| } \leq \| \grd {\tt D}( \Theta^k | \Theta^{k-1} ) \|,
\eeq 
where $\frac{0}{0} = 0$.
Combining with \eqref{eq:grdD} leads to the conclusion.\vspace{-.2cm}

\section{Proof of \Cref{prop:fixedpt}} \label{app:fixedpt}\vspace{-.1cm}
To simplify notations in this proof, we denote
\[ 
\widetilde{\cal L}_{\sf ol}( \Theta; {\bm S} ) = \Phi( \Theta; {\bm S} ) - \Psi(\Theta), 
\]
where ${\bm S} := ( {P}_\mulg, {{\bm \ops}{\bm \latps}}_\mulg, {{\bm \latps}{\bm \latps}}_{\mulg,i} )_{\mulg, i}$ collects the sufficient statistics, and the
non-smooth function $\Psi(\Theta)$ is the regularizer
and $\Phi( \Theta; {\bm S} )$ collects the remaining terms as found in \eqref{eq:surrogate_2stoc}. 

We denote ${\bm S} (\Theta)$ as the population sufficient statistics computed from ${\Theta}$ through \eqref{eq:suffstat_stoc}. Furthermore, $\Theta( {\bm S} ) \in \argmax_{ \Theta } \widetilde{\cal L}_{\sf ol}( \Theta; {\bm S} )$ denotes a maximizer to the surrogate function given the sufficient statistics. Define the smooth function $f(\Theta) := \EE_{\sf DP}[ \log p ( {\bm Y} | \Theta, {\bm Z}, \bm{\Omega} ) ]$ [cf.~\eqref{eq:MAPa}] and  
\beq 
\Gamma = \{ \Theta : 0 \in \grd f( \Theta) - \partial \Psi( \Theta ) \}
\eeq 
is the set of stationary points to the MAP problem \eqref{eq:MAPa} with the modified regularizer. 

If $\overline{\bm S}$ satisfies the fixed point condition \eqref{eq:satpoint} where $\overline{\bm S} = {\bm S} (\overline{\Theta})$, we have $\overline{\Theta} \in \argmax_{\Theta} \widetilde{\cal L}_{\sf ol}( \Theta; \overline{\bm S} )$ such that
\beq \label{eq:fixedpt_cond}
  0 \in \partial \widetilde{\cal L}_{\sf ol}( \overline{\Theta} ; \overline{\bm S} ) = \grd \Phi( \overline{\Theta} ; \overline{\bm S} ) - \partial \Psi(\overline\Theta).
\eeq
By the Jensen's inequality and the fact $f( \overline\Theta ) - \Phi(\overline{\Theta}; \overline{\bm S} ) = {\tt c}'$ for some constant ${\tt c}'$, there exists ${\tt c} > -\infty$ such that
\beq \label{eq:jensen_sat}
f( \Theta ) - f( \overline\Theta ) - \big( \Phi ( {\Theta} ; \overline{\bm S} ) - \Phi ( \overline{\Theta} ; \overline{\bm S} ) \big) \geq {\tt c},~\forall~\Theta,
\eeq 
and the lower bound is achieved when $\Theta = \overline{\Theta}$. It implies
\beq \label{eq:map_equiv}
0 = \grd f( \overline\Theta ) - \grd \Phi( \overline{\Theta} ; \overline{\bm S} ).
\eeq 
Inserting \eqref{eq:map_equiv} into the fixed point condition \eqref{eq:fixedpt_cond} yields
$\overline{\Theta} \in \Gamma$.

On the other hand, let $\overline{\Theta} \in \Gamma$ and observe the inequality \eqref{eq:jensen_sat} and its derived condition \eqref{eq:map_equiv}. Thus,
\beq 
0 \in \grd \Phi( \overline{\Theta} ; \overline{\bm S} ) - \partial \Psi( \overline{\Theta} ).
\eeq 
As the maximizer to $\widetilde{\cal L}_{\sf ol}( \Theta; {\bm S} )$ is unique for any ${\bm S}$, $\Theta( \overline{\bm S})$ is well defined and we conclude that $\overline{\Theta} = \Theta( \overline{\bm S})$.  
Furthermore, it holds that $\overline{\bm S} = {\bm S}( \overline{\Theta} )$, which is a fixed point to \eqref{eq:satpoint}.

\bibliographystyle{IEEEtran}
\bibliography{ref_list}

\end{document}

%% file: Mixture_slp.tex
\begin{tikzpicture}

\begin{axis}[
tick align=outside,
tick pos=left,
x grid style={white!69.0196078431373!black},
xmin=0.015, xmax=0.06,
xtick style={color=black},
y grid style={white!69.0196078431373!black},
ymin=-0.08, ymax=0.08,
ytick style={color=black},
width=6.5cm,height=5.7cm,
]
\addplot [thick, red, mark=o, mark size=3, mark options={solid}, only marks]
table {%
0.034821707	0.047489109
0.03225297	0.010588904
0.035683616	0.031744901
0.034928955	0.049869786
0.034825726	0.026985418
0.039586757	0.014823244
0.039204706	0.03301419
0.037728454	0.049364658
0.040546794	0.041699258
0.040422241	0.037117938
0.03813386	0.042513989
0.028475545	0.024813498
0.042620769	0.050644492
0.026500178	0.007250748
0.039836893	0.007266785
0.038384588	0.042248716
0.029027164	0.041733112
0.028660214	0.019905641
0.022738901	0.028100154
0.029638392	0.013173848
0.046788547	0.046586695
0.032867549	0.039333659
0.041097372	0.034539625
0.032435128	0.043579441
0.036365396	0.009944049
0.0395647	0.037780167
0.037588211	0.029021462
0.030676457	0.022602601
0.044343923	0.020469399
0.038324436	0.024765112
0.028088246	0.019913687
0.028426392	0.023829725
0.043720992	0.043967594
0.035618122	0.03121862
0.041429706	0.050187316
0.045196629	0.047783356
0.0356105	0.037423567
0.03708666	0.03373542
0.032457757	0.038766266
0.043341557	0.028712328
0.043363023	0.047373065
0.033329042	0.017088675
0.023844284	0.025664083
0.033303379	0.033264091
0.039728706	0.043743019
0.032182059	0.023859443
0.037414569	0.049308236
0.034098202	0.041681707
0.03577919	0.049758508
0.030714713	0.045316871
0.037522972	0.031505051
0.032180035	0.03289224
0.040406862	0.030270339
0.041861141	0.025048737
0.038002776	0.039581613
0.019887965	0.014154531
0.029588686	0.037388944
0.042191468	0.037408397
0.028135253	0.026599186
0.032683956	0.025508779
0.027934544	0.032018352
0.021169299	0.007515639
0.034099375	0.018370781
0.033447168	0.020170513
0.027780156	0.024493835
0.039636025	0.02113232
0.030821912	0.050192191
0.036366868	0.034514318
0.025590479	0.022188479
0.037351428	0.043215619
0.02841503	0.022801398
0.034056442	0.011642783
0.03330654	0.011532929
0.03731762	0.026444956
0.043159022	0.043885368
0.033547254	0.042027059
0.036248239	0.031921963
0.024027602	0.006306972
0.023429952	0.01667288
0.032526829	0.029628605
0.030304037	0.03734016
0.03148085	0.012784833
0.036797105	0.032613881
0.039594499	0.037834302
0.032293934	0.0443188
0.028148741	0.026751483
0.035299585	0.041223724
0.03963308	0.023697756
0.046709488	0.056229601
0.034196702	0.019341735
0.030373529	0.001955169
0.027200643	0.027543482
0.034137257	0.034347827
0.035022989	0.039614724
0.046601916	0.042049106
0.032290664	0.051482034
0.02617635	0.029100467
0.036996955	0.033897173
0.037511522	0.027090187
0.037519327	0.018457793
0.048729011	0.046112574
0.037059796	0.043220047
0.046762639	0.047682809
0.039578002	0.038041574
0.026970682	0.030226755
0.0436032	0.045815755
0.044231248	0.045898879
0.028485838	0.037401942
0.03620884	0.018833105
0.031839431	0.018896567
0.038857183	0.030586691
0.045435477	0.057791307
0.031372581	0.029715658
0.037517987	0.046760241
0.039660316	0.027558017
0.036834475	0.032300482
0.037063021	0.024026419
0.024859838	0.034975663
0.032737078	0.017518912
0.027506023	0.045448326
0.037336001	0.018026412
0.030932689	0.034843672
0.038897384	0.038123036
0.03335641	0.044594787
0.032156803	0.038244027
0.044630184	0.028888835
0.033586649	0.01073636
0.026494282	0.019558357
0.031747022	0.028031184
0.031631067	0.036889061
0.040855749	0.059079184
0.037417711	0.028629505
0.027329203	0.010131992
0.028495105	0.036260547
0.029078107	0.023682231
0.027961879	0.016366316
0.031351324	0.028032876
0.039433663	0.040188324
0.044852583	0.039621436
0.041827014	0.020949849
0.033305457	0.029076224
0.03079621	0.028018162
0.04385166	0.045049761
0.031334639	0.02198854
0.036632337	0.042732822
0.02714151	0.021352701
0.0348638	0.026431402
0.035419548	0.027740484
0.041170774	0.053638106
0.036714879	0.032675445
0.0309532	0.029001137
0.045058604	0.042045826
0.036298818	0.024028509
0.040522041	0.038538987
0.030065092	0.033740942
0.046255777	0.048741142
0.036086828	0.021970342
0.02901053	0.009790691
0.042213546	0.041317716
0.035804865	0.017104424
0.030190717	0.028699185
0.027526432	0.012719039
0.036812344	0.021450279
0.045545934	0.056779288
0.035290256	0.042506943
0.033503961	0.03151428
0.037701406	0.028098776
0.035844487	0.025329983
0.038681902	0.024785421
0.042191438	0.018337318
0.038142749	0.025636001
0.029017597	0.024183196
0.042541394	0.062939772
0.032430096	0.034561604
0.040418557	0.044845287
0.034006783	0.04051688
0.034900783	0.029345026
0.032874169	0.037059546
0.034703996	0.019805425
0.035511223	0.023958991
0.028896989	0.03074772
0.038189456	0.047504571
0.036534654	0.048966402
0.049753295	0.029128386
0.03718371	0.055421834
0.04130687	0.036822064
0.024553996	0.007221379
0.031772983	0.023762114
0.042025937	0.045677614
0.030241683	0.011883934
0.042985846	0.030534844
0.03350306	0.033867496
0.041054221	0.040480875
0.032656437	0.028348455
0.037269327	0.033038647
0.038329299	0.021714646
0.037891072	0.041121932
0.035066047	0.027756891
0.041877231	0.032491926
0.044384331	0.048941606
0.034628732	0.030144252
0.024033362	0.034196512
0.020677927	0.013208635
0.032828985	0.027441792
0.04065988	0.033041224
0.030328053	0.026630479
0.036027825	0.039722198
0.038544478	0.040340201
0.045659926	0.042619288
0.023210779	0.018805577
0.029085571	0.019589719
0.032209769	0.002525895
0.037588371	0.02092765
0.032403933	0.015578799
0.047456363	0.068674689
0.040618365	0.029416179
0.039659089	0.047923988
0.028692573	0.0284459
0.033753959	0.030169624
0.029047223	0.032035881
0.038410741	0.041479465
0.033797291	0.044768768
0.034189423	0.021066643
0.031422018	0.019022604
0.029595272	0.025056971
0.03797259	0.035340206
0.047351291	0.04407065
0.042260471	0.048477818
0.034604316	0.012003772
0.036660706	0.027498424
0.039156956	0.039911749
0.034810212	0.042399757
0.03166511	0.030459507
0.040004714	0.042612019
0.036489463	0.029961257
0.021640579	0.01735401
0.036095539	0.022342532
0.03882443	0.037857811
0.031769964	0.024569897
0.037606464	0.033014377
0.021657969	0.017125244
0.035820091	0.021125825
0.040426766	0.033587154
0.035548445	0.034526362
0.040908347	0.042216704
0.028783701	-0.002562106
0.0348896	0.014465732
0.04051848	0.036225245
0.039488504	0.065260775
0.037495922	0.037299139
0.031785308	0.034414425
0.038393582	0.046964835
0.033565365	0.017244675
0.034860527	0.023177519
0.034275548	0.025812143
0.038109864	0.042390846
0.03498031	0.040020075
0.03833516	0.032223559
0.036739773	0.048362703
0.034569501	0.036027263
0.036860962	0.029601998
0.028115278	0.027336902
0.032242332	0.02992825
0.034650642	0.024931664
0.041667022	0.042430677
0.031513455	0.031720193
0.03085128	0.006120284
0.02789889	0.038662269
0.033497084	0.032875174
0.032377254	0.035978308
0.040382898	0.042464784
0.038172244	0.041245883
0.032216229	0.041285549
0.042220885	0.029159094
0.035126265	0.025750048
0.04056202	0.042841185
0.046930172	0.03487365
0.042756635	0.018934563
0.022567425	0.019981276
0.03297295	0.040697521
0.028473836	0.018326119
0.042363788	0.040905999
0.034381951	0.035551629
0.033613631	0.032739266
0.042918686	0.018858044
0.032086421	0.042994562
0.032871207	0.035005192
0.028743814	0.027261488
0.03665754	0.037496218
0.042641998	0.042095803
0.045142463	0.045657961
0.029804071	-0.007651393
0.036128978	0.024417
0.040005306	0.042450818
0.027095913	0.018773475
0.036565207	0.047598932
0.040589827	0.050747074
0.038088741	0.039511606
0.036824264	0.025165689
0.025686594	0.014335893
0.043361915	0.019310106
0.037357643	0.023229448
0.030008758	0.021619683
0.039887302	0.034041029
0.028717443	0.035406904
0.047519501	0.034386329
0.030518834	0.041632591
0.034083341	0.047050867
0.033770782	0.016829495
0.03809241	0.006981578
0.036381345	0.032678824
0.040233083	0.017287743
0.029330104	0.03661632
0.030983155	0.016900332
0.031564973	0.029729739
0.031240871	0.02548551
0.022868113	0.008453867
0.030304431	0.020602712
0.036909239	0.024175586
0.040762361	0.031240803
0.035897098	0.034395814
0.026995116	0.028473412
0.039468549	0.030435275
0.036711773	0.023944225
0.034779302	0.028766729
0.034376281	0.027561726
0.041996978	0.031701051
0.033077701	0.019583772
0.040726298	0.022042397
0.039816828	0.020676053
0.040090959	0.028422483
0.040332291	0.027934429
0.039210698	0.040344128
0.037904809	0.041145905
0.041657603	0.036348521
0.035820713	0.03684944
0.034884232	0.039033382
0.03514232	0.025853867
0.037943882	0.02194579
0.030206327	0.019499505
0.042004657	0.046286737
0.039471474	0.041810105
0.022709669	0.021228896
0.039148344	0.034315353
0.037977008	0.03411393
0.029998374	0.017858963
0.027195018	0.028979251
0.036978912	0.026450846
0.027766527	0.01591574
0.045917547	0.032014335
0.026727061	0.010804355
0.045954649	0.036072311
0.043085902	0.044215603
0.033132002	0.026179391
0.034714622	0.043384119
0.042900561	0.040876238
0.048846929	0.045283291
0.036260019	0.021988673
0.033618717	0.019788759
0.040112567	0.045971319
0.040145008	0.045160756
0.03691771	0.024374313
0.032382056	0.026002318
0.042864869	0.017856298
0.041755237	0.037666773
0.028605079	0.017597214
0.031846153	0.013601122
0.026909712	0.021821001
0.030823713	0.032531746
0.033200103	0.0098969
0.027619228	0.01447401
0.031136539	0.014996532
0.047502247	0.057244215
0.027273805	0.008462228
0.033948289	0.029631852
0.031023059	0.022187679
0.035764591	0.024786618
0.043134998	0.020154313
0.036557086	0.037116671
0.033868493	0.036893695
0.032288773	0.041607408
0.0457233	0.048835928
0.026554573	0.028431585
0.030107926	0.029636054
0.043996769	0.064926465
0.045179395	0.02952863
0.02762192	0.026943195
0.026604388	0.028667383
0.040182244	0.032983441
0.038030116	0.045123107
0.037592316	0.047424177
0.03301657	0.037853714
0.021882463	0.023317212
0.025143807	0.019965863
0.03204144	0.019228843
0.05340409	0.049409257
0.040193331	0.031897314
0.034848617	0.04339814
0.038723592	0.036313616
0.039180485	0.038889444
0.033712066	0.03309501
0.043817011	0.039015623
0.036537669	0.0196548
0.034131194	0.044835105
0.034115662	0.029716376
0.037702445	0.032258569
0.035848882	0.024227942
0.031959132	0.042589852
0.038378844	0.021289183
0.036687356	0.046832844
0.028906653	0.041026138
0.045331847	0.04578895
0.039345578	0.028454763
0.027431572	0.025336367
0.028947461	0.030903285
0.048736838	0.034837648
};
\addplot [thick, blue, mark=o, mark size=3, mark options={solid}, only marks]
table {%
0.037758187	-0.034958687
0.030827564	-0.044788181
0.054810121	-0.043886386
0.020566765	-0.020604261
0.027267384	-0.032480144
0.03804878	-0.054020216
0.045795833	-0.042689707
0.029512168	-0.032894836
0.031846883	-0.04940782
0.02543128	-0.021156641
0.027314295	-0.024500482
0.029293799	-0.026832168
0.022906981	-0.024807999
0.030434464	-0.026303771
0.031143617	-0.035112241
0.033157798	-0.023321078
0.039389371	-0.04625794
0.026473788	-0.020511272
0.041579851	-0.041075932
0.032559038	-0.018147588
0.029780392	-0.019658875
0.039001758	-0.045933347
0.030807647	-0.044108832
0.025211459	-0.019595854
0.031522401	-0.027234168
0.033987573	-0.045472383
0.026506917	-0.03798688
0.030203231	-0.041517443
0.03847486	-0.024632894
0.036556487	-0.0334976
0.036607715	-0.033822112
0.026973217	-0.017010564
0.021118042	-0.026857102
0.033069029	-0.015124501
0.037302249	-0.030187651
0.024693731	-0.036600981
0.052052475	-0.047947766
0.042828884	-0.023218331
0.027047239	-0.02885399
0.033870424	-0.029706292
0.037454704	-0.057840279
0.035304766	-0.033060081
0.038343319	-0.0528508
0.035411888	-0.042441544
0.034385942	-0.042122171
0.031120432	-0.021105099
0.023017699	-0.028707243
0.032970855	-0.037308858
0.03375076	-0.030915823
0.041136678	-0.035022662
0.04079914	-0.053232661
0.032963585	-0.036272942
0.034211334	-0.044420053
0.036690213	-0.04886417
0.031492957	-0.018648809
0.026520998	-0.036418714
0.039811159	-0.052323286
0.034033011	-0.027253082
0.031108288	-0.036474591
0.032129058	-0.025514282
0.032450687	-0.053866498
0.031813975	-0.026928969
0.045539125	-0.055589391
0.035263614	-0.046959114
0.02899619	-0.019336719
0.040724691	-0.028872578
0.039709047	-0.051714916
0.040579709	-0.022101666
0.035542502	-0.045387745
0.029131547	-0.042522738
0.045641346	-0.046587042
0.043938895	-0.035784525
0.031281563	-0.030226469
0.032602608	-0.038514931
0.034633795	-0.038799952
0.029751269	-0.031666705
0.031375747	-0.041515251
0.046020055	-0.041384025
0.035071393	-0.032065145
0.025014117	-0.042567297
0.038014417	-0.017063044
0.038317866	-0.048402255
0.030628061	-0.023624219
0.03493132	-0.051255087
0.036784577	-0.032492951
0.030171772	-0.041162435
0.039556536	-0.050591908
0.037101158	-0.036491138
0.037309011	-0.050808389
0.024003673	-0.0382097
0.037510425	-0.020475983
0.034326857	-0.04079728
0.030914142	-0.013712281
0.031551355	-0.034089973
0.035811366	-0.038683933
0.026850907	-0.03268748
0.047435443	-0.044007281
0.02781885	-0.019078831
0.029519734	-0.031017909
0.027456183	-0.014458398
0.037674077	-0.043982624
0.038858304	-0.052358111
0.031641316	-0.039852568
0.023604695	-0.035375837
0.037658749	-0.031570178
0.041523365	-0.037832334
0.037271486	-0.031728035
0.04308503	-0.04090412
0.023862277	-0.031261947
0.038124558	-0.044830053
0.046391464	-0.057919566
0.024712471	-0.012847633
0.039033606	-0.041266403
0.035604664	-0.047267557
0.031100517	-0.02234736
0.0299985	-0.043505225
0.04170444	-0.065412202
0.030972699	-0.031654085
0.02944435	-0.038411606
0.039419116	-0.047278189
0.030196859	-0.023970947
0.030927001	-0.027185824
0.022528151	-0.024890719
0.030012015	-0.031774894
0.036104856	-0.04225614
0.027720766	-0.036373705
0.027453725	-0.027347744
0.039022577	-0.03775241
0.031528275	-0.031859464
0.039284702	-0.029707309
0.033278309	-0.036288053
0.038725153	-0.058538616
0.031287795	-0.03188399
0.03040491	-0.044785438
0.032632477	-0.034111695
0.029984111	-0.030350035
0.027721612	-0.024256424
0.037974737	-0.016960684
0.032408308	-0.030014485
0.035679274	-0.04275715
0.039999937	-0.034703555
0.037746716	-0.029070443
0.037546719	-0.046039993
0.032273291	-0.030827906
0.030895833	-0.025325123
0.035786283	-0.039896772
0.027585323	-0.018251871
0.039312672	-0.03801061
0.026665517	-0.035990094
0.03857335	-0.036063547
0.037844816	-0.035836078
0.028527563	-0.025607158
0.032092127	-0.043779501
0.037857849	-0.046710538
0.035159188	-0.037295763
0.033741563	-0.050130947
0.038822124	-0.049419933
0.035714594	-0.041176595
0.027605338	-0.035090794
0.035983801	-0.032679756
0.02442742	-0.019269923
0.032211494	-0.060989495
0.043076044	-0.043340004
0.033463329	-0.046766563
0.038477427	-0.048348995
0.042736224	-0.033711154
0.027394603	-0.020543469
0.034751063	-0.025459042
0.039545067	-0.049605034
0.028490984	-0.040960954
0.041702071	-0.048545639
0.026660577	-0.02238557
0.032666812	-0.04654817
0.040769531	-0.042496408
0.029129293	-0.014442266
0.036333587	-0.04787385
0.028497935	-0.020772641
0.036425895	-0.032062734
0.029814189	-0.041842768
0.045917056	-0.048745166
0.026356411	-0.027289499
0.037938676	-0.057433091
0.031131663	-0.031150182
0.031809357	-0.026847011
0.029683385	-0.031738987
0.030774914	-0.0469834
0.028059851	-0.028358692
0.034336304	-0.040178425
0.038890181	-0.031726025
0.035844559	-0.039680216
0.026647814	-0.036166899
0.040147483	-0.034903035
0.033892291	-0.029326183
0.031769664	-0.024303107
0.037111807	-0.027066523
0.029696549	-0.046080505
0.042986471	-0.030695123
0.02101774	-0.027392981
0.04050444	-0.060491302
0.02072799	-0.030628643
0.035533011	-0.042558179
0.040520879	-0.045827395
0.035088487	-0.043507412
0.041120351	-0.033875383
0.047063711	-0.051406677
0.033798332	-0.036695152
0.019271852	-0.029334677
0.036907557	-0.035312759
0.038192178	-0.040289401
0.036800952	-0.03100547
0.03755925	-0.039244314
0.033152206	-0.039739576
0.031499699	-0.036360794
0.045620473	-0.027641149
0.03589358	-0.037399838
0.027280688	-0.015260137
0.039374399	-0.046060857
0.034511731	-0.032367616
0.036706155	-0.054132593
0.025842402	-0.041421719
0.038915027	-0.046324893
0.027899058	-0.038943122
0.037301326	-0.023475205
0.040438257	-0.065934685
0.036955646	-0.05745148
0.04403018	-0.046615675
0.030396283	-0.039876281
0.030060048	-0.022711494
0.038099312	-0.035607085
0.030908128	-0.024478693
0.043840197	-0.038133325
0.032927813	-0.037868515
0.024746016	-0.020958516
0.026563797	-0.03200426
0.030684223	-0.024103523
0.033060847	-0.043010729
0.038398533	-0.051341034
0.036864777	-0.035980556
0.028981476	-0.028851429
0.033333946	-0.024923628
0.033859139	-0.038680641
0.037631858	-0.046476931
0.033704579	-0.031023048
0.032104091	-0.041670162
0.035939276	-0.027238436
0.044681389	-0.032285443
0.028886215	-0.024254654
0.027557429	-0.025427119
0.031171746	-0.048897436
0.0215578	-0.01929508
0.023656141	-0.039484988
0.031384026	-0.044893687
0.017764151	-0.022685203
0.037129876	-0.022426329
0.022535049	-0.020113682
0.039673219	-0.040975928
0.043142234	-0.040081875
0.033615704	-0.015247001
0.024678713	-0.043519655
0.033706191	-0.005709732
0.034715943	-0.021863991
0.035926097	-0.029244021
0.027755004	-0.023994487
0.034202131	-0.038032563
0.026465412	-0.02741612
0.041341275	-0.047542995
0.041494865	-0.037401191
0.028351631	-0.032229503
0.046090047	-0.058639892
0.039118094	-0.045474916
0.03180299	-0.027660452
0.03527099	-0.02482285
0.028971528	-0.041697374
0.038028523	-0.032712197
0.03177426	-0.023808512
0.034801893	-0.031403981
0.042653178	-0.035736914
0.034378441	-0.032447995
0.025472723	-0.036134367
0.025754734	-0.023984657
0.043857608	-0.0375034
0.030537771	-0.021124818
0.033298922	-0.01835726
0.031028499	-0.045218282
0.031389514	-0.031551609
0.034650269	-0.036069389
0.041085767	-0.028675089
0.032452699	-0.029658803
0.037774205	-0.021607476
0.042386675	-0.067286952
0.037741457	-0.035590888
0.030713146	-0.022521618
0.039324821	-0.039253791
0.053547412	-0.044509427
0.039636386	-0.041862027
0.036461613	-0.025562667
0.020231389	-0.01172492
0.034550571	-0.043166613
0.037783584	-0.030216097
0.0370136	-0.047816101
0.040935054	-0.042214857
0.036027309	-0.036335861
0.035652551	-0.015185723
0.026438052	-0.029033433
0.037411233	-0.065061106
0.038150127	-0.050863403
0.029391699	-0.020675799
0.03668478	-0.040631378
0.030430169	-0.018643475
0.032236382	-0.030945858
0.040389879	-0.023247088
0.040155378	-0.05172506
0.029990901	-0.027173778
0.043252563	-0.046061885
0.038229518	-0.039877421
0.025168249	-0.026660577
0.024367683	-0.027439199
0.036590728	-0.027903924
0.036194489	-0.036638179
0.04237493	-0.042561599
0.032587126	-0.042797099
0.024008313	-0.030556382
0.03623502	-0.03370847
0.038442372	-0.036692842
0.035108531	-0.054703204
0.043739096	-0.038475272
0.035665863	-0.047872924
0.030360553	-0.039265251
0.036748858	-0.047957123
0.031386081	-0.028882102
0.035979475	-0.03441286
0.029129516	-0.032542005
0.032478673	-0.025041273
0.040025685	-0.045858097
0.034615626	-0.041806865
0.027686075	-0.021747669
0.035836213	-0.012458127
0.044040745	-0.067785658
0.033582871	-0.02651423
0.041932441	-0.026173519
0.032202441	-0.032189858
0.03580178	-0.039005291
0.026569257	-0.012338682
0.039464649	-0.049284166
0.039517394	-0.042016875
0.028950574	-0.018939024
0.01994284	-0.027591685
0.030200386	-0.024214468
0.028327661	-0.032588879
0.035895708	-0.042592235
0.034366166	-0.055308513
0.033948356	-0.018797142
0.041796196	-0.043524872
0.036455021	-0.038395601
0.02528099	-0.022895539
0.034927281	-0.048632672
0.042605627	-0.036121002
0.033998165	-0.025115096
0.034866087	-0.028456574
0.041038796	-0.04115355
0.027697866	-0.034089226
0.03379316	-0.004611142
0.030620871	-0.027884397
0.031208013	-0.034758539
0.030063441	-0.019629974
0.030626274	-0.041632597
0.039619517	-0.036342058
0.034070964	-0.047307388
0.032971141	-0.048976935
0.034568238	-0.043987687
0.040554235	-0.04713925
0.045266295	-0.033222689
0.031129951	-0.021134299
0.027147072	-0.051390496
0.028879111	-0.028356921
0.039500056	-0.031908261
0.034116048	-0.042643868
0.037536551	-0.042038667
0.043716069	-0.046895407
0.024621587	-0.019991312
0.032886713	-0.039822999
0.03620759	-0.024327595
0.046098259	-0.03839575
0.046585927	-0.056551291
};
\end{axis}

\end{tikzpicture}

%% file: Mixture_wlp.tex
\begin{tikzpicture}

\begin{axis}[
tick align=outside,
tick pos=left,
x grid style={white!69.0196078431373!black},
xmin=0.015, xmax=0.06,
xtick style={color=black},
y grid style={white!69.0196078431373!black},
ymin=-0.15, ymax=0.15,
ytick style={color=black},
width=6.5cm,height=6cm,
]
\addplot [thick, red, mark=o, mark size=3, mark options={solid}, only marks]
table {%
0.033982321	-0.035072159
0.032230298	0.069970125
0.035559299	-0.019651545
0.034323028	-0.06728304
0.034504543	-0.031944277
0.038799853	0.034991553
0.038530883	0.043918664
0.036917211	-0.054053401
0.040051426	-0.031631711
0.03984524	0.031207203
0.037828614	-0.054380921
0.028454705	0.002485821
0.041613107	-0.021947125
0.026645021	0.050690887
0.040313546	0.055920162
0.037620036	-0.047763393
0.028120256	-0.041462294
0.028609139	-0.001556889
0.022293252	-0.03702726
0.029778028	0.016780503
0.046072129	-0.04129061
0.031895445	-0.042443943
0.040806055	-0.038321929
0.031742182	-0.03801376
0.036711245	0.009897889
0.038869635	-0.012420262
0.037356837	0.013692122
0.03034663	0.006392534
0.044012541	0.023875508
0.038021976	-0.007867978
0.028070484	0.001523283
0.027955371	0.017647621
0.042996803	-0.048876619
0.035482683	-0.030396881
0.041031942	-0.063642733
0.0438492	-0.014840771
0.035201802	-0.00569529
0.037001482	0.045280493
0.032092419	-0.009245036
0.043675297	-0.022977205
0.042584166	0.017081182
0.033635807	-0.04598113
0.023146905	-0.040475552
0.033324979	-0.014733032
0.038899126	0.003916114
0.031833995	-0.003598308
0.03659686	-0.03967006
0.033725786	-0.033034944
0.035307369	-0.065696654
0.029698219	-0.05346323
0.037337246	0.021770154
0.031284297	0.002479418
0.039866452	-0.030024493
0.041715139	0.059550325
0.037321644	0.006502719
0.019483029	-0.003764617
0.029079199	-0.055170216
0.04162042	-0.025838399
0.028127758	-0.004991226
0.032223534	0.049767592
0.02746418	-0.01675804
0.021291745	0.014482096
0.033886044	0.050307628
0.033355367	0.013690511
0.027238946	-0.018410555
0.039740283	-0.037760857
0.030072908	-0.102983902
0.035497142	-0.014421783
0.025614045	-0.041921718
0.036825676	-0.022944081
0.028124272	0.011970451
0.034063317	0.00833436
0.033107866	0.005511302
0.03690487	0.005136198
0.042660601	-0.007692878
0.033300093	-0.022949413
0.035753272	-0.020953288
0.024051185	0.055273372
0.023310392	0.027381488
0.031965126	0.002878753
0.02960665	-0.00884238
0.03095744	0.04975203
0.036017329	0.028070701
0.038821275	0.027701491
0.031362541	-0.014863296
0.027587901	-0.012011545
0.034851864	-0.048640987
0.039332058	0.017320783
0.045697472	-0.051301026
0.034075312	0.031647039
0.030508321	0.025455138
0.026622141	0.003970903
0.033631737	0.025668997
0.034367406	-0.0331357
0.04596334	-0.012076687
0.0313783	-0.054780421
0.025910566	-0.036273878
0.036794032	-0.048421287
0.037149179	0.012563933
0.037579418	0.035849758
0.048223754	0.006773446
0.036489238	-0.017976106
0.045662119	-0.03874433
0.039025818	0.006468831
0.02624043	-0.030085266
0.042717273	-0.017836834
0.043677954	-0.02455903
0.027535218	-0.009279215
0.03644763	-0.021372745
0.031763397	0.023346677
0.038883302	0.00351312
0.044702038	-0.06012114
0.030755349	-0.006854597
0.036606283	-0.018699281
0.039442977	0.027121925
0.036390859	0.014990767
0.036693945	-0.002633423
0.023839485	-0.027118239
0.032848975	-0.006481979
0.02666633	-0.050681297
0.037630557	0.009983971
0.030418489	-0.032537341
0.038544461	-0.064699783
0.032465127	-0.054619478
0.031342867	0.012121263
0.044759402	0.021632541
0.033522641	0.033762924
0.026205511	0.00514955
0.031767998	-0.038864423
0.030999503	-0.036344134
0.040193169	-0.079938262
0.037091691	0.055104445
0.027315195	0.006307671
0.027521053	-0.029255007
0.028755007	0.006520611
0.027790819	0.003383647
0.030817831	-0.036144139
0.038867312	-0.005070922
0.043776546	0.024384801
0.042419983	-0.020050662
0.032789704	0.035049872
0.030195533	-0.008352782
0.043041304	-0.048708681
0.031159543	-0.002942397
0.036052074	-0.045561287
0.026972618	-0.008411607
0.034635464	0.007264696
0.034950797	-0.023124025
0.040326036	-0.021511947
0.036015597	0.001771017
0.030585864	-0.009875823
0.044504088	0.006586374
0.035711253	0.009608948
0.039964988	0.004672145
0.029421547	-0.057556235
0.045984	-0.039361236
0.03529412	0.073673697
0.029272308	0.063269717
0.041331349	-0.020526359
0.035673747	0.002782393
0.029602462	-0.018835613
0.027828755	0.027061061
0.03648013	0.050771941
0.044405477	-0.094817537
0.034930785	-0.072984982
0.033275579	-0.018512345
0.037168422	0.013869375
0.035190902	-0.036209846
0.038357232	0.023093997
0.041879192	0.059650091
0.037866048	0.022492157
0.028709797	0.036327064
0.041737852	-0.100473651
0.032310136	-0.045596784
0.039117633	-0.032242921
0.033740015	-0.055288759
0.03427007	0.018836624
0.03266102	-0.007342762
0.034113737	0.038076556
0.035356819	0.024071578
0.028484524	-0.049124522
0.037453292	-0.052651252
0.035789876	-0.044006802
0.049716758	0.01795096
0.036267069	-0.02763483
0.040798009	-0.023837373
0.024457799	0.035257
0.031468473	0.009129355
0.040912367	0.037824146
0.029982804	0.016652036
0.04269304	0.011453436
0.033090067	-0.014132243
0.040350587	0.000279254
0.032111127	-0.006711224
0.037143925	-0.008006271
0.038747761	0.011775955
0.037394357	-0.062532364
0.034974992	0.01494606
0.041347349	0.04325361
0.043905411	-0.04012301
0.034517735	-0.066155739
0.02327123	-0.043164059
0.020282751	0.005978514
0.032662014	-0.029653994
0.040028234	0.014207904
0.030182509	0.003419983
0.035420045	-0.071675496
0.038005325	-0.029270846
0.044949515	-0.028947496
0.022766665	0.004378064
0.028736413	0.024798203
0.032284406	0.071432616
0.036984317	0.052060083
0.032009028	0.034988552
0.046255732	-0.142650071
0.040079751	0.00807398
0.039230333	-0.0332175
0.027845718	-0.010740935
0.033317543	-0.019330412
0.028425198	-0.02498757
0.037375292	-0.043977806
0.032850005	-0.050763016
0.033671183	-0.0150275
0.031673359	0.04022898
0.029486911	-0.022170373
0.037623252	-0.024832406
0.046579362	-0.015295319
0.041835676	-0.066740218
0.034557277	0.04033684
0.036330974	-0.008207382
0.039108675	-0.023715349
0.034209658	-0.041212376
0.031391047	-0.015727966
0.039450638	-0.020886396
0.036519269	0.039683626
0.021606819	0.001780454
0.035776196	-0.022458177
0.038286653	-0.010866869
0.031931239	-0.027548101
0.037342798	-0.031795651
0.021269753	-0.046424828
0.035593033	-0.001652685
0.039426514	0.011995729
0.035292704	-0.050418841
0.039992723	-0.003869379
0.028646363	0.056102313
0.034732687	0.049354991
0.039842325	0.01930192
0.038136542	-0.035972144
0.036824217	-0.019814531
0.031314432	-0.068340549
0.037326491	-0.042731682
0.03326823	0.038594934
0.03505142	-0.014469961
0.034318502	0.027865239
0.037182721	-0.050462833
0.03416591	-0.017815886
0.03749287	-0.009306753
0.036217971	-0.066726473
0.034047176	-0.023715861
0.03646051	0.044069817
0.028344207	-0.012154429
0.032148455	-0.018457728
0.034532372	0.007061552
0.041341128	-0.022038357
0.030599169	0.001897245
0.031071534	0.038774061
0.027514591	-0.047133171
0.032861271	-0.025959041
0.032346974	-0.047797758
0.039681474	-0.011519728
0.037461309	-0.014904167
0.031569858	-0.06171851
0.041776175	-0.003801023
0.034746944	0.003810169
0.040060624	-0.016723551
0.046946575	0.009392734
0.042893586	0.046516749
0.022512376	-0.010092155
0.032228557	-0.094102418
0.028348467	0.029701351
0.041935252	-0.020366573
0.033799478	-0.010411159
0.033265614	-0.015663739
0.043222181	0.034041856
0.03098489	-0.033290465
0.032454803	-0.040679774
0.028268528	0.022666783
0.036259385	-0.017672936
0.042151269	-0.029289356
0.044942846	-0.052417073
0.030565631	0.082911954
0.036322414	-0.023625575
0.039568146	-0.031369855
0.026736516	0.025342573
0.035830856	-0.068408219
0.039595265	-0.017915947
0.037613806	-0.034354651
0.036292721	-0.019286075
0.0258148	0.044038679
0.043455338	0.010781783
0.037377399	0.028348343
0.03011668	-0.022468369
0.039232783	-0.018881657
0.027840252	-0.031594653
0.047060604	0.003643741
0.030081631	-0.063726876
0.033350453	-0.109430162
0.033904656	0.056947472
0.038287015	0.060344092
0.036253666	-0.055099467
0.040256047	0.003322063
0.028599612	-0.027302827
0.030670723	0.033699437
0.030731889	-0.028049009
0.031230784	-0.000927481
0.023195948	-0.018294741
0.030138612	-0.028273582
0.037080747	-0.016175559
0.039770885	0.022009757
0.035598376	-0.046012434
0.027064708	0.014940747
0.039213364	0.019018506
0.037040578	0.038937088
0.035142061	-0.037445429
0.034219516	-0.015001793
0.041441223	-0.028332619
0.032867176	0.058458507
0.040566846	0.008568799
0.039865731	0.004942557
0.039398956	0.037136064
0.040364043	0.028168737
0.038497234	0.007078534
0.036738646	-0.004699719
0.040929454	0.008447122
0.035291957	-0.000726204
0.034335511	-0.035803357
0.034707287	-0.00850597
0.03828072	-0.024390537
0.030566273	0.011975454
0.041870842	-0.023590466
0.038729764	-0.003390498
0.022460893	-0.00546513
0.038684563	0.017173208
0.037235448	-0.017894244
0.029780097	0.004243028
0.026744977	-0.017709941
0.037063804	-0.000764444
0.027276032	0.008805726
0.045536665	0.000276552
0.027026996	-0.001269069
0.045404984	-0.03155725
0.042053039	-0.01197658
0.032485143	-0.013537039
0.034340934	-0.056560831
0.042733042	-0.031453988
0.0479404	-0.058517043
0.036056653	0.025443313
0.03340297	0.007146487
0.039258477	-0.042136475
0.039940372	-0.03025159
0.03687714	0.031272035
0.032146261	-0.040218733
0.042633825	0.02094081
0.041694798	-0.008029742
0.028566839	0.030624249
0.031778955	0.022101272
0.02674101	-0.003774143
0.030611668	-0.023347458
0.033470828	0.064237326
0.027053661	0.008338287
0.031098907	0.031304982
0.046982573	-0.042272681
0.027495928	0.021160731
0.033672098	-0.02124971
0.030576284	0.003866405
0.03566663	-0.006355863
0.043833367	-0.014584033
0.03587594	0.023370237
0.033481779	-0.010175297
0.031387697	-0.054297736
0.044790163	-0.072924584
0.026074018	-0.040847858
0.0294736	0.006485076
0.042935592	-0.060860054
0.04458986	0.011599257
0.027083179	-0.025408908
0.02647847	-0.030331427
0.040107587	-0.01430051
0.037563118	-0.008025487
0.037151444	-0.050990563
0.032643527	-0.071557417
0.021678932	-0.039198198
0.024822102	0.03117219
0.031507663	0.043174851
0.053052888	-0.053191828
0.03993223	0.015860927
0.033761189	-0.048313852
0.037784063	0.025846527
0.038546162	-0.042958364
0.033351808	-0.017523793
0.043070979	0.001553974
0.036677417	-0.012540161
0.033546091	-0.047129764
0.033825037	-0.014666548
0.037319415	-0.00375131
0.035481814	0.017929741
0.031235668	-0.057953133
0.038597833	0.036048429
0.035971968	-0.071863127
0.028253607	-0.010752323
0.044683636	-0.026646046
0.039062111	0.031831183
0.027079194	0.012953683
0.028507188	-0.006688177
0.048242921	0.009423347
};
\addplot [thick, blue, mark=o, mark size=3, mark options={solid}, only marks]
table {%
0.038160566	0.002579069
0.031022822	0.010135177
0.055661456	-0.05629147
0.020565252	0.024668199
0.027799882	0.009070737
0.03837863	0.084091541
0.04655276	0.015043892
0.030347511	0.018653074
0.032225073	0.023710879
0.025748729	-0.006490114
0.027674182	-0.028521485
0.029894294	-0.036970072
0.02314458	0.009701037
0.031204726	0.033009536
0.031296235	-0.012670788
0.033764201	0.027775915
0.039530376	0.017906721
0.027189795	-0.033310565
0.042503396	-0.000317003
0.03266073	-0.017036831
0.030589913	-0.033590369
0.039440091	0.02379578
0.031430694	0.026592877
0.025481682	-0.036101209
0.032198189	-0.035640169
0.033763358	0.059988743
0.026662761	0.02924051
0.03031893	0.018774443
0.038930604	-0.025932132
0.03707288	0.024467068
0.036927122	0.031765799
0.027289452	-0.021437374
0.021079056	0.049205947
0.033408593	-0.065369375
0.03821461	-0.017322673
0.025097765	0.043758827
0.052201477	0.039130899
0.043801962	-0.009511605
0.027327786	-0.022988401
0.034478177	0.011331131
0.037430576	0.078072107
0.035899333	-0.023477025
0.038715816	0.024188996
0.035967628	5.79E-05
0.035183499	0.02347406
0.031916952	0.019905479
0.023886163	0.02484626
0.033595893	0.019489027
0.033917712	0.022690716
0.04140068	0.003547735
0.040846389	0.029936509
0.033418474	0.038122004
0.034466354	0.029584718
0.03677588	0.031006924
0.031638468	-0.030954318
0.026610748	0.031388246
0.040280299	0.023790236
0.034068984	-0.002703063
0.031435863	0.030953844
0.032200185	-0.003253472
0.033202335	0.06190226
0.032727308	0.008665496
0.045730708	0.045382876
0.035579305	0.041847918
0.029118518	0.039220138
0.04098804	-0.011545851
0.039967963	0.080286084
0.04113189	-0.041318239
0.036172027	0.007453081
0.029289518	0.05778147
0.046475366	-0.048489673
0.044857848	-0.013316397
0.031568377	0.0261302
0.033480636	0.051448071
0.035381179	0.02012487
0.029991342	0.017001793
0.032401229	0.063768948
0.046086711	-0.044568402
0.035653442	0.018508892
0.025025695	0.068712383
0.038621463	-0.04564434
0.038651346	0.015478516
0.031255403	-0.019659062
0.035225595	0.096266996
0.037650286	-0.009995508
0.030419798	0.05558893
0.039850912	0.00019507
0.037672803	0.035329021
0.037370923	0.014891893
0.02393891	0.024247545
0.038112247	-0.050359703
0.034774998	0.00924517
0.031306619	-0.062793268
0.032171549	0.018069868
0.035904228	-0.01429913
0.026901619	0.036846899
0.047827065	0.029919554
0.027934725	0.00080287
0.030016934	-0.02074664
0.027786649	-0.019029914
0.038182595	0.043859458
0.039184176	0.037006076
0.032248614	0.067509301
0.024166314	0.017406501
0.038308026	0.00746764
0.041823446	0.020224943
0.037887167	-0.027863706
0.043872934	0.001750066
0.024169938	0.001543167
0.038450632	0.049871381
0.047080214	0.049087149
0.025085609	-0.03306596
0.03918319	0.006223276
0.035713747	0.088562355
0.031453124	0.009753236
0.030456951	0.023334671
0.042234563	0.111407265
0.031560283	0.004628816
0.029749904	0.006304148
0.040096172	0.04561898
0.030696233	-0.035775717
0.031447367	0.028131523
0.02284227	-0.009409066
0.029895613	0.007044034
0.036339847	0.036409237
0.02799225	0.049508127
0.028091354	-0.01754047
0.039309026	0.040072547
0.031910508	0.032405635
0.039609525	-0.05672422
0.033943652	0.026140589
0.039016535	0.054717288
0.031701558	0.035072987
0.030705029	0.007229233
0.03311958	0.036428878
0.03061192	-0.000823773
0.028195132	-0.0254473
0.038973562	-0.012182964
0.032961561	0.015595019
0.036738173	-0.001873451
0.0406091	-0.000856725
0.038021781	-0.011521804
0.037984099	0.009526176
0.033022785	-0.002831871
0.031303239	-0.000577553
0.036612096	0.050704033
0.028010998	-0.009667924
0.039726601	0.034606527
0.027125272	0.049476316
0.039455115	0.016668214
0.038196274	-0.00604687
0.02953269	-0.017719009
0.032486525	0.04764822
0.038134563	0.045381551
0.03597581	-0.000773931
0.034397082	0.027694576
0.039611763	0.052281585
0.035849465	0.003949411
0.027697942	0.02078881
0.036558832	0.003028661
0.024764641	0.003935446
0.032019871	0.108385141
0.043840429	0.012680692
0.033887554	0.041902655
0.038678416	0.005939831
0.043510144	-0.014646392
0.027616094	0.023170344
0.03545756	0.017020256
0.040119344	0.043618133
0.02844477	0.052318814
0.041650034	0.038113871
0.027108665	-0.002989099
0.033186818	0.032073771
0.041395658	0.038389234
0.029471349	-0.05750447
0.036737276	0.078072316
0.02888637	-0.019095155
0.036727791	-0.016492238
0.02999832	0.026134834
0.046977033	0.025358019
0.027067888	0.012260775
0.038687767	0.098525542
0.03192486	0.002286639
0.032198588	0.027281773
0.030279956	0.015784642
0.031107591	0.050781359
0.028835769	-0.009803553
0.0350628	0.022545866
0.039539484	-0.039958396
0.036178571	0.015834137
0.027276675	0.065772226
0.040711185	0.004489134
0.034280863	0.0065394
0.032571784	0.009041111
0.037695094	-0.009776047
0.030400032	0.033888564
0.04390452	0.02233077
0.020879405	0.00284023
0.041200072	0.089315675
0.021146637	0.053789965
0.035783071	-0.001951575
0.041100579	0.009058409
0.035351335	0.012391359
0.041354682	0.013131219
0.047670259	0.002203308
0.034195624	-0.005477883
0.019671222	0.042666238
0.037210344	-0.003704198
0.038586546	-0.031257489
0.037690189	0.0026393
0.038247862	0.009648606
0.033944523	0.010950591
0.031779663	0.005431095
0.046081379	-0.022692136
0.036701054	-0.029313433
0.027831286	-0.05759416
0.039842883	0.025432591
0.034929865	-0.015880831
0.036855884	0.085569783
0.026087744	0.046019971
0.039815962	0.009782168
0.028159693	0.033802602
0.037801747	-0.018561218
0.040688597	0.063710052
0.037412677	0.001486185
0.044335594	-0.001373123
0.030795117	0.090361742
0.030100795	0.000192184
0.038823656	0.040093885
0.031057165	0.008273984
0.044528716	-0.022574361
0.033448912	0.037467478
0.025044806	0.033469078
0.027152066	-0.003454401
0.031214744	-0.000263596
0.032800628	0.012373274
0.038966214	0.013367502
0.037495893	0.002362899
0.029568047	0.015280295
0.034104495	-0.02644678
0.034522078	0.00261763
0.038019623	0.029846228
0.034338235	-0.001285744
0.032459321	0.011716675
0.036326774	0.010945049
0.045479423	-0.014902182
0.029472801	0.014994649
0.027724561	-0.022898508
0.03132679	0.016565504
0.021770877	-0.000652832
0.024024019	0.034731694
0.031877648	0.037669382
0.017998888	0.048214504
0.037155945	-0.029597931
0.022892053	0.018207833
0.040107138	0.006813045
0.043949235	0.021393732
0.034010695	-0.041230035
0.024522164	0.054449224
0.033881415	-0.022058237
0.035170198	-0.003383567
0.036280322	-0.038502416
0.028027639	-0.015988078
0.034299735	0.026547881
0.026737655	-0.011238184
0.042007125	0.021379346
0.041808901	-0.035142821
0.028546552	0.019887625
0.046939773	0.035911628
0.039664145	0.021532414
0.032086156	-0.047421954
0.035685046	-0.031478945
0.029201767	0.003881101
0.038947559	-0.005925439
0.031787973	0.008547118
0.035289591	0.000579134
0.043500174	0.001235277
0.035135503	-0.025834209
0.025677196	0.029346336
0.026220049	0.042987087
0.044403674	0.005478369
0.031150135	-0.058446804
0.033987628	-0.020851182
0.031576907	0.047368585
0.032212195	0.001945722
0.035059455	0.018260603
0.041454286	-0.014173195
0.032916849	-0.058607234
0.03799199	-0.048197595
0.042847355	0.067391703
0.038286905	0.021831595
0.031367732	-0.010508389
0.0399542	-0.002947602
0.053713548	-0.004187328
0.040311667	-0.020214915
0.036794479	-0.066423729
0.020540329	-0.030729053
0.035062159	0.033776388
0.038037601	-0.018705993
0.03718687	0.00912624
0.041480307	-0.003041284
0.03682465	-0.007332686
0.035993573	-0.057402392
0.026740717	0.00318495
0.037898325	0.113829582
0.038410635	0.04334587
0.029748588	-0.031731846
0.036916603	0.009560673
0.031060694	-0.012523361
0.0325898	-0.0002193
0.041310817	-0.044119159
0.040893693	0.043184797
0.030290849	-0.003390268
0.043559175	-0.034042602
0.038943919	0.033195938
0.025644277	0.004768877
0.023838579	0.051791906
0.037090826	-0.035356061
0.036678396	0.022007203
0.042763301	0.01398205
0.032589363	0.039031469
0.024340201	0.032091807
0.036383938	0.009420336
0.039203899	-0.001214169
0.035409279	0.025226218
0.04449518	0.022294462
0.036053276	0.066894442
0.030635724	0.008797017
0.037111515	0.059437122
0.031648168	-0.02117805
0.036084711	0.009516986
0.02909348	0.05420964
0.033193816	-5.47E-06
0.040132158	0.030601053
0.034827172	0.078912624
0.028068434	-0.001188681
0.036019591	-0.050340549
0.044806256	0.039296176
0.034784575	-0.045992337
0.042275503	-0.024508152
0.032935215	0.013473336
0.036587114	0.009270552
0.026997808	-0.005842907
0.039943017	0.011412502
0.040304383	-0.012341607
0.029193752	-0.025829922
0.020408724	0.039790111
0.030127821	-0.041428539
0.028775816	0.023089544
0.036393199	0.024514953
0.034189046	0.105252918
0.03479388	-0.078873034
0.042253496	0.034386015
0.036957091	0.001045195
0.026000282	-0.021001947
0.035573709	0.041263851
0.043308636	-0.034859663
0.034464462	-0.045614829
0.035552252	0.022427373
0.042138154	-0.006546893
0.027996666	0.032954392
0.034619358	-0.047854083
0.030674944	0.004074365
0.032104373	-0.016558319
0.030610183	-0.037414908
0.030734207	0.039743476
0.040038048	-0.029865936
0.034158835	0.031686008
0.033014652	0.04207578
0.035533204	0.004104606
0.041057886	0.02506829
0.045177115	0.012342349
0.03148555	-0.011052562
0.026994447	0.096186504
0.029101807	-0.001688577
0.040266585	-0.007006299
0.034240839	0.029857263
0.037802296	0.059113143
0.044486058	0.015605418
0.025126783	-0.009007996
0.033152053	0.030327863
0.037026777	-0.01265358
0.046744837	0.001844548
0.046928217	0.052210991
};
\end{axis}

\end{tikzpicture}

%% file: legend_batchEm.tex
\begin{tikzpicture}
    \begin{customlegend}[legend columns=4, legend style={column sep=0.1ex, font=\normalsize, cells={align=left}},legend entries={Alg.~\ref{alg:em} $\gamma=0$ , Alg.~\ref{alg:em} $\gamma=0.2$ , Alg.~\ref{alg:em} $\gamma=0.4$ , Alg.~\ref{alg:em} $\gamma=0.6$, GLMM\cite{maretic2020graph} ,Spectral clustering, SpecTemp \cite{segarra2017network}, Kalofolias \cite{kalofolias2016learn}}]
    \addlegendimage{very thick, blue, mark=o, mark size=4, mark options={solid}, sharp plot}
    \addlegendimage{very thick, red, mark=o, mark size=4, mark options={solid}, sharp plot}
    \addlegendimage{very thick, color0, mark=o, mark size=4, mark options={solid}, sharp plot}
    \addlegendimage{very thick, green!50!black, mark=o, mark size=4, mark options={solid}, sharp plot}
    \addlegendimage{thick, orange, mark=diamond, mark size=4, mark options={solid}, sharp plot}
    \addlegendimage{thick, yellow!50!black, mark=square, mark size=4, mark options={solid}, sharp plot}
    \addlegendimage{thick, green, mark=triangle, mark size=4, mark options={solid}, sharp plot}
    \addlegendimage{thick, black, mark=oplus, mark size=4, mark options={solid}, sharp plot}
    \end{customlegend}
    \end{tikzpicture}

%% file: BatchEM_wlf_vsC_errorrate.tex
\begin{tikzpicture}

\definecolor{color0}{rgb}{0,0.75,0.75}

\begin{axis}[
legend cell align={left},
legend columns=1,
legend style={at={(0.5,0.8)}, anchor=north, draw=white!80.0!black},
tick align=outside,
tick pos=both,
title={},
x grid style={white!69.01960784313725!black},
xlabel={\Large Number of graphs $\mulG$},
xmajorgrids,
xmin=2, xmax=5,
xtick style={color=black},
y grid style={white!69.01960784313725!black},
ylabel={\Large Error rate},
ymajorgrids,
ymin=0, ymax=1,
ytick style={color=black},
width=7cm,height=6cm,
]
\addplot [ultra thick, blue, mark=o, mark size=4, mark options={solid}]
table {%
2 0.0255000000000001
3 0.0306666666666667
4 0.0660000000000000
5 0.0807999999999999
};
\addplot [ultra thick, red, mark=o, mark size=4, mark options={solid}]
table {%
2 0.0350000000000001
3 0.0463333333333336
4 0.0632499999999999
5 0.0679999999999998
};
\addplot [ultra thick, color0, mark=o, mark size=4, mark options={solid}]
table {%
2 0.0620000000000002
3 0.0556666666666669
4 0.0912500000000001
5 0.123200000000000
};
\addplot [ultra thick, green!50!black, mark=o, mark size=4, mark options={solid}]
table {%
2 0.102000000000000
3 0.129000000000000
4 0.181250000000000
5 0.257400000000000
};
\addplot [ultra thick, orange, mark=diamond, mark size=4, mark options={solid}]
table {%
2 0.917500000000000
3 0.914333333333333
4 0.896000000000000
5 0.892800000000000
};

\addplot [ultra thick, yellow!50!black, mark=square, mark size=4, mark options={solid}]
table {%
2 0.4045
3 0.5477
4 0.6077
5 0.6418
};

\addplot [ultra thick, green, mark=triangle, mark size=4, mark options={solid}]
table {%
2 0.791
3 0.784
4 0.8
5 0.7804
};
							
\addplot [ultra thick, black, mark=oplus, mark size=4, mark options={solid}]
table {%
2 0.897
3 0.884
4 0.881
5 0.8742
};
\end{axis}

\end{tikzpicture}

%% file: BatchEM_slf_vsC_errorrate.tex
\begin{tikzpicture}

\definecolor{color0}{rgb}{0,0.75,0.75}

\begin{axis}[
legend cell align={left},
legend columns=1,
legend style={at={(0.5,0.8)}, anchor=north, draw=white!80.0!black},
tick align=outside,
tick pos=both,
title={},
x grid style={white!69.01960784313725!black},
xlabel={\Large Number of graphs $\mulG$},
xmajorgrids,
xmin=2, xmax=5,
xtick style={color=black},
y grid style={white!69.01960784313725!black},
ylabel={},
ymajorgrids,
ymin=0, ymax=1,
ytick style={color=black},
width=7cm,height=6cm,
]
\addplot [ultra thick, blue, mark=o, mark size=4, mark options={solid}]
table {%
2 0.00150000000000006
3 0
4 0.000749999999999917
5 0.000599999999999934
};
\addplot [ultra thick, red, mark=o, mark size=4, mark options={solid}]
table {%
2 0.000499999999999945
3 0.000666666666666593
4 0.000999999999999890
5 0.000294117647058778
};
\addplot [ultra thick, color0, mark=o, mark size=4, mark options={solid}]
table {%
2 0.00349999999999995
3 0.00200000000000000
4 0.00124999999999997
5 0
};
\addplot [ultra thick, green!50!black, mark=o, mark size=4, mark options={solid}]
table {%
2 0.00349999999999995
3 0.000999999999999890
4 0.00200000000000000
5 0
};
\addplot [ultra thick, orange, mark=diamond, mark size=4, mark options={solid}]
table {%
2 0.906000000000000
3 0.899333333333333
4 0.883750000000000
5 0.886200000000000
};

\addplot [ultra thick, yellow!50!black, mark=square, mark size=4, mark options={solid}]
table {%
2 0
3 0
4 0.000250000000000000
5 0.00119999999999999
};

\addplot [ultra thick, green, mark=triangle, mark size=4, mark options={solid}]
table {%
2 0.501500000000000
3 0.499000000000000
4 0.479750000000000
5 0.502400000000000
};
							
\addplot [ultra thick, black, mark=oplus, mark size=4, mark options={solid}]
table {%
2 0.899500000000000
3 0.882333333333333
4 0.868500000000000
5 0.858800000000000
};
\end{axis}

\end{tikzpicture}

%% file: BatchEM_wlf_vsC_NMI.tex
\begin{tikzpicture}

\definecolor{color0}{rgb}{0,0.75,0.75}

\begin{axis}[
legend cell align={left},
legend columns=3,
legend style={at={(0.5,1.05)}, anchor=north, draw=white!80.0!black},
tick align=outside,
tick pos=both,
title={},
x grid style={white!69.01960784313725!black},
xlabel={\Large Number of graphs $\mulG$},
xmajorgrids,
xmin=2, xmax=5,
xtick style={color=black},
y grid style={white!69.01960784313725!black},
ylabel={\Large NMI},
ymajorgrids,
ymin=0, ymax=1,
ytick style={color=black},
width=7cm,height=6cm,
]
\addplot [very thick, blue, mark=o, mark size=4, mark options={solid}]
table {%
2 0.999772812453221
3 0.990719632090800
4 0.934294350795343
5 0.930055496753029
};
\addplot [very thick, red, mark=o, mark size=4, mark options={solid}]
table {%
2 0.999367236045843
3 0.983210593036605
4 0.955611725547166
5 0.944724081399395
};
\addplot [very thick, color0, mark=o, mark size=4, mark options={solid}]
table {%
2 0.974739009400790
3 0.975636716911821
4 0.912836722497581
5 0.860679919048355
};
\addplot [very thick, green!50!black, mark=o, mark size=4, mark options={solid}]
table {%
2 0.954659733658390
3 0.912168046383270
4 0.787753443833495
5 0.662964081933621
};
\addplot [thick, orange, mark=otimes, mark size=4, mark options={solid}]
table {%
2 0.0665916256362695
3 0.0385832778283009
4 0.0366769945788574
5 0.0329016318502047
};

\addplot [thick, yellow!50!black, mark=square, mark size=4, mark options={solid}]
table {%
2 0.0120
3 0.0101
4 0.0090
5 0.0097
};

\end{axis}

\end{tikzpicture}

%% file: BatchEM_slf_vsC_NMI.tex
\begin{tikzpicture}

\definecolor{color0}{rgb}{0,0.75,0.75}

\begin{axis}[
legend cell align={left},
legend columns=3,
legend style={at={(0.5,1.05)}, anchor=north, draw=white!80.0!black},
tick align=outside,
tick pos=both,
title={},
x grid style={white!69.01960784313725!black},
xlabel={\Large Number of graphs $\mulG$},
xmajorgrids,
xmin=2, xmax=5,
xtick style={color=black},
y grid style={white!69.01960784313725!black},
ylabel={},
ymajorgrids,
ymin=0, ymax=1,
ytick style={color=black},
width=7cm,height=6cm,
]
\addplot [ultra thick, blue, mark=o, mark size=4, mark options={solid}]
table {%
2 1
3 1
4 1
5 1
};
\addplot [ultra thick, red, mark=o, mark size=4, mark options={solid}]
table {%
2 1
3 1
4 1
5 1
};
\addplot [ultra thick, color0, mark=o, mark size=4, mark options={solid}]
table {%
2 1
3 1
4 1
5 1
};
\addplot [ultra thick, green!50!black, mark=o, mark size=4, mark options={solid}]
table {%
2 1
3 1
4 1
5 1
};
\addplot [ultra thick, orange, mark=otimes, mark size=4, mark options={solid}]
table {%
2 0.975583033279172
3 0.906081687399239
4 0.903799924852085
5 0.913213740345692
};

\addplot [ultra thick, yellow!50!black, mark=square, mark size=4, mark options={solid}]
table {%
2 0.991526099798088
3 0.968995366729282
4 0.940577926413862
5 0.862013381338245
};

\end{axis}

\end{tikzpicture}

%% file: BatchEM_objf.tex
\begin{tikzpicture}

\definecolor{color0}{rgb}{0.967797559291991,0.441274560091574,0.53581031550587}
\definecolor{color1}{rgb}{0.680418912779335,0.615149751467757,0.194054521114453}
\definecolor{color2}{rgb}{0.201253172212011,0.690792081537903,0.479667611892753}
\definecolor{color3}{rgb}{0.219799566082832,0.662515787685034,0.773209315931721}

\begin{axis}[
legend cell align={left},
legend style={fill opacity=0.8, draw opacity=1, text opacity=1, at={(0.97,0.03)}, anchor=south east, draw=white!80!black},
tick align=outside,
tick pos=left,
x grid style={white!69.0196078431373!black},
xlabel={\Large Iteration $k$},
xmajorgrids,
xmin=0, xmax=100,
xtick style={color=black},
xtick={1,20,40,60,80,100},
xticklabels={$1$,$20$,$40$,$60$,$80$,$100$},
y grid style={white!69.0196078431373!black},
ylabel={\Large ${\cal L}(\Theta)$},
ymajorgrids,
ymin=-3.44954340365925, ymax=-0.258151071016631,
ytick style={color=black},
width=7cm,height=8cm,
]
\addplot [ultra thick, color0]
table {%
1 -2.09484063856831
2 -1.54683609359941
3 -1.19219484227772
4 -1.15837150673391
5 -1.15050263185936
6 -1.1470496987561
7 -1.11801955589636
8 -1.0979667746355
9 -1.01488445714017
10 -1.01482309295325
11 -1.01482832171489
12 -1.01481846057473
13 -1.01482255992346
14 -1.01481296955265
15 -1.01482719420888
16 -1.01462854716013
17 -1.0148248705645
18 -1.01482756829168
19 -1.01481624877196
20 -1.01476002459499
21 -1.01481730365104
22 -1.01482348231475
23 -1.01482530025878
24 -1.0148269136533
25 -1.01481798962536
26 -1.01482274638636
27 -1.01482570041999
28 -1.01482790118523
29 -1.01482462812934
30 -1.01478225008308
31 -1.01481896796695
32 -1.01480668065742
33 -1.01481280775546
34 -1.01482525126153
35 -1.01482472181792
36 -1.01482333477812
37 -1.01482258635527
38 -1.01479678389058
39 -1.0148168273044
40 -1.01482389248321
41 -1.01481883782774
42 -1.01482424068714
43 -1.01482371161406
44 -1.0148254971901
45 -1.01482201174287
46 -1.01480303125694
47 -1.01482306002245
48 -1.01577284126562
49 -1.01482983454469
50 -1.01480514587946
51 -1.01482732461965
52 -1.01477522269882
53 -1.0148190755355
54 -1.01481081988069
55 -1.0148283656201
56 -1.01482417696778
57 -1.01482791851176
58 -1.01480314761807
59 -1.01482685531018
60 -1.01482199508618
61 -1.01482654700909
62 -1.01482837245158
63 -1.01482811387327
64 -1.01483227846739
65 -1.01482136689874
66 -1.0148145364969
67 -1.01482367545697
68 -1.01482640823873
69 -1.014827665811
70 -1.0147987612258
71 -1.01478199701672
72 -1.01478385887439
73 -1.01479152970565
74 -1.0148209550649
75 -1.01480029955307
76 -1.0148198978196
77 -1.01482147723962
78 -1.01482406726881
79 -1.01475185610125
80 -1.01483434725
81 -1.01482417444329
82 -1.01482419982931
83 -1.01482461323834
84 -1.01480710619029
85 -1.01483734580738
86 -1.01481887858944
87 -1.01482388336085
88 -1.0148204574175
89 -1.01482348509821
90 -1.01474412566527
91 -1.01482284847854
92 -1.01482638396677
93 -1.01479811203453
94 -1.01481514504699
95 -1.01482095921501
96 -1.01482892627663
97 -1.01478097164668
98 -1.01482350093102
99 -1.01609669035663
100 -1.01482445290276
};
\addlegendentry{$\mulG=2$ $\gamma=0$ }
\addplot [ultra thick, dashed, color0]
table {%
1 -1.20591812536515
2 -1.09748242286848
3 -0.719073791355327
4 -0.520627217780824
5 -0.427746254193997
6 -0.407072135839418
7 -0.404838316933247
8 -0.404709183947445
9 -0.404700076029623
10 -0.40474246149932
11 -0.404774188145114
12 -0.404700251700067
13 -0.40442632607284
14 -0.403929336617954
15 -0.403505934407064
16 -0.403330396555754
17 -0.40329013137776
18 -0.40328102724749
19 -0.403275319652593
20 -0.403277140244215
21 -0.403668219136196
22 -0.403214358864023
23 -0.403266054183321
24 -0.403281625248599
25 -0.403275524208911
26 -0.403278693747184
27 -0.403277300864402
28 -0.403274070826327
29 -0.403282840862348
30 -0.403283338453307
31 -0.403273241167449
32 -0.403278055728745
33 -0.403277229212355
34 -0.403284090263475
35 -0.403292865131702
36 -0.403274551778121
37 -0.403276410382886
38 -0.403275044054866
39 -0.403279897565158
40 -0.403273081953741
41 -0.403277128451725
42 -0.403276388764848
43 -0.403277851466388
44 -0.403276211971538
45 -0.403279294783948
46 -0.403276026541799
47 -0.403281066401513
48 -0.403278819003678
49 -0.403274586051486
50 -0.403276504167868
51 -0.40327727974827
52 -0.403278316622554
53 -0.403279458585291
54 -0.403284657337262
55 -0.40327432499561
56 -0.40328384500053
57 -0.403274202148779
58 -0.403277175115931
59 -0.403290151757706
60 -0.403274133021854
61 -0.403266884979143
62 -0.403278161247737
63 -0.403292647459269
64 -0.403277233018824
65 -0.403273260331888
66 -0.403274762474772
67 -0.4032740416841
68 -0.40327708899083
69 -0.403277858312477
70 -0.40327642589554
71 -0.403275334796636
72 -0.403274976525653
73 -0.403275469730718
74 -0.403251037253277
75 -0.403265594205608
76 -0.403275807993618
77 -0.403270118534263
78 -0.403279591856424
79 -0.40328171899417
80 -0.403276349320934
81 -0.403275171146335
82 -0.403280872736774
83 -0.40327689010734
84 -0.403294320788559
85 -0.403269263773382
86 -0.403275960841239
87 -0.403274815985688
88 -0.403273386402063
89 -0.403280822370682
90 -0.403276661650533
91 -0.403285949562536
92 -0.403277044034358
93 -0.403273995060725
94 -0.403283983352829
95 -0.403282083354026
96 -0.403273912108468
97 -0.403279703079466
98 -0.403272575343617
99 -0.403275752442616
100 -0.403275862730163
};
\addlegendentry{$\mulG=2$ $\gamma=0.6$ }
\addplot [ultra thick, color1]
table {%
1 -2.55736055826364
2 -1.97463933392145
3 -1.12540874377137
4 -1.05092782964067
5 -1.05074205386582
6 -1.05074179000978
7 -1.0507400043468
8 -1.05073678812821
9 -1.05073999265187
10 -1.05074280725831
11 -1.05065438704159
12 -1.05074281153371
13 -1.0507315248084
14 -1.05073754962682
15 -1.05074141072783
16 -1.05074005054053
17 -1.05073549573167
18 -1.05073544127465
19 -1.05074254927513
20 -1.05073981838129
21 -1.05073829244775
22 -1.05074179483958
23 -1.05073932385742
24 -1.0507400545922
25 -1.05074184737954
26 -1.05073727395242
27 -1.0507358373341
28 -1.05073960988405
29 -1.05073419511901
30 -1.05073727455415
31 -1.05073054188961
32 -1.05073742635733
33 -1.05067464374511
34 -1.05074196134273
35 -1.05073873951442
36 -1.05073716961895
37 -1.05075910029236
38 -1.05073121176116
39 -1.05074164517517
40 -1.05073747604947
41 -1.05070328438837
42 -1.05073888811377
43 -1.0507294635497
44 -1.05074061819997
45 -1.05074222243019
46 -1.050740298525
47 -1.05074257843186
48 -1.05074051294127
49 -1.05073626667589
50 -1.05073587375241
51 -1.05072883182583
52 -1.05074130321707
53 -1.05074131022501
54 -1.05073946008644
55 -1.05074054778097
56 -1.05073586317483
57 -1.05073127828937
58 -1.0507417469429
59 -1.05073865153747
60 -1.05073698864586
61 -1.05073309812202
62 -1.05074137966768
63 -1.05074023768082
64 -1.05074477963383
65 -1.05074175277966
66 -1.05074205946811
67 -1.05074051277676
68 -1.05073350901672
69 -1.05073646987326
70 -1.05074118193768
71 -1.05073611455086
72 -1.05074177803726
73 -1.05074014707948
74 -1.05074253643744
75 -1.05074013843131
76 -1.05074095540484
77 -1.05074242855408
78 -1.0507341324795
79 -1.05073271435847
80 -1.05073712635985
81 -1.05073015109844
82 -1.05073823705451
83 -1.05074337516211
84 -1.05072829549274
85 -1.05072883567897
86 -1.05074358825996
87 -1.05073835289157
88 -1.05074067665172
89 -1.05073851827659
90 -1.05074009022787
91 -1.05074087770532
92 -1.05074437454621
93 -1.05073980489208
94 -1.05073853453634
95 -1.050741572248
96 -1.05074258640363
97 -1.05074139415736
98 -1.05074219009711
99 -1.05073754776353
100 -1.05074189421287
};
\addlegendentry{$\mulG=3$ $\gamma=0$ }
\addplot [ultra thick, dashed, color1]
table {%
1 -1.72006587260248
2 -1.51938583545567
3 -0.879592375898862
4 -0.733573146291047
5 -0.683102460647597
6 -0.649265311672472
7 -0.632168942184468
8 -0.604507014885359
9 -0.564201107023148
10 -0.517302129410271
11 -0.482110305654698
12 -0.468367339373421
13 -0.466938744673738
14 -0.46696487278364
15 -0.466787853834864
16 -0.466615938541347
17 -0.466338837173924
18 -0.466110830850231
19 -0.466026840335212
20 -0.465989227149577
21 -0.465982756776573
22 -0.465982979568283
23 -0.465985334516319
24 -0.465951109409384
25 -0.465974631549734
26 -0.465981598680844
27 -0.465978509613099
28 -0.465969058249628
29 -0.465978893313576
30 -0.465977866797901
31 -0.46597710104129
32 -0.465975037704594
33 -0.466011426480631
34 -0.465964110017845
35 -0.465888235750088
36 -0.465963547443986
37 -0.465974446245524
38 -0.465974729266437
39 -0.46597768095084
40 -0.465957344940396
41 -0.465969851786262
42 -0.465976975118874
43 -0.465976757179297
44 -0.465967192003349
45 -0.465975271653936
46 -0.46595993600857
47 -0.465975580643739
48 -0.465919347637855
49 -0.465974042770915
50 -0.465975548442504
51 -0.465973642117773
52 -0.466072583490025
53 -0.465953604703526
54 -0.465974177840102
55 -0.465983206049187
56 -0.465963092833848
57 -0.465983228696907
58 -0.46597296889247
59 -0.465971815914659
60 -0.465975457146822
61 -0.465992384564888
62 -0.465975162189818
63 -0.46597854397567
64 -0.46598158652147
65 -0.465977749520272
66 -0.465983534087917
67 -0.465977566685109
68 -0.46598101928199
69 -0.465977247105538
70 -0.465980175786864
71 -0.46597531880604
72 -0.465954278332664
73 -0.46594798150218
74 -0.46596750883735
75 -0.465989409007808
76 -0.465970803379178
77 -0.465970311143644
78 -0.465970904762946
79 -0.465970044329319
80 -0.465976242199761
81 -0.465980576949197
82 -0.4659739978188
83 -0.465994785700204
84 -0.465970821195156
85 -0.465979275063187
86 -0.465974618848391
87 -0.465978369200823
88 -0.465978036434019
89 -0.465984163187006
90 -0.465977662992738
91 -0.46597944763604
92 -0.465977935311231
93 -0.465978314158053
94 -0.465933897862879
95 -0.465948321143033
96 -0.465972646376996
97 -0.465970726223415
98 -0.465969127441076
99 -0.465973026259563
100 -0.466003834345774
};
\addlegendentry{$\mulG=3$ $\gamma=0.6$ }
\addplot [ultra thick, color2]
table {%
1 -3.04652603662276
2 -2.91380460162698
3 -1.56264676443837
4 -1.198201015746
5 -1.19661983040782
6 -1.19660483005685
7 -1.1966185034097
8 -1.19660500146847
9 -1.19661884328099
10 -1.1966183257245
11 -1.19662145946594
12 -1.19661880127299
13 -1.19661883649078
14 -1.19661879965097
15 -1.19661957587932
16 -1.19661958789087
17 -1.19661624108164
18 -1.19664325475437
19 -1.19661887116299
20 -1.19661839312876
21 -1.19661858756358
22 -1.19661856824101
23 -1.19662368757805
24 -1.19661934886725
25 -1.19661886011842
26 -1.19661923224181
27 -1.19661951372826
28 -1.19661822920662
29 -1.19662009508341
30 -1.19661907072665
31 -1.19661903621856
32 -1.19661840515446
33 -1.19661912943587
34 -1.19661849360476
35 -1.1966184747815
36 -1.19661842701884
37 -1.19661954125797
38 -1.19661409631634
39 -1.1966189296959
40 -1.19662348431147
41 -1.19661862352754
42 -1.19661820727975
43 -1.19662164795166
44 -1.1966190878896
45 -1.19661852941444
46 -1.19661943775494
47 -1.1966185945844
48 -1.1966184773276
49 -1.19662085827425
50 -1.19661868574094
51 -1.19661728589876
52 -1.19661876224147
53 -1.19661967756739
54 -1.19661839051739
55 -1.19661850321704
56 -1.1966185812908
57 -1.19661860306356
58 -1.19661968568252
59 -1.19661893244146
60 -1.19661896353764
61 -1.19661849609851
62 -1.19660610336313
63 -1.19662048383801
64 -1.19661848416385
65 -1.19661863168283
66 -1.1966182170176
67 -1.19661870977172
68 -1.19661857684096
69 -1.19662496131842
70 -1.19661318721468
71 -1.19661800175721
72 -1.1966217316845
73 -1.19661820399572
74 -1.19661926201586
75 -1.19661542397676
76 -1.19661856806313
77 -1.19661877910822
78 -1.19661835842177
79 -1.19661970424757
80 -1.19663620529229
81 -1.19661872382151
82 -1.19661852774657
83 -1.19661830726721
84 -1.19663323336407
85 -1.19661851147847
86 -1.19661953387781
87 -1.1966183836195
88 -1.19661853584146
89 -1.19661918018794
90 -1.19661996783175
91 -1.19661819696567
92 -1.19662053805656
93 -1.19661844649061
94 -1.19661999352152
95 -1.1966192144951
96 -1.19661870333128
97 -1.19661865020821
98 -1.1966185530408
99 -1.19661725071906
100 -1.19662306586938
};
\addlegendentry{$\mulG=4$ $\gamma=0$ }
\addplot [ultra thick, dashed, color2]
table {%
1 -2.0390829648509
2 -1.95710846142564
3 -1.37583083966088
4 -0.865578705699241
5 -0.729513795928674
6 -0.635985763409818
7 -0.580411715810353
8 -0.522409294207631
9 -0.506510987750445
10 -0.500571260768398
11 -0.49866447079326
12 -0.498433252948347
13 -0.498340059366186
14 -0.498267142633577
15 -0.498265255672237
16 -0.498264770005819
17 -0.498272585917292
18 -0.498300052464604
19 -0.498312112043619
20 -0.498429922242885
21 -0.49832707947988
22 -0.49835252095512
23 -0.498378189697
24 -0.498365176572587
25 -0.498450171909398
26 -0.498389087540623
27 -0.498408723697987
28 -0.498418906868301
29 -0.498432462506183
30 -0.498414552867256
31 -0.498382271571645
32 -0.498299346220863
33 -0.498178607327521
34 -0.498024696260373
35 -0.497851046576853
36 -0.497807781818932
37 -0.497824375793725
38 -0.497816390760341
39 -0.497831592671808
40 -0.49785215562471
41 -0.497832023037962
42 -0.497796862732164
43 -0.497846058747186
44 -0.497851954594523
45 -0.497846187405811
46 -0.497849147190471
47 -0.497852522023775
48 -0.497841952651148
49 -0.497847296032722
50 -0.497846540548564
51 -0.497856337892477
52 -0.497926474676138
53 -0.497834627305139
54 -0.497848858607775
55 -0.497941671001559
56 -0.497812014561064
57 -0.497891699328418
58 -0.497826569617972
59 -0.497845611306926
60 -0.497854812325868
61 -0.497851810979636
62 -0.497847212506096
63 -0.49786909452617
64 -0.497820216905063
65 -0.497842808236275
66 -0.497881438792285
67 -0.497865074678264
68 -0.497840422919622
69 -0.497836836569507
70 -0.497823882991009
71 -0.497830553863657
72 -0.49785554860087
73 -0.497855774473803
74 -0.49784183226285
75 -0.497857552977621
76 -0.497844821158087
77 -0.497841188792343
78 -0.497862172860629
79 -0.497823257434695
80 -0.497833906710223
81 -0.497851363524833
82 -0.497837065349576
83 -0.497850078461582
84 -0.497844045214699
85 -0.497860149392346
86 -0.497859356281108
87 -0.497852856943534
88 -0.497849330082427
89 -0.497849803446506
90 -0.497848730975788
91 -0.497860718375872
92 -0.497839135358542
93 -0.497839746496634
94 -0.497836380072846
95 -0.497847225279431
96 -0.497854632269235
97 -0.497809944610014
98 -0.497808092963971
99 -0.497840525109397
100 -0.497845029173756
};
\addlegendentry{$\mulG=4$ $\gamma=0.6$ }
\addplot [ultra thick, color3]
table {%
1 -3.30448011581186
2 -2.9415789747639
3 -1.55090130390734
4 -1.35413941539638
5 -1.30167108230826
6 -1.25269489504307
7 -1.2338022173905
8 -1.22292110683468
9 -1.18683165176125
10 -1.13723353175612
11 -1.12849447183726
12 -1.12848016603915
13 -1.12849423177757
14 -1.12848708997681
15 -1.12848858394084
16 -1.12858827280384
17 -1.1284881720283
18 -1.12849896414758
19 -1.12849123982653
20 -1.1284857599575
21 -1.12847785315502
22 -1.12848912321532
23 -1.12850392436672
24 -1.1284940143502
25 -1.12849158775471
26 -1.12849857597834
27 -1.12848464036147
28 -1.12846481879311
29 -1.12850009025618
30 -1.12849905530002
31 -1.12849658290753
32 -1.12849249786402
33 -1.12850066595843
34 -1.12848134271828
35 -1.12875939662182
36 -1.1284839818414
37 -1.12848873150444
38 -1.12848630795189
39 -1.12850228757968
40 -1.1284911118509
41 -1.12848927338193
42 -1.1284548688077
43 -1.12848613384188
44 -1.12848930591863
45 -1.12848501151836
46 -1.12851041936832
47 -1.12845047995324
48 -1.12845287792459
49 -1.12848461194677
50 -1.12849772803713
51 -1.12847860253526
52 -1.12849718187199
53 -1.1284914678714
54 -1.12846652574152
55 -1.12848931346754
56 -1.12842688625584
57 -1.12847727446598
58 -1.12849892650825
59 -1.12847965962223
60 -1.12846432212696
61 -1.12849474666567
62 -1.12849473080994
63 -1.12849137783413
64 -1.12847525032502
65 -1.12850452638953
66 -1.12839787192216
67 -1.1284174269983
68 -1.12849382951491
69 -1.12848384547941
70 -1.12849268247864
71 -1.12835465272662
72 -1.12850914135047
73 -1.12850160980897
74 -1.12849643929231
75 -1.12850108829506
76 -1.1285036165409
77 -1.12847344167195
78 -1.12845202226099
79 -1.12849724961908
80 -1.12848925261468
81 -1.12849767007137
82 -1.12849515382202
83 -1.12849509964939
84 -1.12849545198254
85 -1.12849178109315
86 -1.12849923292919
87 -1.12849071045195
88 -1.12849922815813
89 -1.12849710722955
90 -1.12847931875203
91 -1.12850043244712
92 -1.12849200817744
93 -1.12851320651383
94 -1.12842241631658
95 -1.12849687576993
96 -1.12830749794926
97 -1.12852086821007
98 -1.12850160278632
99 -1.12849626582766
100 -1.12850068573837
};
\addlegendentry{$\mulG=5$ $\gamma=0$ }
\addplot [ultra thick, dashed, color3]
table {%
1 -2.27952533257444
2 -2.21529865918752
3 -1.54167608687909
4 -0.799202320191983
5 -0.656999390122632
6 -0.598665849834386
7 -0.555210736675907
8 -0.543564564219747
9 -0.535075154961315
10 -0.533603827914599
11 -0.531804701909559
12 -0.530988694968331
13 -0.530367740311727
14 -0.529510353366858
15 -0.529207606897754
16 -0.529166625563449
17 -0.529160337264159
18 -0.529158696052884
19 -0.529116580603828
20 -0.529075881819453
21 -0.529027041834221
22 -0.529045845493675
23 -0.52888588322033
24 -0.528860837718152
25 -0.528840850266707
26 -0.528863497710055
27 -0.528847877845986
28 -0.528846423854618
29 -0.528854307792431
30 -0.528851522410754
31 -0.52886228281057
32 -0.528871445739236
33 -0.528869244768575
34 -0.528874927054374
35 -0.528888990795203
36 -0.528862437281088
37 -0.528836474307342
38 -0.528862023050363
39 -0.52889172671086
40 -0.528841129711808
41 -0.52887489555021
42 -0.528870239952505
43 -0.528879748691076
44 -0.528877773124211
45 -0.528906135513554
46 -0.528852422514084
47 -0.52887468314275
48 -0.528883729894943
49 -0.528874235717553
50 -0.528879991562946
51 -0.528880919650235
52 -0.528884160978998
53 -0.528868916502787
54 -0.528885123287269
55 -0.528881114713367
56 -0.528887508904122
57 -0.528881972497573
58 -0.528947863981871
59 -0.529406379675053
60 -0.530029286743544
61 -0.528453563520307
62 -0.528737231832409
63 -0.528778379985679
64 -0.52882899874795
65 -0.528846737022916
66 -0.528909249577719
67 -0.528908755566157
68 -0.528869410139276
69 -0.528894162049974
70 -0.528860183119079
71 -0.528877369698819
72 -0.528887275559134
73 -0.528872268988563
74 -0.52886626752089
75 -0.528898694615906
76 -0.528875271445958
77 -0.528876495156671
78 -0.528873418046155
79 -0.528900124373226
80 -0.528865944100165
81 -0.528885126105839
82 -0.528878585348227
83 -0.528886862520338
84 -0.528870287580308
85 -0.528875590970629
86 -0.528889783551186
87 -0.528880702121852
88 -0.528923034445611
89 -0.528846205223557
90 -0.528874843753378
91 -0.528866680999627
92 -0.528871097519468
93 -0.528875495425387
94 -0.52887839761058
95 -0.528883995516303
96 -0.528884918882949
97 -0.528905392886371
98 -0.528860993386604
99 -0.52888067946759
100 -0.528873148551748
};
\addlegendentry{$\mulG=5$ $\gamma=0.6$ }
\end{axis}

\end{tikzpicture}

%% file: BatchEM_slp_objf.tex
\begin{tikzpicture}

\definecolor{color0}{rgb}{0.967797559291991,0.441274560091574,0.53581031550587}
\definecolor{color1}{rgb}{0.680418912779335,0.615149751467757,0.194054521114453}
\definecolor{color2}{rgb}{0.201253172212011,0.690792081537903,0.479667611892753}
\definecolor{color3}{rgb}{0.219799566082832,0.662515787685034,0.773209315931721}

\begin{axis}[
legend cell align={left},
legend style={fill opacity=0.8, draw opacity=1, text opacity=1, at={(0.97,0.03)}, anchor=south east, draw=white!80!black},
tick align=outside,
tick pos=left,
x grid style={white!69.0196078431373!black},
xlabel={\Large Iteration $k$},
xmajorgrids,
xmin=0, xmax=100,
xtick style={color=black},
xtick={1,20,40,60,80,100},
xticklabels={$1$,$20$,$40$,$60$,$80$,$100$},
y grid style={white!69.0196078431373!black},
ylabel={},
ymajorgrids,
ymin=-36.0989247458917, ymax=-17.4211947501405,
ytick style={color=black},
width=7cm,height=8cm,
]
\addplot [ultra thick, color0]
table {%
1 -24.093054041259
2 -21.132717539645
3 -21.1324231824284
4 -21.132179754398
5 -21.1324496168363
6 -21.1325438838221
7 -21.1324392609682
8 -21.132679442613
9 -21.1324160814078
10 -21.1324785159046
11 -21.132487975597
12 -21.1325007406779
13 -21.13261200967
14 -21.1325559395577
15 -21.1326791281662
16 -21.1323308835393
17 -21.1326300009063
18 -21.1325421436669
19 -21.1326376541042
20 -21.1322567459188
21 -21.1325544493048
22 -21.1320764265102
23 -21.1326796482533
24 -21.1325174830443
25 -21.1322850751623
26 -21.13989770559
27 -21.1318954006192
28 -21.1325317608352
29 -21.1316543985182
30 -21.1325822716142
31 -21.1325216067119
32 -21.1326484742848
33 -21.1323577170278
34 -21.1320457007159
35 -21.1326076209577
36 -21.132685295491
37 -21.1323515561202
38 -21.1328304658536
39 -21.1325447209323
40 -21.1331464286501
41 -21.1323914406489
42 -21.1323837484273
43 -21.1325457815122
44 -21.1327354899676
45 -21.1324158615319
46 -21.1325049242251
47 -21.1324321829374
48 -21.1326322063948
49 -21.1450748109934
50 -21.1323538878469
51 -21.1326380521796
52 -21.132340697697
53 -21.1325152213165
54 -21.129763305801
55 -21.1328975490855
56 -21.133182531255
57 -21.1320767178492
58 -21.1316931832716
59 -21.1325364330201
60 -21.1321022697037
61 -21.132512199103
62 -21.1312754768426
63 -21.1325506536458
64 -21.1324635076895
65 -21.1320486472394
66 -21.1325060722857
67 -21.1321775344777
68 -21.1324548368358
69 -21.1325740641702
70 -21.1326813014088
71 -21.132455441573
72 -21.1320465118439
73 -21.1325927821544
74 -21.1326358111892
75 -21.1322074650253
76 -21.1324542335442
77 -21.1330233770546
78 -21.1352358985358
79 -21.1322330426478
80 -21.1325015242793
81 -21.1320765592333
82 -21.1326474514247
83 -21.1313374889313
84 -21.1308236497151
85 -21.1324608376154
86 -21.1325705135357
87 -21.1325756757352
88 -21.1325713464879
89 -21.1326857444058
90 -21.1324319856936
91 -21.1324177747522
92 -21.1316678333887
93 -21.1296091094373
94 -21.1318644765741
95 -21.1296487542929
96 -21.1319534560176
97 -21.1314291297304
98 -21.1324221962421
99 -21.1322423495623
100 -21.1322438559114
};
\addplot [ultra thick, dashed, color0]
table {%
1 -21.1017167234141
2 -18.4193939318596
3 -18.2705796182396
4 -18.2708992866792
5 -18.2705663514811
6 -18.2713439765375
7 -18.2709492718634
8 -18.2711179785275
9 -18.2710069976295
10 -18.2705708551783
11 -18.2707320052108
12 -18.2705226607193
13 -18.2705362359357
14 -18.2710937175151
15 -18.2709160342329
16 -18.2725271718474
17 -18.2705753781491
18 -18.2752121343207
19 -18.2706329952338
20 -18.2705947422585
21 -18.2705775949186
22 -18.2701824772201
23 -18.276833550429
24 -18.2705735150738
25 -18.2718149650148
26 -18.2753063756046
27 -18.2714006656635
28 -18.2703813895726
29 -18.2706106372419
30 -18.271077711231
31 -18.2706463876656
32 -18.2704747312357
33 -18.2708967810674
34 -18.2706140509347
35 -18.2705386777274
36 -18.2727034042216
37 -18.2705769069915
38 -18.2888860601779
39 -18.271028272589
40 -18.2706219539929
41 -18.2704525056441
42 -18.271913606775
43 -18.2705870887796
44 -18.2708179376654
45 -18.2717721783191
46 -18.2716782925407
47 -18.2709152267571
48 -18.2755384401321
49 -18.2728808200702
50 -18.2730974091109
51 -18.2705340573153
52 -18.2706251111455
53 -18.2704702491914
54 -18.2704288555682
55 -18.2709770727833
56 -18.2704687731847
57 -18.270730871437
58 -18.271076823859
59 -18.2717805204722
60 -18.2717682206634
61 -18.2705270987181
62 -18.270575967241
63 -18.2736199726089
64 -18.2704772166478
65 -18.2710220017951
66 -18.2705093821006
67 -18.2706739204896
68 -18.2707689050198
69 -18.2704582649126
70 -18.2712484392132
71 -18.2711380460898
72 -18.2710401081509
73 -18.2707837464433
74 -18.2704323168175
75 -18.2716097634605
76 -18.270465013947
77 -18.2705103061795
78 -18.2705709742328
79 -18.2705060942825
80 -18.2708019572415
81 -18.2704269999721
82 -18.2710591528441
83 -18.2705554491563
84 -18.2710370632826
85 -18.270413867002
86 -18.2706987517517
87 -18.2704973350464
88 -18.2705489618934
89 -18.2711268814754
90 -18.273551350813
91 -18.270677740205
92 -18.2716278824138
93 -18.2721630409605
94 -18.2707083100971
95 -18.2707588615909
96 -18.2712556575935
97 -18.2706636869614
98 -18.2703892188901
99 -18.2704951073639
100 -18.2704532864496
};
\addplot [ultra thick, color1]
table {%
1 -26.1842388587704
2 -21.950851151446
3 -21.9544692447711
4 -21.9500565908399
5 -21.953561655838
6 -21.9499580526699
7 -21.9502690118201
8 -21.950814609036
9 -21.9500959485299
10 -21.9509485931498
11 -21.9506105972607
12 -21.9507905611703
13 -21.9823984530097
14 -21.9526723735159
15 -21.9531972907467
16 -21.9502121399305
17 -21.9516346320012
18 -21.9499421002692
19 -21.9504349235606
20 -21.9501351158624
21 -21.952336444566
22 -21.9560990409164
23 -21.9500736524164
24 -21.9498321756286
25 -21.9515533322147
26 -21.9508510823966
27 -21.9504540780327
28 -21.9512896309698
29 -21.9511608682884
30 -21.9503380138745
31 -21.9516093124903
32 -21.9499628236502
33 -21.9502325099187
34 -21.9516690656932
35 -21.9504650889542
36 -21.951570790131
37 -21.9585708673344
38 -21.9509108633411
39 -21.9511329789
40 -21.9516610518381
41 -21.9506451684931
42 -21.9509396673706
43 -21.9509444297651
44 -21.9567788837809
45 -21.9513224552049
46 -21.9506255365859
47 -21.9499505776984
48 -21.9529141629467
49 -21.951161905903
50 -21.9502870408896
51 -21.9507030227329
52 -21.9479450614003
53 -21.9504098746325
54 -21.9502030767901
55 -21.9500505491937
56 -21.950227255631
57 -21.9508591877859
58 -21.9504905323804
59 -21.9512892183598
60 -21.9508152057642
61 -21.9504475594548
62 -21.9502898791568
63 -21.9596217289032
64 -21.9530919471794
65 -21.9502525476488
66 -21.9527084100585
67 -21.9543075258752
68 -21.9506044515627
69 -21.9537290596554
70 -21.9523248429019
71 -21.9547050217215
72 -21.9552655744164
73 -21.9531981190387
74 -21.9503973577064
75 -21.9501106196884
76 -21.9501428994365
77 -21.9504066188347
78 -21.9525079554655
79 -21.951902937984
80 -21.9512362577242
81 -21.950484277208
82 -21.9504042318904
83 -21.9512182917373
84 -21.9507026342661
85 -21.9503084247242
86 -21.9593490879261
87 -21.9509568417661
88 -21.9510596031191
89 -21.9507626189413
90 -21.9501480346201
91 -21.949998662548
92 -21.9531189288554
93 -21.9504201845453
94 -21.9505114863526
95 -21.9505809655467
96 -21.9504730547555
97 -21.9501078917789
98 -21.9503617884256
99 -21.9579368875672
100 -21.9501228223792
};
\addplot [ultra thick, dashed, color1]
table {%
1 -21.8515558039513
2 -18.7047119081979
3 -18.7391150992437
4 -18.7381104999185
5 -18.7369278787488
6 -18.7662313981474
7 -18.7444293414976
8 -18.7369118772113
9 -18.7371749834783
10 -18.7695069352812
11 -18.7366899026191
12 -18.7411045654196
13 -18.7368448252255
14 -18.7411925438731
15 -18.7366606634266
16 -18.7402004558406
17 -18.7388987739534
18 -18.7372121646341
19 -18.7407163201202
20 -18.7730726899481
21 -18.7425544686555
22 -18.7374797150539
23 -18.7375293059228
24 -18.7367659851471
25 -18.7373971089525
26 -18.7373612083352
27 -18.7480550498721
28 -18.7366559669021
29 -18.7368343325362
30 -18.7372334610622
31 -18.7368260816008
32 -18.7372961011451
33 -18.7360198660048
34 -18.7368750368651
35 -18.7390723351025
36 -18.7370718462568
37 -18.7390022400781
38 -18.7365753905054
39 -18.7371571276512
40 -18.7429098688581
41 -18.7402453021115
42 -18.7370791245398
43 -18.7370588158658
44 -18.7389567136284
45 -18.7374675696414
46 -18.7384541592075
47 -18.7371641150918
48 -18.7378500146793
49 -18.7443299401428
50 -18.7368900191226
51 -18.7465704254238
52 -18.7368933302461
53 -18.7366299293031
54 -18.7378190016041
55 -18.7391844902089
56 -18.7379157708679
57 -18.7373615252177
58 -18.7381513385576
59 -18.7394685804498
60 -18.7388323519
61 -18.7371918069956
62 -18.7369999512254
63 -18.73677812999
64 -18.7381432572593
65 -18.7368180882408
66 -18.7366943070428
67 -18.7368508899813
68 -18.7379246814597
69 -18.7387868797331
70 -18.7389557651476
71 -18.739078568945
72 -18.73686616746
73 -18.7377716585561
74 -18.7374353152481
75 -18.7375982486033
76 -18.7378477066322
77 -18.7368543083281
78 -18.7367530449181
79 -18.7372109305382
80 -18.7364298597077
81 -18.7389812069179
82 -18.7371027567836
83 -18.7388365202577
84 -18.7369915064756
85 -18.7369315176727
86 -18.7379531858429
87 -18.7378265221815
88 -18.7385867385994
89 -18.7390321711446
90 -18.7373462515205
91 -18.7373609559718
92 -18.7374034175176
93 -18.7378708671782
94 -18.7415559686627
95 -18.7369823415858
96 -18.7377491748361
97 -18.7371496942708
98 -18.737375707695
99 -18.7370909580815
100 -18.7373506263668
};
\addplot [ultra thick, color2]
table {%
1 -33.3499133908772
2 -27.2046070621256
3 -27.1607334912455
4 -27.160028032158
5 -27.160799858971
6 -27.1605772959963
7 -27.1620008900118
8 -27.1604075164494
9 -27.1601864581273
10 -27.1612563038236
11 -27.1616280097207
12 -27.1603202718431
13 -27.1613620137654
14 -27.1600531963438
15 -27.1599760556254
16 -27.1607238042976
17 -27.1885293850171
18 -27.1612151002503
19 -27.1598571111201
20 -27.1605537034342
21 -27.1606211894443
22 -27.1604295431127
23 -27.1597185333842
24 -27.1617720253711
25 -27.1618556902842
26 -27.1801093768293
27 -27.1614522026994
28 -27.1597916136717
29 -27.1613608160376
30 -27.1602535642159
31 -27.1617579120636
32 -27.1600542769062
33 -27.1603486932598
34 -27.1679676992644
35 -27.1635863758928
36 -27.2327128245736
37 -27.1618182627817
38 -27.1602931354234
39 -27.1660111430695
40 -27.1603297945516
41 -27.1618610526035
42 -27.1610826816082
43 -27.1607138608603
44 -27.1601905164742
45 -27.160285076057
46 -27.1613649966782
47 -27.1625891445672
48 -27.1601689755562
49 -27.1603623372449
50 -27.1611225319949
51 -27.16023077387
52 -27.1638737142726
53 -27.1602251896863
54 -27.1613984447065
55 -27.1603907608471
56 -27.1636585967844
57 -27.160919093983
58 -27.1621952799994
59 -27.1596606286372
60 -27.1606618041174
61 -27.1643735834102
62 -27.1604779023366
63 -27.1613525033317
64 -27.1601246538296
65 -27.1611317970376
66 -27.1604011495413
67 -27.1621522525854
68 -27.1589956259574
69 -27.1601156792896
70 -27.1614769212081
71 -27.1610556211161
72 -27.1605987800497
73 -27.1648754137673
74 -27.1608508024777
75 -27.1606833822195
76 -27.1644567715205
77 -27.2283983426606
78 -27.1611054652495
79 -27.1661034159102
80 -27.1599901671647
81 -27.1622310178867
82 -27.1597148127284
83 -27.1600950172201
84 -27.1636726262243
85 -27.1701638577425
86 -27.1622850692816
87 -27.1608312515913
88 -27.1604159604542
89 -27.1613542517867
90 -27.1614523872535
91 -27.1597830292706
92 -27.1607036904797
93 -27.1606296144898
94 -27.1613367760882
95 -27.1619224969219
96 -27.1602077069941
97 -27.1593478542183
98 -27.1612853031868
99 -27.1603430141202
100 -27.1608072707883
};
\addplot [ultra thick, dashed, color2]
table {%
1 -27.0754392601923
2 -25.2298959152268
3 -25.0323415401043
4 -25.0327488091835
5 -25.0316165565044
6 -25.0314702003291
7 -25.0314725481388
8 -25.0335504341078
9 -25.0337129346441
10 -25.034777242517
11 -25.032127167038
12 -25.0327338700527
13 -25.033698720175
14 -25.0390543676027
15 -25.0313355122071
16 -25.0314892790117
17 -25.0321502680613
18 -25.0314089077384
19 -25.0346125835091
20 -25.0314251542654
21 -25.0317027835181
22 -25.0350954828778
23 -25.032484158493
24 -25.0313242243793
25 -25.031642070356
26 -25.043050672652
27 -25.0322454337844
28 -25.0542769457184
29 -25.0316837632274
30 -25.0320085613737
31 -25.0329559833689
32 -25.0318375314158
33 -25.0922140671874
34 -25.0321405190051
35 -25.0320269005879
36 -25.032158436277
37 -25.0317237151853
38 -25.0325586510042
39 -25.0319501683724
40 -25.0327342143451
41 -25.0313959490525
42 -25.0322038958229
43 -25.0317378439156
44 -25.0315597336
45 -25.0316647634701
46 -25.0319770533601
47 -25.0326471728939
48 -25.0366900243289
49 -25.0318735552179
50 -25.0319047078727
51 -25.0317341031663
52 -25.0317399895165
53 -25.0313701384207
54 -25.0370618270248
55 -25.0321446905326
56 -25.0337483908167
57 -25.0321852829445
58 -25.0322940132571
59 -25.0314274929403
60 -25.0339541912898
61 -25.0317597249233
62 -25.0416490970781
63 -25.0508644438796
64 -25.1097579019052
65 -25.0318291903894
66 -25.0336301854499
67 -25.0316939369075
68 -25.0483397352386
69 -25.0328795186238
70 -25.0320226507254
71 -25.0317383265753
72 -25.0320055977586
73 -25.0332231833446
74 -25.03277874076
75 -25.0312660688962
76 -25.0406588823872
77 -25.0315179836432
78 -25.0317897415558
79 -25.0328037531627
80 -25.032359354963
81 -25.0328835068277
82 -25.0321785725409
83 -25.0317003134174
84 -25.0318878118534
85 -25.031822710742
86 -25.0318776940609
87 -25.0332890901062
88 -25.0330037859928
89 -25.0313666197617
90 -25.0321706390337
91 -25.0321358375673
92 -25.0342632346406
93 -25.0346120917556
94 -25.031808969874
95 -25.0314873630005
96 -25.0365697703464
97 -25.0324770301241
98 -25.033242930957
99 -25.0318628000659
100 -25.0315249457936
};
\addplot [ultra thick, color3]
table {%
1 -35.2499370188121
2 -31.5611946036797
3 -29.248434811537
4 -28.9763720586756
5 -28.9764544049214
6 -28.9830344390109
7 -28.9763020599564
8 -28.9777405406046
9 -28.9752880666664
10 -28.9761473469041
11 -28.9771586647036
12 -28.9771612874405
13 -28.9789513262626
14 -28.9793149207105
15 -28.976339263733
16 -28.9808310036962
17 -28.975448161152
18 -28.9793150807408
19 -28.994845143341
20 -28.9756684027329
21 -28.976523043692
22 -28.9768532656523
23 -28.9768355515924
24 -28.9771573083133
25 -28.9779644871221
26 -28.9766160477985
27 -28.9809042967322
28 -28.9761642802326
29 -28.978748638016
30 -28.9771167142341
31 -28.9850506045318
32 -28.9758526302313
33 -28.9858003435994
34 -28.9770710796048
35 -28.9760595445906
36 -28.9773899471556
37 -28.9759634918662
38 -28.9764401780809
39 -28.9760756506858
40 -28.9759238219998
41 -28.9788188915839
42 -28.9778969882371
43 -28.9777711034369
44 -28.9793295115263
45 -28.9802032004755
46 -28.9776976082663
47 -28.9763119370512
48 -28.9760212538733
49 -28.9776869178847
50 -28.9880464146963
51 -28.9762541038918
52 -28.9794166073579
53 -28.9775389761019
54 -28.980372397117
55 -28.9794370929906
56 -28.9777921276535
57 -28.9775082562458
58 -28.9773354829866
59 -28.9789377568899
60 -28.9792848287709
61 -28.9766652273927
62 -28.9760448116511
63 -28.9766461797778
64 -28.9762360972722
65 -28.9758989970464
66 -28.9873350737671
67 -28.9768424525724
68 -28.9772547518154
69 -28.9781927519151
70 -28.9823936582351
71 -28.9767521540678
72 -28.9766151105284
73 -28.9761123632541
74 -28.9758597873603
75 -28.9769913276116
76 -28.9763034582049
77 -28.9765062873766
78 -28.9775071599351
79 -28.9760361491395
80 -28.9787287886664
81 -28.984834156756
82 -28.9765276928473
83 -28.97803565146
84 -28.977482185751
85 -28.9801669889472
86 -28.9779189900247
87 -28.977085691554
88 -28.9766527199299
89 -28.9803095442701
90 -28.9784317330089
91 -28.9842434662259
92 -28.9763982559591
93 -28.9759860316211
94 -28.9763071157559
95 -28.9777233259275
96 -28.9814244465902
97 -28.9766801236488
98 -28.9800341428409
99 -28.9801672839796
100 -28.9762612974251
};
\addplot [ultra thick, dashed, color3]
table {%
1 -34.1625269289383
2 -31.0578695643488
3 -29.4548077307657
4 -29.4564361745315
5 -29.4541891562517
6 -29.4539141051539
7 -29.4539285555021
8 -29.457948985167
9 -29.4559143925002
10 -29.4543131978803
11 -29.4555888245305
12 -29.4607392493683
13 -29.4555386914944
14 -29.4543330205892
15 -29.4545247179827
16 -29.4568935471364
17 -29.4585340448054
18 -29.455460148755
19 -29.4553825125586
20 -29.454147523183
21 -29.455212141079
22 -29.4565838006834
23 -29.4532657742584
24 -29.4575335595113
25 -29.4542666634261
26 -29.4541603736112
27 -29.4567213796274
28 -29.4552597743881
29 -29.4562235746907
30 -29.4567140191032
31 -29.4551925290015
32 -29.4532380439027
33 -29.457656569547
34 -29.4571013523533
35 -29.4568044624431
36 -29.4533807640939
37 -29.4558884376568
38 -29.455145779762
39 -29.4542929304223
40 -29.4551959381103
41 -29.4553030260336
42 -29.4874497714375
43 -29.4541049416007
44 -29.4564369685982
45 -29.4540183467314
46 -29.4543731639167
47 -29.4554807105742
48 -29.4747354067201
49 -29.6873237665894
50 -29.5890916749966
51 -29.4556705017465
52 -29.4550271834608
53 -29.457058757699
54 -29.4573188731663
55 -29.4561556216305
56 -29.5413179801234
57 -29.4692265153423
58 -29.4533829224255
59 -29.4554960756166
60 -29.4571487553685
61 -29.4579381333911
62 -29.4536436514881
63 -29.4535907792489
64 -29.4538810036904
65 -29.456836634027
66 -29.454072315032
67 -29.4544836178316
68 -29.4535903087592
69 -29.4538952358293
70 -29.4544640009726
71 -29.4559433904947
72 -29.4540125138834
73 -29.4553162488068
74 -29.454538605098
75 -29.4575895760555
76 -29.4571147109732
77 -29.4567860769984
78 -29.4536997109457
79 -29.4665746426821
80 -29.4566560682169
81 -29.4558408427188
82 -29.4543719737271
83 -29.4727655345195
84 -29.4560901802913
85 -29.4537968286028
86 -29.4596844953771
87 -29.472010586915
88 -29.4545830009514
89 -29.4538831050561
90 -29.4546448179358
91 -29.4536878977521
92 -29.4555735190727
93 -29.4652187025149
94 -29.5015887270772
95 -29.457729121399
96 -29.4568744021988
97 -29.4563885323421
98 -29.4538817339601
99 -29.4529178345414
100 -29.4554160941515
};
\end{axis}

\end{tikzpicture}

%% file: onlineEM_vsm_Errorrate.tex
\begin{tikzpicture}

\definecolor{color0}{rgb}{0.967797559291991,0.441274560091574,0.53581031550587}
\definecolor{color1}{rgb}{0.680418912779335,0.615149751467757,0.194054521114453}
\definecolor{color2}{rgb}{0.201253172212011,0.690792081537903,0.479667611892753}
\definecolor{color3}{rgb}{0.219799566082832,0.662515787685034,0.773209315931721}

\begin{axis}[
legend cell align={left},
legend style={fill opacity=0.8, draw opacity=1, text opacity=1, draw=white!80!black},
tick align=outside,
tick pos=left,
x grid style={white!69.0196078431373!black},
xlabel={\Large Time $t$},
xmajorgrids,
xmin=1, xmax=11,
xtick style={color=black},
xtick={1,2,4,6,8,10},
xticklabels={$0$,$300\mulG$,$900\mulG$,$1500\mulG$,$2100\mulG$,$2700\mulG$},
y grid style={white!69.0196078431373!black},
ylabel={\Large Error rate },
ymajorgrids,
ymin=0.00640197207455656, ymax=0.507687990353923,
ytick style={color=black},
width=7cm,height=6cm,
]
\path [fill=color0, fill opacity=0.7]
(axis cs:1,0.354477760372027)
--(axis cs:1,0.348855572961306)
--(axis cs:2,0.16846278271069)
--(axis cs:3,0.142966623401267)
--(axis cs:4,0.11999127640308)
--(axis cs:5,0.111875835281489)
--(axis cs:6,0.106988002850928)
--(axis cs:7,0.102030091825272)
--(axis cs:8,0.0993347450341577)
--(axis cs:9,0.0976941780320094)
--(axis cs:10,0.0938150558935051)
--(axis cs:11,0.0927385091397302)
--(axis cs:11,0.0950392686380475)
--(axis cs:11,0.0950392686380475)
--(axis cs:10,0.0961849441064948)
--(axis cs:9,0.100083599745768)
--(axis cs:8,0.101776366076953)
--(axis cs:7,0.104636574841394)
--(axis cs:6,0.109678663815738)
--(axis cs:5,0.114790831385178)
--(axis cs:4,0.123342056930253)
--(axis cs:3,0.147033376598733)
--(axis cs:2,0.173759439511532)
--(axis cs:1,0.354477760372027)
--cycle;

\path [fill=color1, fill opacity=0.7]
(axis cs:1,0.377475493204443)
--(axis cs:1,0.373062141204159)
--(axis cs:2,0.167031738537326)
--(axis cs:3,0.152373306110201)
--(axis cs:4,0.131088438421652)
--(axis cs:5,0.117153547780816)
--(axis cs:6,0.112892060632741)
--(axis cs:7,0.114007086615705)
--(axis cs:8,0.111933064757778)
--(axis cs:9,0.10870154302227)
--(axis cs:10,0.104423899411884)
--(axis cs:11,0.101282879325315)
--(axis cs:11,0.103018195943502)
--(axis cs:11,0.103018195943502)
--(axis cs:10,0.106328788760159)
--(axis cs:9,0.110653295687408)
--(axis cs:8,0.113873386855125)
--(axis cs:7,0.116100440266015)
--(axis cs:6,0.115064928614571)
--(axis cs:5,0.11940559200413)
--(axis cs:4,0.133427690610606)
--(axis cs:3,0.155153575610229)
--(axis cs:2,0.170602670064824)
--(axis cs:1,0.377475493204443)
--cycle;

\path [fill=color2, fill opacity=0.7]
(axis cs:1,0.420283449997503)
--(axis cs:1,0.417216550002497)
--(axis cs:2,0.283382940422159)
--(axis cs:3,0.261022117764393)
--(axis cs:4,0.254616439202031)
--(axis cs:5,0.248396365414131)
--(axis cs:6,0.248199728949557)
--(axis cs:7,0.24811372646711)
--(axis cs:8,0.245031823367642)
--(axis cs:9,0.245012808792063)
--(axis cs:10,0.241962000878907)
--(axis cs:11,0.241962000878907)
--(axis cs:11,0.245537999121093)
--(axis cs:11,0.245537999121093)
--(axis cs:10,0.245537999121093)
--(axis cs:9,0.248737191207937)
--(axis cs:8,0.248718176632358)
--(axis cs:7,0.25188627353289)
--(axis cs:6,0.251800271050443)
--(axis cs:5,0.251603634585869)
--(axis cs:4,0.257883560797969)
--(axis cs:3,0.263977882235607)
--(axis cs:2,0.285367059577841)
--(axis cs:1,0.420283449997503)
--cycle;

\path [fill=color3, fill opacity=0.7]
(axis cs:1,0.352021668944945)
--(axis cs:1,0.349796512873237)
--(axis cs:2,0.332700148895736)
--(axis cs:3,0.323371339313421)
--(axis cs:4,0.318659433287094)
--(axis cs:5,0.300525160122325)
--(axis cs:6,0.300545611790656)
--(axis cs:7,0.287773221388771)
--(axis cs:8,0.289571641279623)
--(axis cs:9,0.287818807695338)
--(axis cs:10,0.286016663391779)
--(axis cs:11,0.284269310501639)
--(axis cs:11,0.286639780407452)
--(axis cs:11,0.286639780407452)
--(axis cs:10,0.288528791153676)
--(axis cs:9,0.290363010486481)
--(axis cs:8,0.292246540538559)
--(axis cs:7,0.290408596793047)
--(axis cs:6,0.303090751845707)
--(axis cs:5,0.303111203514038)
--(axis cs:4,0.321340566712906)
--(axis cs:3,0.325719569777488)
--(axis cs:2,0.334572578376992)
--(axis cs:1,0.352021668944945)
--cycle;

table {%
1 0.351666666666667
2 0.171111111111111
3 0.145
4 0.121666666666667
5 0.113333333333333
6 0.108333333333333
7 0.103333333333333
8 0.100555555555556
9 0.0988888888888888
10 0.0949999999999999
11 0.0938888888888888
};
\addplot [very thick, color1, mark=square, mark size=4, mark options={solid}]
table {%
1 0.375268817204301
2 0.168817204301075
3 0.153763440860215
4 0.132258064516129
5 0.118279569892473
6 0.113978494623656
7 0.11505376344086
8 0.112903225806452
9 0.109677419354839
10 0.105376344086022
11 0.102150537634409
};
\addplot [very thick, color2, mark=diamond, mark size=4, mark options={solid}]
table {%
1 0.41875
2 0.284375
3 0.2625
4 0.25625
5 0.25
6 0.25
7 0.25
8 0.246875
9 0.246875
10 0.24375
11 0.24375
};
\addplot [very thick, color3, mark=oplus, mark size=4, mark options={solid}]
table {%
1 0.350909090909091
2 0.333636363636364
3 0.324545454545455
4 0.32
5 0.301818181818182
6 0.301818181818182
7 0.289090909090909
8 0.290909090909091
9 0.289090909090909
10 0.287272727272727
11 0.285454545454545
};

\end{axis}

\end{tikzpicture}

%% file: onlineEM_vsm_Errorarate_slp.tex
\begin{tikzpicture}

\definecolor{color0}{rgb}{0.967797559291991,0.441274560091574,0.53581031550587}
\definecolor{color1}{rgb}{0.680418912779335,0.615149751467757,0.194054521114453}
\definecolor{color2}{rgb}{0.201253172212011,0.690792081537903,0.479667611892753}
\definecolor{color3}{rgb}{0.219799566082832,0.662515787685034,0.773209315931721}

\begin{axis}[
legend cell align={left},
legend style={fill opacity=0.8, draw opacity=1, text opacity=1, draw=white!80!black},
tick align=outside,
tick pos=left,
x grid style={white!69.0196078431373!black},
xlabel={\Large Time $t$},
xmajorgrids,
xmin=1, xmax=11,
xtick style={color=black},
xtick={1,2,4,6,8,10},
xticklabels={$0$,$300\mulG$,$900\mulG$,$1500\mulG$,$2100\mulG$,$2700\mulG$},
y grid style={white!69.0196078431373!black},
ylabel={},
ymajorgrids,
ymin=-0.000655590686207076, ymax=0.0137674044103486,
ytick style={color=black},
width=7cm,height=6cm,
]
\path [fill=color0, fill opacity=0.7]
(axis cs:1,0)
--(axis cs:1,0)
--(axis cs:2,0)
--(axis cs:3,0)
--(axis cs:4,0)
--(axis cs:5,0)
--(axis cs:6,0)
--(axis cs:7,0)
--(axis cs:8,0)
--(axis cs:9,0)
--(axis cs:10,0)
--(axis cs:11,0)
--(axis cs:11,0)
--(axis cs:11,0)
--(axis cs:10,0)
--(axis cs:9,0)
--(axis cs:8,0)
--(axis cs:7,0)
--(axis cs:6,0)
--(axis cs:5,0)
--(axis cs:4,0)
--(axis cs:3,0)
--(axis cs:2,0)
--(axis cs:1,0)
--cycle;

\path [fill=color1, fill opacity=0.7]
(axis cs:1,0.00689784183852396)
--(axis cs:1,0.00643549149480937)
--(axis cs:2,0)
--(axis cs:3,0)
--(axis cs:4,0)
--(axis cs:5,0)
--(axis cs:6,0)
--(axis cs:7,0)
--(axis cs:8,0)
--(axis cs:9,0)
--(axis cs:10,0)
--(axis cs:11,0)
--(axis cs:11,0)
--(axis cs:11,0)
--(axis cs:10,0)
--(axis cs:9,0)
--(axis cs:8,0)
--(axis cs:7,0)
--(axis cs:6,0)
--(axis cs:5,0)
--(axis cs:4,0)
--(axis cs:3,0)
--(axis cs:2,0)
--(axis cs:1,0.00689784183852396)
--cycle;

\path [fill=color2, fill opacity=0.7]
(axis cs:1,0)
--(axis cs:1,0)
--(axis cs:2,0)
--(axis cs:3,0)
--(axis cs:4,0)
--(axis cs:5,0)
--(axis cs:6,0)
--(axis cs:7,0)
--(axis cs:8,0)
--(axis cs:9,0)
--(axis cs:10,0)
--(axis cs:11,0)
--(axis cs:11,0)
--(axis cs:11,0)
--(axis cs:10,0)
--(axis cs:9,0)
--(axis cs:8,0)
--(axis cs:7,0)
--(axis cs:6,0)
--(axis cs:5,0)
--(axis cs:4,0)
--(axis cs:3,0)
--(axis cs:2,0)
--(axis cs:1,0)
--cycle;

\path [fill=color3, fill opacity=0.7]
(axis cs:1,0.0131118137241415)
--(axis cs:1,0.0123427317304039)
--(axis cs:2,0.00174096412782169)
--(axis cs:3,0.00174096412782169)
--(axis cs:4,0.00174096412782169)
--(axis cs:5,0.00174096412782169)
--(axis cs:6,0.00174096412782169)
--(axis cs:7,0.00174096412782169)
--(axis cs:8,0.00174096412782169)
--(axis cs:9,0.00174096412782169)
--(axis cs:10,0.00174096412782169)
--(axis cs:11,0.00174096412782169)
--(axis cs:11,0.00189539950854193)
--(axis cs:11,0.00189539950854193)
--(axis cs:10,0.00189539950854193)
--(axis cs:9,0.00189539950854193)
--(axis cs:8,0.00189539950854193)
--(axis cs:7,0.00189539950854193)
--(axis cs:6,0.00189539950854193)
--(axis cs:5,0.00189539950854193)
--(axis cs:4,0.00189539950854193)
--(axis cs:3,0.00189539950854193)
--(axis cs:2,0.00189539950854193)
--(axis cs:1,0.0131118137241415)
--cycle;

\addplot [very thick, color0, mark=o, mark size=4, mark options={solid}]
table {%
1 0
2 0
3 0
4 0
5 0
6 0
7 0
8 0
9 0
10 0
11 0
};
\addlegendentry{Alg.~\ref{alg:olem}, $\mulG=2$}
\addplot [very thick, color1, mark=square, mark size=4, mark options={solid}]
table {%
1 0.00666666666666667
2 0
3 0
4 0
5 0
6 0
7 0
8 0
9 0
10 0
11 0
};
\addlegendentry{Alg.~\ref{alg:olem}, $\mulG=3$}
\addplot [very thick, color2, mark=diamond, mark size=4, mark options={solid}]
table {%
1 0
2 0
3 0
4 0
5 0
6 0
7 0
8 0
9 0
10 0
11 0
};
\addlegendentry{Alg.~\ref{alg:olem}, $\mulG=4$}
\addplot [very thick, color3, mark=oplus, mark size=4, mark options={solid}]
table {%
1 0.0127272727272727
2 0.00181818181818181
3 0.00181818181818181
4 0.00181818181818181
5 0.00181818181818181
6 0.00181818181818181
7 0.00181818181818181
8 0.00181818181818181
9 0.00181818181818181
10 0.00181818181818181
11 0.00181818181818181
};
\addlegendentry{Alg.~\ref{alg:olem}, $\mulG=5$}

\end{axis}

\end{tikzpicture}

%% file: onlineEM_vsm_NMI.tex
\begin{tikzpicture}

\definecolor{color0}{rgb}{0.967797559291991,0.441274560091574,0.53581031550587}
\definecolor{color1}{rgb}{0.680418912779335,0.615149751467757,0.194054521114453}
\definecolor{color2}{rgb}{0.201253172212011,0.690792081537903,0.479667611892753}
\definecolor{color3}{rgb}{0.219799566082832,0.662515787685034,0.773209315931721}

\begin{axis}[
legend cell align={left},
legend style={fill opacity=0.8, draw opacity=1, text opacity=1, draw=white!80!black},
tick align=outside,
tick pos=left,
x grid style={white!69.0196078431373!black},
xlabel={\Large Time $t$},
xmajorgrids,
xmin=1, xmax=11,
xtick style={color=black},
xtick={1,2,4,6,8,10},
xticklabels={$0$,$300\mulG$,$900\mulG$,$1500\mulG$,$2100\mulG$,$2700\mulG$},
y grid style={white!69.0196078431373!black},
ylabel={\Large NMI},
ymajorgrids,
ymin=0, ymax=1.0,
ytick style={color=black},
width=7cm,height=6cm,
]
\path [fill=color0, fill opacity=0.7]
(axis cs:1,0.140168274429962)
--(axis cs:1,0.131391144992933)
--(axis cs:2,0.815624301381582)
--(axis cs:3,0.92565645956745)
--(axis cs:4,0.968619258884642)
--(axis cs:5,0.993969175488427)
--(axis cs:6,0.999674397879854)
--(axis cs:7,0.999698601226363)
--(axis cs:8,0.999747084053425)
--(axis cs:9,0.999747084053425)
--(axis cs:10,0.999722827814598)
--(axis cs:11,0.999722827814598)
--(axis cs:11,0.999755447594543)
--(axis cs:11,0.999755447594543)
--(axis cs:10,0.999755447594543)
--(axis cs:9,0.999778613180528)
--(axis cs:8,0.999778613180528)
--(axis cs:7,0.999732246518164)
--(axis cs:6,0.999709009642077)
--(axis cs:5,0.995485363030182)
--(axis cs:4,0.973566750779311)
--(axis cs:3,0.935084666685799)
--(axis cs:2,0.829373746910042)
--(axis cs:1,0.140168274429962)
--cycle;

\path [fill=color1, fill opacity=0.7]
(axis cs:1,0.253316504297325)
--(axis cs:1,0.249355704568276)
--(axis cs:2,0.776869780367418)
--(axis cs:3,0.896178829185393)
--(axis cs:4,0.943192522733612)
--(axis cs:5,0.953822528185983)
--(axis cs:6,0.95482824287719)
--(axis cs:7,0.955876073491619)
--(axis cs:8,0.957146953551418)
--(axis cs:9,0.958243883239415)
--(axis cs:10,0.959883817591965)
--(axis cs:11,0.962007429746893)
--(axis cs:11,0.96592261519986)
--(axis cs:11,0.96592261519986)
--(axis cs:10,0.963970434747876)
--(axis cs:9,0.962470110554093)
--(axis cs:8,0.961462193636599)
--(axis cs:7,0.960301413858399)
--(axis cs:6,0.959330291551316)
--(axis cs:5,0.958340117584826)
--(axis cs:4,0.947855999320047)
--(axis cs:3,0.901877663836656)
--(axis cs:2,0.783801043163998)
--(axis cs:1,0.253316504297325)
--cycle;

\path [fill=color2, fill opacity=0.7]
(axis cs:1,0.32577524234713)
--(axis cs:1,0.323721885466683)
--(axis cs:2,0.655002921799173)
--(axis cs:3,0.730600124734728)
--(axis cs:4,0.771402623296405)
--(axis cs:5,0.800444351766844)
--(axis cs:6,0.82065840554895)
--(axis cs:7,0.836080595503403)
--(axis cs:8,0.848939641011117)
--(axis cs:9,0.858244954180877)
--(axis cs:10,0.866776642486466)
--(axis cs:11,0.875601906660252)
--(axis cs:11,0.879406460654777)
--(axis cs:11,0.879406460654777)
--(axis cs:10,0.870692255479546)
--(axis cs:9,0.862290802886109)
--(axis cs:8,0.853070941551068)
--(axis cs:7,0.840356189794411)
--(axis cs:6,0.825041618592274)
--(axis cs:5,0.804987630126294)
--(axis cs:4,0.776127364964868)
--(axis cs:3,0.735657227158808)
--(axis cs:2,0.660115156131329)
--(axis cs:1,0.32577524234713)
--cycle;

\path [fill=color3, fill opacity=0.7]
(axis cs:1,0.364113818324618)
--(axis cs:1,0.362744061344878)
--(axis cs:2,0.669822777856628)
--(axis cs:3,0.780741121582969)
--(axis cs:4,0.814874869918577)
--(axis cs:5,0.828517054017732)
--(axis cs:6,0.838167977953256)
--(axis cs:7,0.84585085431513)
--(axis cs:8,0.855294615984173)
--(axis cs:9,0.859890683196992)
--(axis cs:10,0.864134186717661)
--(axis cs:11,0.866343381884736)
--(axis cs:11,0.870186782104073)
--(axis cs:11,0.870186782104073)
--(axis cs:10,0.868014543017363)
--(axis cs:9,0.863854462887596)
--(axis cs:8,0.859326153309375)
--(axis cs:7,0.850073768318858)
--(axis cs:6,0.842486695856631)
--(axis cs:5,0.832848388549906)
--(axis cs:4,0.819238438514714)
--(axis cs:3,0.785014553509601)
--(axis cs:2,0.67345071652205)
--(axis cs:1,0.364113818324618)
--cycle;

\addplot [very thick, color0, mark=o, mark size=4, mark options={solid}]
table {%
1 0.135779709711448
2 0.822499024145812
3 0.930370563126624
4 0.971093004831977
5 0.994727269259305
6 0.999691703760965
7 0.999715423872263
8 0.999762848616976
9 0.999762848616976
10 0.99973913770457
11 0.99973913770457
};

\addplot [very thick, color1, mark=square, mark size=4, mark options={solid}]
table {%
1 0.2513361044328
2 0.780335411765708
3 0.899028246511025
4 0.945524261026829
5 0.956081322885404
6 0.957079267214253
7 0.958088743675009
8 0.959304573594009
9 0.960356996896754
10 0.96192712616992
11 0.963965022473376
};

\addplot [very thick, color2, mark=diamond, mark size=4, mark options={solid}]
table {%
1 0.324748563906906
2 0.657559038965251
3 0.733128675946768
4 0.773764994130637
5 0.802715990946569
6 0.822850012070612
7 0.838218392648907
8 0.851005291281092
9 0.860267878533493
10 0.868734448983006
11 0.877504183657514
};

\addplot [very thick, color3, mark=oplus, mark size=4, mark options={solid}]
table {%
1 0.363428939834748
2 0.671636747189339
3 0.782877837546285
4 0.817056654216645
5 0.830682721283819
6 0.840327336904943
7 0.847962311316994
8 0.857310384646774
9 0.861872573042294
10 0.866074364867512
11 0.868265081994404
};

\end{axis}

\end{tikzpicture}

%% file: onlineEM_vsm_NMI_slp.tex
\begin{tikzpicture}

\definecolor{color0}{rgb}{0.967797559291991,0.441274560091574,0.53581031550587}
\definecolor{color1}{rgb}{0.680418912779335,0.615149751467757,0.194054521114453}
\definecolor{color2}{rgb}{0.201253172212011,0.690792081537903,0.479667611892753}
\definecolor{color3}{rgb}{0.219799566082832,0.662515787685034,0.773209315931721}

\begin{axis}[
legend cell align={left},
legend style={fill opacity=0.8, draw opacity=1, text opacity=1, draw=white!80!black},
tick align=outside,
tick pos=left,
x grid style={white!69.0196078431373!black},
xlabel={\Large Time $t$},
xmajorgrids,
xmin=1, xmax=11,
xtick style={color=black},
xtick={1,2,4,6,8,10},
xticklabels={$0$,$300\mulG$,$900\mulG$,$1500\mulG$,$2100\mulG$,$2700\mulG$},
y grid style={white!69.0196078431373!black},
ylabel={},
ymajorgrids,
ymin=0.974684728932767, ymax=1.00120548909844,
ytick style={color=black},
width=7cm,height=6cm,
]
\path [fill=color0, fill opacity=0.7]
(axis cs:1,0.993364024860643)
--(axis cs:1,0.992461021845007)
--(axis cs:2,1)
--(axis cs:3,1)
--(axis cs:4,1)
--(axis cs:5,1)
--(axis cs:6,1)
--(axis cs:7,1)
--(axis cs:8,1)
--(axis cs:9,1)
--(axis cs:10,1)
--(axis cs:11,1)
--(axis cs:11,1)
--(axis cs:11,1)
--(axis cs:10,1)
--(axis cs:9,1)
--(axis cs:8,1)
--(axis cs:7,1)
--(axis cs:6,1)
--(axis cs:5,1)
--(axis cs:4,1)
--(axis cs:3,1)
--(axis cs:2,1)
--(axis cs:1,0.993364024860643)
--cycle;

\path [fill=color1, fill opacity=0.7]
(axis cs:1,0.992190896777524)
--(axis cs:1,0.991605004353798)
--(axis cs:2,1)
--(axis cs:3,1)
--(axis cs:4,1)
--(axis cs:5,1)
--(axis cs:6,1)
--(axis cs:7,1)
--(axis cs:8,1)
--(axis cs:9,1)
--(axis cs:10,1)
--(axis cs:11,1)
--(axis cs:11,1)
--(axis cs:11,1)
--(axis cs:10,1)
--(axis cs:9,1)
--(axis cs:8,1)
--(axis cs:7,1)
--(axis cs:6,1)
--(axis cs:5,1)
--(axis cs:4,1)
--(axis cs:3,1)
--(axis cs:2,1)
--(axis cs:1,0.992190896777524)
--cycle;

\path [fill=color2, fill opacity=0.7]
(axis cs:1,1)
--(axis cs:1,1)
--(axis cs:2,1)
--(axis cs:3,1)
--(axis cs:4,1)
--(axis cs:5,1)
--(axis cs:6,1)
--(axis cs:7,1)
--(axis cs:8,1)
--(axis cs:9,1)
--(axis cs:10,1)
--(axis cs:11,1)
--(axis cs:11,1)
--(axis cs:11,1)
--(axis cs:10,1)
--(axis cs:9,1)
--(axis cs:8,1)
--(axis cs:7,1)
--(axis cs:6,1)
--(axis cs:5,1)
--(axis cs:4,1)
--(axis cs:3,1)
--(axis cs:2,1)
--(axis cs:1,1)
--cycle;

\path [fill=color3, fill opacity=0.7]
(axis cs:1,0.977053124891917)
--(axis cs:1,0.975890218031207)
--(axis cs:2,0.997640990783181)
--(axis cs:3,0.999817334895808)
--(axis cs:4,1)
--(axis cs:5,1)
--(axis cs:6,1)
--(axis cs:7,1)
--(axis cs:8,1)
--(axis cs:9,1)
--(axis cs:10,1)
--(axis cs:11,1)
--(axis cs:11,1)
--(axis cs:11,1)
--(axis cs:10,1)
--(axis cs:9,1)
--(axis cs:8,1)
--(axis cs:7,1)
--(axis cs:6,1)
--(axis cs:5,1)
--(axis cs:4,1)
--(axis cs:3,0.999832218277798)
--(axis cs:2,0.997833200649692)
--(axis cs:1,0.977053124891917)
--cycle;

\addplot [very thick, color0, mark=o, mark size=4, mark options={solid}]
table {%
1 0.992912523352825
2 1
3 1
4 1
5 1
6 1
7 1
8 1
9 1
10 1
11 1
};

\addplot [very thick, color1, mark=square, mark size=4, mark options={solid}]
table {%
1 0.991897950565661
2 1
3 1
4 1
5 1
6 1
7 1
8 1
9 1
10 1
11 1
};

\addplot [very thick, color2, mark=diamond, mark size=4, mark options={solid}]
table {%
1 1
2 1
3 1
4 1
5 1
6 1
7 1
8 1
9 1
10 1
11 1
};

\addplot [very thick, color3, mark=oplus, mark size=4, mark options={solid}]
table {%
1 0.976471671461562
2 0.997737095716437
3 0.999824776586803
4 1
5 1
6 1
7 1
8 1
9 1
10 1
11 1
};

\end{axis}

\end{tikzpicture}

%% file: onlineEM_vsm_Fvalue.tex
\begin{tikzpicture}

\definecolor{color0}{rgb}{0.967797559291991,0.441274560091574,0.53581031550587}
\definecolor{color1}{rgb}{0.680418912779335,0.615149751467757,0.194054521114453}
\definecolor{color2}{rgb}{0.201253172212011,0.690792081537903,0.479667611892753}
\definecolor{color3}{rgb}{0.219799566082832,0.662515787685034,0.773209315931721}

\begin{axis}[
legend cell align={left},
legend style={fill opacity=0.8, draw opacity=1, text opacity=1, at={(0.97,0.03)}, anchor=south east, draw=white!80!black},
tick align=outside,
tick pos=left,
x grid style={white!69.0196078431373!black},
xlabel={\Large Time $t$},
xmajorgrids,
xmin=1, xmax=11,
xtick style={color=black},
xtick={1,2,4,6,8,10},
xticklabels={$0$,$300\mulG$,$900\mulG$,$1500\mulG$,$2100\mulG$,$2700\mulG$},
y grid style={white!69.0196078431373!black},
ylabel={\Large ${\cal L}( \Theta )$ },
ymajorgrids,
ymin=-26.5758164092242, ymax=-12.7312444249395,
ytick style={color=black},
width=7cm,height=8cm,
]
\addplot [very thick, color0, mark=o, mark size=4, mark options={solid}]
table {%
1 -19.4776853427977
2 -15.5456350339941
3 -15.0087987542489
4 -14.4770480682752
5 -14.0080345925933
6 -13.7046516972137
7 -13.5010407466317
8 -13.37938843499
9 -13.3849109250883
10 -13.3687934099704
11 -13.3605431514979
};
\addplot [very thick, color1, mark=square, mark size=4, mark options={solid}]
table {%
1 -20.1328908127433
2 -14.4193303626999
3 -14.2686674756995
4 -14.2957089703631
5 -14.3109087652123
6 -14.3221093791633
7 -14.3235450104399
8 -14.3211814698677
9 -14.3355683442965
10 -14.3470556610771
11 -14.356088312547
};
\addplot [very thick, color2, mark=diamond, mark size=4, mark options={solid}]
table {%
1 -25.9465176826658
2 -16.7208291460792
3 -15.9734579336743
4 -15.5438044915815
5 -15.2020903332985
6 -15.0833840756794
7 -15.0900552192566
8 -15.0862355032081
9 -15.0841788199203
10 -15.0764962507597
11 -15.0716918541999
};
\addplot [very thick, color3, mark=oplus, mark size=4, mark options={solid}]
table {%
1 -25.0750926528999
2 -18.6556090887248
3 -18.2087941080126
4 -18.1406454198424
5 -18.1266048179212
6 -18.1467783306045
7 -18.1758010687963
8 -18.1872126992063
9 -18.1917767858293
10 -18.1951080314476
11 -18.1965363230617
};
\end{axis}

\end{tikzpicture}

%% file: onlineEM_vsm_Fvalue_slp.tex
\begin{tikzpicture}

\definecolor{color0}{rgb}{0.967797559291991,0.441274560091574,0.53581031550587}
\definecolor{color1}{rgb}{0.680418912779335,0.615149751467757,0.194054521114453}
\definecolor{color2}{rgb}{0.201253172212011,0.690792081537903,0.479667611892753}
\definecolor{color3}{rgb}{0.219799566082832,0.662515787685034,0.773209315931721}

\begin{axis}[
legend cell align={left},
legend style={fill opacity=0.8, draw opacity=1, text opacity=1, at={(0.97,0.03)}, anchor=south east, draw=white!80!black},
tick align=outside,
tick pos=left,
x grid style={white!69.0196078431373!black},
xlabel={\Large Time $t$},
xmajorgrids,
xmin=1, xmax=11,
xtick style={color=black},
xtick={1,2,4,6,8,10},
xticklabels={$0$,$300\mulG$,$900\mulG$,$1500\mulG$,$2100\mulG$,$2700\mulG$},
y grid style={white!69.0196078431373!black},
ylabel={ },
ymajorgrids,
ymin=-42.1257560627507, ymax=-15.1566236504008,
ytick style={color=black},
width=7cm,height=8cm,
]
\addplot [very thick, color0, mark=o, mark size=4, mark options={solid}]
table {%
1 -19.5619417967264
2 -17.0429502049785
3 -16.6815037569471
4 -16.4974864325423
5 -16.4311501240835
6 -16.4112410610282
7 -16.4002875695269
8 -16.3930953559536
9 -16.3824933055076
10 -16.3843789291111
11 -16.3860454208045
};
\addplot [very thick, color1, mark=square, mark size=4, mark options={solid}]
table {%
1 -23.0392461457309
2 -19.6018613977018
3 -19.0621055613387
4 -18.9098656512242
5 -18.849044748459
6 -18.8043057299781
7 -18.7923337024877
8 -18.7919986696435
9 -18.793437075571
10 -18.7963945579723
11 -18.7906773313085
};
\addplot [very thick, color2, mark=diamond, mark size=4, mark options={solid}]
table {%
1 -30.8649519670745
2 -25.2494877073934
3 -24.6700601950237
4 -24.4509192006556
5 -24.3492411340082
6 -24.3105321371386
7 -24.312639740914
8 -24.3206905979661
9 -24.3356076640995
10 -24.3396510504779
11 -24.3460852042126
};
\addplot [very thick, color3, mark=oplus, mark size=4, mark options={solid}]
table {%
1 -40.8998864076439
2 -32.1263752895992
3 -30.4610304143309
4 -29.7077944535335
5 -29.314302278564
6 -29.0585161376538
7 -28.8579898109934
8 -28.6891919648087
9 -28.5467121293402
10 -28.4214645821721
11 -28.3062449687926
};
\end{axis}

\end{tikzpicture}

%% file: onlineEMdetector_baw.tex
\begin{tikzpicture}

\definecolor{color0}{rgb}{0.12156862745098,0.466666666666667,0.705882352941177}
\definecolor{color1}{rgb}{0.967797559291991,0.441274560091574,0.53581031550587}
\definecolor{color2}{rgb}{0.680418912779335,0.615149751467757,0.194054521114453}
\definecolor{color3}{rgb}{0,0.75,0.75}

\begin{groupplot}[group style={group size=1 by 3,vertical sep=0cm},   scale only axis,
    width=1.1\textwidth,
    height=3cm,ticklabel style = {font=\Large}]
\nextgroupplot[
scaled x ticks=manual:{}{\pgfmathparse{#1}},
tick align=outside,
tick pos=left,
x grid style={white!69.0196078431373!black},
xmin=0, xmax=1100,
xtick={1,200,400,600,800,1000},
xtick style={color=black},
xticklabels={},
y grid style={white!69.0196078431373!black},
ymin=0, ymax=4.1,
ytick style={color=black},
]
\addplot [ultra thick, black]
table {%
1 0
12 0
23 0.00606799066251562
34 0.00866165159331628
45 0.039533546299089
56 0.0787551450261069
67 0.120985502955121
78 0.162811937780505
89 0.204041766090736
100 0.244024122027506
111 0.282565143661351
122 0.323498980427034
133 0.361108051198398
144 0.402415585786048
155 0.440856414964699
166 0.480426723650275
177 0.522866952358312
188 0.568263333349513
199 0.611238473946935
210 0.650879846529854
221 0.692434084162919
232 0.734113836360022
243 0.775016968362264
254 0.816318399499269
265 0.859573672648768
276 0.903525058307971
287 0.945434953978984
298 0.988655483012721
309 1.02743116865094
320 1.07300078388929
331 1.1209482451421
342 1.14707384631908
353 1.1746894125419
364 1.2192972546182
375 1.26123338161781
386 1.30506993436925
397 1.34335465355107
408 1.38406228677058
419 1.42693478998389
430 1.47523641193593
441 1.5252947413539
452 1.57585026781197
463 1.6242433749657
474 1.67585264305398
485 1.72591501761102
496 1.77800076498954
507 1.82784346419929
518 1.87790705818375
529 1.92828147172761
540 1.97994924004932
551 2.02965147084193
562 2.07659809797057
573 2.12115794710826
584 2.16344172500819
595 2.19663070539431
606 2.24147996278427
617 2.28747986738949
628 2.33831371472577
639 2.39221539480112
650 2.44529755083201
661 2.49664199397625
672 2.54756171668257
683 2.59689680375692
694 2.64831628267887
705 2.69579680761292
716 2.74642841427965
727 2.79964677974802
738 2.84954081714983
749 2.90175922491445
760 2.949171203837
771 2.9862787113981
782 3.01589598457213
793 3.04919291609034
804 3.08301835356868
815 3.11993207037017
826 3.15833864744244
837 3.20139689529849
848 3.24453326936322
859 3.28090624056616
870 3.31046699363882
881 3.3409214967938
892 3.3723639301049
903 3.40461406308107
914 3.43789559465917
925 3.47195804654378
936 3.5166955910143
947 3.56372127260864
958 3.61044242486996
969 3.6570576575652
980 3.69972811721868
991 3.73585590951749
1002 3.76803216331901
1013 3.80096068557674
1024 3.83724837065095
1035 3.87392850153521
1046 3.92327911560172
1057 3.97139363021611
1068 4.01194794255609
1079 4.04689527737079
1090 4.08058389162819
};
\addplot [ultra thick, red]
table {%
800 -0.0277123396832821
800 5.45731780517138
};

\nextgroupplot[
scaled x ticks=manual:{}{\pgfmathparse{#1}},
tick align=outside,
tick pos=left,
x grid style={white!69.0196078431373!black},
xmin=0, xmax=1100,
xtick={1,200,400,600,800,1000},
xtick style={color=black},
xticklabels={},
y grid style={white!69.0196078431373!black},
ymin=0, ymax=4.5,
ytick={0,1,2,3},
ytick style={color=black},
]
\addplot [ultra thick, orange]
table {%
1 3.00435154811848
12 2.06366819111457
23 1.64903279555595
34 3.52106449053139
45 3.06763432542951
56 2.83915776210176
67 1.83093656890663
78 3.45261300656003
89 3.05225937247925
100 3.50893113370385
111 2.92803583697077
122 1.69992150046988
133 2.30805260439062
144 2.25833596706403
155 3.14710001580832
166 2.15541557580888
177 1.93570465215397
188 2.12331849955722
199 1.96272225283686
210 2.2944964131422
221 2.33627593976313
232 2.9375073974923
243 2.44697132891371
254 1.89566012414257
265 2.36445789160822
276 2.26386413998039
287 4.13981831191904
298 2.06370774728911
309 3.51642611514519
320 1.40735558378864
331 2.38350737954379
342 2.44035106178963
353 1.13563966480515
364 1.6828785390925
375 2.0782282877411
386 1.28028178157649
397 3.25059374782782
408 3.04882957123018
419 2.66744722595245
430 1.29170081289961
441 2.67958466650623
452 2.41686848792955
463 1.78730161270307
474 1.97213096138899
485 2.00049179864278
496 3.4149799694034
507 0.788276972426739
518 1.6406639213706
529 2.85260679662289
540 1.635890712548
551 3.07700764010928
562 1.76577456444834
573 1.92041914847606
584 2.2713600563657
595 2.63815129050179
606 2.28719801695967
617 2.72396775615932
628 1.78822657304891
639 1.23111067789267
650 2.47194004838762
661 2.04756717379779
672 1.48893757742003
683 2.37756986932166
694 3.15195901002709
705 2.7734016461175
716 1.80772677193649
727 1.12120796459385
738 2.23989330647522
749 3.05512577852625
760 3.67410959116166
771 2.32262983379404
782 1.56370519768395
793 2.10289304191822
804 3.69114615830343
815 3.12729691269405
826 3.11573311192341
837 0.898195988399382
848 2.951571312392
859 2.54559129058433
870 2.07049952145913
881 3.38529297125496
892 2.36950533624154
903 3.14375663492089
914 1.39329770925485
925 3.33657404227768
936 3.01852070249282
947 3.86046095758814
958 3.07945214048181
969 3.17948844578227
980 2.50739828845026
991 2.69443304657202
1002 2.45030112360695
1013 1.36541762997486
1024 3.31010331043187
1035 3.08847192630452
1046 3.05649165124578
1057 1.21119688974292
1068 2.49771024466542
1079 1.49534519071718
1090 2.94761603092019
};
\addplot [ultra thick, red]
table {%
800 -0.0277123396832821
800 5.45731780517138
};

\nextgroupplot[
tick align=outside,
tick pos=left,
x grid style={white!69.0196078431373!black},
xmin=0, xmax=1100,
xtick={1,200,400,600,800,1000},
xticklabels={\large $1$,$200$,$400$,$600$,$800$,$1000$},
xtick style={color=black},
xlabel={\Huge Time $t$},
y grid style={white!69.0196078431373!black},
ymin=0, ymax=4.5,
ytick={0,1,2,3},
ytick style={color=black},
]
\addplot [ultra thick, color3]
table {%
1 1.15995568701927
12 1.05495728237149
23 1.15173808183206
34 1.07236750811546
45 1.44688919369806
56 0.945089299191915
67 1.13447868570783
78 1.64069467122831
89 1.33547240613039
100 1.05305869293311
111 1.49975923491855
122 1.40953297101069
133 1.04822353454925
144 1.00802842243081
155 0.912579065735803
166 1.50151318880004
177 1.05410878517328
188 1.38511107987324
199 1.26889110447993
210 1.02502750282966
221 1.04976341806378
232 1.09457099489525
243 1.16238944209385
254 1.32198170397192
265 1.47150050728883
276 1.29626206536309
287 0.888374983333159
298 1.30244217248653
309 1.05170698827704
320 0.944962372852373
331 1.21306933952497
342 1.04716811993247
353 1.20833930599549
364 2.03841195871954
375 1.34673295184926
386 1.40648401941229
397 1.39339240420347
408 1.21764709926624
419 1.11569648605095
430 1.31937056540242
441 1.08102362447239
452 1.3522287878854
463 1.1459265466743
474 1.64262608647931
485 1.53245441759953
496 0.940123304595661
507 2.06614593608869
518 0.985091891377733
529 1.3447854409494
540 0.933369398050871
551 1.00692471547415
562 1.24215376926482
573 1.4895055910634
584 2.15571946298882
595 1.06369140735718
606 1.01192158638554
617 1.01052975590547
628 1.0382341087506
639 1.12199603762967
650 1.05563465316063
661 1.09812233280244
672 1.27426737537536
683 0.933650960983148
694 1.19101275102461
705 0.853453070628588
716 1.22446220578446
727 1.20469050422183
738 1.76562657708103
749 0.872689121226735
760 1.12056979852043
771 1.35715654872274
782 1.41662386787023
793 1.19340301775271
804 3.51716445400767
815 3.34704735617662
826 2.09549805137119
837 2.13392073973891
848 2.16423999598171
859 2.09112209179373
870 3.06605500342261
881 3.77493075278562
892 2.10693273832142
903 1.95518345851395
914 3.11409094484523
925 4.44663115171189
936 2.05964702388588
947 2.51202740670895
958 1.90805618385439
969 3.16310846833915
980 2.63958546216915
991 2.61889129978696
1002 2.74319811149806
1013 2.95644806526381
1024 2.34158800761083
1035 2.29044666299093
1046 2.91015749558404
1057 1.81242128027763
1068 2.94765639959133
1079 2.27254446655159
1090 3.12008224549395
};
\addplot [ultra thick, red]
table {%
800 -0.0277123396832821
800 5.45731780517138
};
\end{groupplot}

\draw ({$(current bounding box.south east)!1.01!(current bounding box.south west)$}|-{$(current bounding box.south west)!0.25!(current bounding box.north west)$}) node[
  scale=1,
  anchor= west,
  text=black,
  rotate=90.0
]{\Huge Detector value};
\end{tikzpicture}

%% file: onlineEMdetector_babl.tex
\begin{tikzpicture}

\definecolor{color0}{rgb}{0.12156862745098,0.466666666666667,0.705882352941177}
\definecolor{color1}{rgb}{0.967797559291991,0.441274560091574,0.53581031550587}
\definecolor{color2}{rgb}{0.680418912779335,0.615149751467757,0.194054521114453}
\definecolor{color3}{rgb}{0,0.75,0.75}

\begin{groupplot}[group style={group size=1 by 3,vertical sep=0cm},   scale only axis,
    width=1.1\textwidth,
    height=3cm,legend pos=north west,ticklabel style = {font=\Large}]
\nextgroupplot[
scaled x ticks=manual:{}{\pgfmathparse{#1}},
tick align=outside,
tick pos=left,
x grid style={white!69.0196078431373!black},
xmin=0, xmax=1100,
xtick={1,200,400,600,800,1000},
xtick style={color=black},
xticklabels={},
y grid style={white!69.0196078431373!black},
ymin=0, ymax=1,
ytick style={color=black},
]
\addplot [ultra thick, black]
table {%
1 0
12 0
23 0
34 0
45 0
56 0
67 0
78 0
89 0
100 0
111 0
122 0
133 0.00364678275625924
144 0.00122414120856273
155 0
166 0
177 0
188 0
199 0
210 0
221 0
232 0
243 0
254 0
265 0
276 0
287 0.0159588696131944
298 0.0634842493033807
309 0.0115395228772895
320 0
331 0
342 0
353 0
364 0
375 0
386 0.0291367483576074
397 0
408 0
419 0
430 0.131412564647484
441 0.00144880536838822
452 0
463 0
474 0
485 0
496 0.00135766497628928
507 0
518 0
529 0.00430408742063674
540 0
551 0
562 0
573 0
584 0
595 0
606 0
617 0
628 0
639 0
650 0
661 0.00183344812920767
672 0
683 0
694 0.00401192888967801
705 0.0173701016501221
716 0.0333886191206132
727 0
738 0.00551470299933284
749 0.0446795744526575
760 0
771 0
782 0
793 0
804 0
815 0.00671597057806431
826 0.0899986379506032
837 0.441732286331399
848 0.499404265061881
859 0.405530854140313
870 0.290729395365629
881 0.133007584854882
892 0.082901524246277
903 0.125535221266705
914 0.0763261202612861
925 0.0739017137743995
936 0.0731916417197699
947 0.194994580740219
958 0.339562330197975
969 0.504082346304244
980 0.748575133011056
991 0.613433720889441
1002 0.47343510986918
1013 0.43434913567692
1024 0.296821783337176
1035 0.299581642030069
1046 0.40048666011436
1057 0.404761839478882
1068 0.268858081490942
1079 0.119644230465538
1090 0.243557979226745
};
\addlegendentry{\LARGE CUSUM \cite{kaushik2021network}}
\addplot [ultra thick, red]
table {%
800 -0.0277123396832821
800 3.45731780517138
};

\nextgroupplot[
scaled x ticks=manual:{}{\pgfmathparse{#1}},
tick align=outside,
tick pos=left,
x grid style={white!69.0196078431373!black},
xmin=0, xmax=1100,
xtick={1,200,400,600,800,1000},
xtick style={color=black},
xticklabels={},
y grid style={white!69.0196078431373!black},
ymin=0, ymax=1,
ytick={0,0.2,0.4,0.6,0.8},
ytick style={color=black},
]
\addplot [ultra thick, orange]
table {%
1 0.103983115310728
13 0.110890858953606
25 0.0759178324387165
37 0.161923804926836
49 0.114098911497536
61 0.0724420429375234
73 0.0535841855030233
85 0.112746741215376
97 0.142996051002554
109 0.158256556691396
121 0.157652620293906
133 0.147062520019533
145 0.138646513081429
157 0.166401240534708
169 0.0913300351377394
181 0.115111487657495
193 0.125079918996774
205 0.0645648799220482
217 0.099815681672652
229 0.153990631753638
241 0.112207816693382
253 0.0808936072744275
265 0.0820791270989065
277 0.202483296676337
289 0.206315345719548
301 0.100838609523547
313 0.0762625893252061
325 0.161018299057875
337 0.111781247604853
349 0.0559237459864392
361 0.0629566874001104
373 0.0859399194372266
385 0.221347901133275
397 0.0798821225214259
409 0.149529544819112
421 0.079467710041622
433 0.102483057550336
445 0.123988331921186
457 0.0396876140269837
469 0.0855520051864164
481 0.131495775522235
493 0.0566669450931201
505 0.107721820799855
517 0.116500651899364
529 0.130683618427273
541 0.0842071367074371
553 0.0763821849915692
565 0.173916239121194
577 0.169583989783684
589 0.169588136184621
601 0.0805289797647098
613 0.145294595481562
625 0.141862424284207
637 0.122255045197742
649 0.164166362156963
661 0.188106042322234
673 0.141367078514261
685 0.121404870943884
697 0.135104557838242
709 0.138728253993324
721 0.0791245412184839
733 0.0947673309423587
745 0.104761684894514
757 0.0995462303882266
769 0.041514045379811
781 0.135563038616399
793 0.148417857385458
805 0.467876093139331
817 0.579353535666767
829 0.32161052569875
841 0.355042870785467
853 0.516604303680619
865 0.333172768174719
877 0.223770979368471
889 0.366615320648554
901 1.1923649928525
913 0.296388431319766
925 0.256538669436362
937 0.327745340185517
949 0.190593471103114
961 0.66981914428115
973 0.595981051262499
985 0.19632923654015
997 0.610970611300582
1009 0.461427549940887
1021 0.196428488341081
1033 0.639393124254397
1045 0.709598627185429
1057 0.61566985853718
1069 0.398237504666093
1081 0.338390156007349
1093 0.680974899469802
1105 0.764776421236457
1117 0.566934554158704
1129 0.584316119681235
1141 0.462998665420487
1153 0.435209750969021
1165 0.841981515955146
1177 0.393992353969129
1189 0.581924918702988
};
\addlegendentry{\LARGE BSMSD \cite{isufi2018blind}}
\addplot [ultra thick, red]
table {%
800 -0.0277123396832821
800 3.45731780517138
};

\nextgroupplot[
tick align=outside,
tick pos=left,
x grid style={white!69.0196078431373!black},
xmin=0, xmax=1100,
xtick={1,200,400,600,800,1000},
xticklabels={\large $1$,$200$,$400$,$600$,$800$,$1000$},
xtick style={color=black},
xlabel={\Huge Time $t$},
y grid style={white!69.0196078431373!black},
ymin=0, ymax=1,
ytick={0,0.2,0.4,0.6,0.8},
ytick style={color=black},
]
\addplot [ultra thick, color3]
table {%
1 0.262793902628373
13 0.35962784051243
25 0.283090797724608
37 0.263442422050961
49 0.261375314095533
61 0.290578486847949
73 0.171751350780432
85 0.295997148120957
97 0.218238794441054
109 0.378060416859588
121 0.331334037070822
133 0.347202069583399
145 0.355073937463701
157 0.378970538109098
169 0.197596946564412
181 0.223299120812332
193 0.252525630675041
205 0.313179852118361
217 0.229864674003688
229 0.312011383855594
241 0.284412041053995
253 0.41058960435731
265 0.259258130503093
277 0.442497763650053
289 0.312501744277082
301 0.326535821708909
313 0.447592671355713
325 0.248363107119848
337 0.329761569745584
349 0.286931623458419
361 0.334520928777712
373 0.36559832524858
385 0.275684599283173
397 0.240425955035203
409 0.220064393160246
421 0.388823932594822
433 0.311730598177264
445 0.302465245886446
457 0.182566789997453
469 0.320185624066071
481 0.506452484947281
493 0.258605035058289
505 0.254863455887918
517 0.223851555996863
529 0.29326385980999
541 0.284469676594516
553 0.285691260683309
565 0.236756564352787
577 0.262274319301273
589 0.390370991516775
601 0.27222936255407
613 0.267532723384729
625 0.170820730091733
637 0.324030393298261
649 0.358975552155541
661 0.244926369202937
673 0.305605290013438
685 0.268582225585108
697 0.255723433188661
709 0.397615979322398
721 0.269053388519139
733 0.243687182190179
745 0.318018533316307
757 0.194805917544987
769 0.258812485080984
781 0.230025001852188
793 0.336914290308705
805 0.500274800607859
817 0.561220898513211
829 0.474192293911983
841 0.405356090788963
853 1.01263026488113
865 0.596397432235005
877 0.463250757701268
889 0.659786982080748
901 0.752394464176983
913 0.473138321220506
925 0.459358155647231
937 0.682564110229659
949 0.444240844509944
961 0.7247384728987
973 0.70005380741866
985 0.416226211380087
997 0.758196276972016
1009 0.669130880146506
1021 0.422898721616348
1033 0.629831838437387
1045 0.809109740093699
1057 0.496524743798717
1069 0.409715424472936
1081 0.365920476610693
1093 0.698657481325651
1105 0.715192636469153
1117 0.760394354999208
1129 0.618002891717324
1141 0.710612065070104
1153 0.560331276677429
1165 0.684587596866028
1177 0.598242861632498
1189 0.489136776426498
};
\addlegendentry{\LARGE Alg.~\ref{alg:olem} w/ \eqref{eq:detector_p}}
\addplot [ultra thick, red]
table {%
800 -0.0277123396832821
800 3.45731780517138
};
\end{groupplot}
\end{tikzpicture}

%% file: onlineEMdetector_cpw.tex
\begin{tikzpicture}

\definecolor{color0}{rgb}{0.12156862745098,0.466666666666667,0.705882352941177}
\definecolor{color1}{rgb}{0.967797559291991,0.441274560091574,0.53581031550587}
\definecolor{color2}{rgb}{0.680418912779335,0.615149751467757,0.194054521114453}
\definecolor{color3}{rgb}{0,0.75,0.75}

\begin{groupplot}[group style={group size=1 by 3,vertical sep=0cm},   scale only axis,
    width=1.1\textwidth,
    height=3cm,ticklabel style = {font=\Large}]
\nextgroupplot[
scaled x ticks=manual:{}{\pgfmathparse{#1}},
tick align=outside,
tick pos=left,
x grid style={white!69.0196078431373!black},
xmin=0, xmax=1100,
xtick={1,200,400,600,800,1000},
xtick style={color=black},
xticklabels={},
y grid style={white!69.0196078431373!black},
ymin=0, ymax=4,
ytick style={color=black},
]
\addplot [ultra thick, black]
table {%
1 0
12 0.0403604500391417
23 0.0803911996499683
34 0.120708908578734
45 0.162521346461739
56 0.205764652950029
67 0.247952467810723
78 0.291059721960052
89 0.334823452026845
100 0.377590848514867
111 0.421434031956983
122 0.461930362819919
133 0.501472272342402
144 0.539627156143271
155 0.574128971154633
166 0.610538360664915
177 0.643132559848829
188 0.675144436340333
199 0.710183713916704
210 0.750058433474797
221 0.796151364633395
232 0.846957169219105
243 0.899790021135044
254 0.951409039343545
265 1.00013664887072
276 1.04698336974168
287 1.09029466801967
298 1.12857983769114
309 1.16863210243657
320 1.20847712670364
331 1.25071230476388
342 1.29764037549283
353 1.3464876336167
364 1.39700542682937
375 1.44753982824915
386 1.49657191072168
397 1.54029488192475
408 1.58037507632897
419 1.6187516623402
430 1.65583480611878
441 1.69398111490397
452 1.73557671676532
463 1.77493069781385
474 1.81375267252836
485 1.85115276484746
496 1.88790705953204
507 1.92445449530783
518 1.96359739483689
529 2.00307853500048
540 2.04254618175779
551 2.08106934557669
562 2.11825622507509
573 2.15114310890657
584 2.18226655057923
595 2.21648960684881
606 2.25202795297177
617 2.29048258175354
628 2.33096093265189
639 2.37099730134901
650 2.4089036103822
661 2.44282533840273
672 2.47677383130495
683 2.51338185621256
694 2.55328326710629
705 2.59724755770771
716 2.64317563034182
727 2.68966133682139
738 2.734736941415
749 2.77796410575551
760 2.81992751486008
771 2.85949144432612
782 2.89621763432021
793 2.93008985382316
804 2.96326847158638
815 2.99776385943194
826 3.02954681591809
837 3.062354095416
848 3.09711222352673
859 3.13336469670194
870 3.16961064128645
881 3.20682064703732
892 3.24097118690823
903 3.27408651876913
914 3.30871456136481
925 3.34592481010861
936 3.3822277624126
947 3.41830269603324
958 3.4549968358039
969 3.48902980770923
980 3.52310578057727
991 3.56044714694759
1002 3.59828453593352
1013 3.63962540003167
1024 3.68362622012996
1035 3.72530006431336
1046 3.76856331667097
1057 3.81268339818864
1068 3.85402195310493
1079 3.89427158169131
1090 3.93379924508392
};
\addplot [ultra thick, red]
table {%
800 -0.0277123396832821
800 5.45731780517138
};

\nextgroupplot[
scaled x ticks=manual:{}{\pgfmathparse{#1}},
tick align=outside,
tick pos=left,
x grid style={white!69.0196078431373!black},
xmin=0, xmax=1100,
xtick={1,200,400,600,800,1000},
xtick style={color=black},
xticklabels={},
y grid style={white!69.0196078431373!black},
ymin=0, ymax=4,
ytick={0,1,2,3},
ytick style={color=black},
]
\addplot [ultra thick, orange]
table {%
1 3.29890734404162
12 0.663235422466477
23 2.39432301585484
34 1.1817252840827
45 0.841705742800874
56 0.868501606875384
67 0.832355532653904
78 1.09620393180231
89 0.739625771488922
100 1.66130270461101
111 1.58433567585971
122 0.646218205449906
133 0.908045781691068
144 0.981130238958663
155 1.08419988300123
166 0.976684936431125
177 2.0921335222615
188 0.99518659992987
199 1.06679511556012
210 1.13093746145898
221 0.775389161378529
232 1.10763992043991
243 2.20290164911291
254 0.27092526383784
265 0.949836767928088
276 2.17048343495533
287 1.92076984664135
298 1.1236379013341
309 1.56325498330646
320 1.4357829328598
331 1.80268805867051
342 0.691886253895084
353 0.836543388602308
364 0.620824197819937
375 1.98788911093549
386 1.36226500140025
397 1.13808745983554
408 0.767775985030156
419 0.819239667452389
430 0.882483328511563
441 0.51396754405404
452 0.609196840504072
463 0.748432781786921
474 0.566526510303983
485 1.02182529317334
496 0.811891894040106
507 0.540442483880648
518 0.882027960783441
529 1.20235319097107
540 1.40426985210883
551 0.130698121446475
562 0.648825724715652
573 0.57548844280528
584 0.682749763546645
595 0.795226236618781
606 0.49490843900689
617 1.53198099428163
628 1.77538347219721
639 1.41902768665622
650 1.16192841334653
661 2.07664366619303
672 0.884593531464962
683 0.814823124883695
694 0.447458691583629
705 0.533315317536633
716 0.451727165981983
727 1.8018736069889
738 0.646478918310778
749 0.886110341907583
760 0.968973968280507
771 0.535886829376924
782 0.716981417207558
793 0.360166823885823
804 0.367793101837086
815 0.697008579190744
826 0.5106891885916
837 0.788571194436744
848 0.490595192083688
859 1.23485841104266
870 1.32476576014548
881 0.554005166172154
892 0.558474203986761
903 0.462631931458124
914 0.629396407638544
925 0.55050283616272
936 0.642104762849221
947 0.749039213996797
958 0.669963766503326
969 1.37209499441051
980 1.59980512310846
991 0.386786208158864
1002 1.20150257829187
1013 1.04555291048976
1024 0.374027459148512
1035 1.58635428484474
1046 0.302914397025818
1057 1.18022750454698
1068 1.71602871817991
1079 0.430760662812804
1090 0.428809255793252
};
\addplot [ultra thick, red]
table {%
800 -0.0277123396832821
800 5.45731780517138
};

\nextgroupplot[
tick align=outside,
tick pos=left,
x grid style={white!69.0196078431373!black},
xmin=0, xmax=1100,
xtick={1,200,400,600,800,1000},
xticklabels={\large $1$,$200$,$400$,$600$,$800$,$1000$},
xtick style={color=black},
xlabel={\Huge Time $t$},
y grid style={white!69.0196078431373!black},
ymin=0, ymax=4,
ytick={0,1,2,3},
ytick style={color=black},
]
\addplot [ultra thick, color3]
table {%
1 1.07211942418173
12 0.715947683282418
23 1.17808987411628
34 1.13514230769246
45 1.11255704364284
56 1.24001630812461
67 1.02946367167134
78 1.11319934227878
89 1.02468578763655
100 1.21757497819085
111 1.35952009689909
122 1.07433877307794
133 1.2180073736432
144 1.27324460442352
155 1.12541614487718
166 1.26581683464391
177 1.20972269233243
188 1.08748112223126
199 1.01379356715491
210 1.26838842023468
221 0.80890632149986
232 0.816733117246237
243 1.07143923724885
254 1.00036769239559
265 1.05522351584314
276 0.875731549530379
287 0.973276889975028
298 1.07746747723095
309 0.963662304303101
320 1.15753342743408
331 1.0291658559702
342 1.0731765355519
353 1.2629036467666
364 0.886132870949205
375 0.938477859073403
386 0.961788197723102
397 1.45702393746841
408 1.06856805366874
419 1.06805016367823
430 1.27189282127743
441 1.21331445297525
452 1.28123362176665
463 1.04616130745126
474 0.962320022371588
485 1.25571121755179
496 1.24787481119553
507 0.877983464224105
518 1.27843227201749
529 0.978340097010716
540 0.847723150547945
551 1.01920576133787
562 0.868238349886276
573 1.0383624786342
584 1.06471470158244
595 1.09827543793191
606 0.952192288375604
617 1.46982216818714
628 0.960591442905632
639 1.42838219364312
650 1.33685802595909
661 0.946699544843545
672 1.40823474484416
683 1.03298597229877
694 1.01496948887988
705 1.03460579720796
716 0.868246872844831
727 1.52939409550329
738 1.11284351785515
749 1.15088255423602
760 0.9613370304895
771 1.23541226142783
782 1.01780647898686
793 0.943622753211617
804 2.26710602085175
815 2.46421379723028
826 1.98973053096946
837 1.7603023379994
848 2.44139115588785
859 2.20286292404824
870 1.99649535382686
881 2.33189603636696
892 1.97648033812367
903 2.41319342449945
914 1.93025747413201
925 1.92270744107869
936 2.96821085348566
947 2.47340721319822
958 2.47775319312353
969 2.41582205047812
980 2.24398218486577
991 1.79764083791757
1002 1.68840038204324
1013 2.30459741163814
1024 2.64222663950012
1035 1.88612935835322
1046 1.99557444773247
1057 1.91218368784862
1068 1.94419495296311
1079 1.8243893467867
1090 2.40125159474508
};
\addplot [ultra thick, red]
table {%
800 -0.0277123396832821
800 5.45731780517138
};
\end{groupplot}

\draw ({$(current bounding box.south east)!1.01!(current bounding box.south west)$}|-{$(current bounding box.south west)!0.25!(current bounding box.north west)$}) node[
  scale=1,
  anchor= west,
  text=black,
  rotate=90.0
]{\Huge Detector value};
\end{tikzpicture}

%% file: onlineEMdetector_cpbl.tex
\begin{tikzpicture}

\definecolor{color0}{rgb}{0.12156862745098,0.466666666666667,0.705882352941177}
\definecolor{color1}{rgb}{0.967797559291991,0.441274560091574,0.53581031550587}
\definecolor{color2}{rgb}{0.680418912779335,0.615149751467757,0.194054521114453}
\definecolor{color3}{rgb}{0,0.75,0.75}

\begin{groupplot}[group style={group size=1 by 3,vertical sep=0cm},   scale only axis,
    width=1.1\textwidth,
    height=3cm,legend pos=north west,ticklabel style = {font=\Large}]
\nextgroupplot[
scaled x ticks=manual:{}{\pgfmathparse{#1}},
tick align=outside,
tick pos=left,
x grid style={white!69.0196078431373!black},
xmin=0, xmax=1100,
xtick={1,200,400,600,800,1000},
xtick style={color=black},
xticklabels={},
y grid style={white!69.0196078431373!black},
ymin=0, ymax=5,
ytick style={color=black},
]
\addplot [ultra thick, black]
table {%
1 0
12 0
23 0
34 0.120811291466853
45 0.211165697246017
56 0.101366418134837
67 0
78 0
89 0
100 0
111 0
122 0
133 0
144 0.0379972752512715
155 0.0518425915066714
166 0.0228308985130568
177 0
188 0
199 0
210 0
221 0
232 0
243 0
254 0.0307949645749812
265 0.0727949185094162
276 0.145693991324103
287 0.510865485409263
298 0.776392081695127
309 0.625983628965649
320 0.385607456545328
331 0.0962589580720367
342 0
353 0
364 0.0744177743794222
375 0.178009676565879
386 0.0431756050445864
397 0.0924039798003795
408 0.0523943366881029
419 0.00157995862910638
430 0.0170456443162951
441 0.108811468029992
452 0.471865695112577
463 0.799933706990982
474 0.774794084002454
485 0.989555889722671
496 1.33448211401725
507 1.55642669391451
518 1.627188664496
529 1.64290788730292
540 1.70581521303439
551 1.84652477206441
562 1.87131956150347
573 1.77628454276826
584 1.91342012156282
595 2.03575158795179
606 2.01937093562224
617 2.00169319535698
628 1.87048608809965
639 1.95485654480581
650 2.02921078254483
661 1.83802368139826
672 1.88508022480583
683 2.02206476680117
694 2.05439974902912
705 2.05241162401749
716 1.86950090422759
727 1.51510661991537
738 1.37497336651652
749 1.10953392841065
760 0.806217576832495
771 0.422370592981745
782 0.110996860136265
793 0
804 0
815 0.0107476983283737
826 0.205760411952739
837 0.272149002646889
848 0.316482983750742
859 0.339484818285592
870 0.329709122518424
881 0.432521647905922
892 0.687250762916224
903 1.27799807298062
914 2.06214357573035
925 2.04798765672089
936 2.1721059645432
947 2.39755078355987
958 2.69291666892931
969 2.76861177024472
980 3.09181199232633
991 3.45177495035515
1002 3.71985599417247
1013 3.70216188503903
1024 3.61725878251215
1035 3.80659107702908
1046 3.9813777411353
1057 4.15733121652139
1068 4.32539365021382
1079 4.38256864354818
1090 4.24865716639579
};
\addplot [ultra thick, red]
table {%
800 -0.0277123396832821
800 6.45731780517138
};

\nextgroupplot[
scaled x ticks=manual:{}{\pgfmathparse{#1}},
tick align=outside,
tick pos=left,
x grid style={white!69.0196078431373!black},
xmin=0, xmax=1100,
xtick={1,200,400,600,800,1000},
xtick style={color=black},
xticklabels={},
y grid style={white!69.0196078431373!black},
ymin=0, ymax=5,
ytick={0,2,4},
ytick style={color=black},
]
\addplot [ultra thick, orange]
table {%
1 0.47109777846988
13 0.458018016549132
25 0.259834632795712
37 0.347719117197804
49 0.0899179776339692
61 0.310314472479753
73 0.348820862636567
85 0.274879139118858
97 0.351326683064841
109 0.337155340304915
121 0.435267523860979
133 0.657807230397857
145 0.421594630911036
157 0.118947748004243
169 0.424572339122921
181 0.547356232499946
193 0.311326159233201
205 0.785457289526408
217 0.13353226866606
229 0.30400032036541
241 0.649130064785082
253 0.536778535629195
265 0.371230999985153
277 0.487922675761301
289 0.335251807362884
301 0.596745368794572
313 0.31582972406365
325 0.356118047813284
337 0.287102098459236
349 0.455171253318965
361 0.530952259139688
373 0.733316594866912
385 0.447587885832696
397 0.319160828634764
409 0.0886504261696933
421 0.463390429700803
433 0.364667610595681
445 0.306549216566691
457 0.468664176864024
469 0.261106054449356
481 0.629735058613477
493 0.434906587913639
505 0.339627189852452
517 0.496165544608693
529 0.345666260132932
541 0.29940539723903
553 0.485991199955194
565 0.212138670133495
577 0.376902150182771
589 0.310455881177124
601 0.224486500749738
613 0.25682286808731
625 0.629627949885075
637 0.279461643777258
649 0.47643295750071
661 0.214744213362106
673 0.453708289728506
685 0.561384411422917
697 0.346615572842745
709 0.409950022792209
721 0.295457838191598
733 0.271578368563114
745 0.666744482262548
757 0.453754999638012
769 0.306704528208142
781 0.423503891353312
793 0.304241136766781
805 3.06658823125403
817 1.19935062990892
829 3.20323347144823
841 1.07180910346393
853 0.932498395046588
865 3.21969445720029
877 0.745874796388163
889 0.830169996942362
901 2.32645898587291
913 0.600773211704523
925 4.94166818924033
937 2.62988529105807
949 1.37165732402302
961 2.06782408997737
973 1.48992591509435
985 1.34270939040336
997 0.988965889681009
1009 0.966205048975568
1021 3.46442789069462
1033 1.04528582859543
1045 2.3244998078043
1057 0.801923389620469
1069 2.83803307444012
1081 1.22334911173912
1093 1.28612183994271
1105 1.04018390263626
1117 1.22976878898532
1129 1.50567519070004
1141 1.01587644760615
1153 1.70831660327084
1165 1.19015745010685
1177 2.68626233357738
1189 2.00092071572274
};
\addplot [ultra thick, red]
table {%
800 -0.0277123396832821
800 6.45731780517138
};

\nextgroupplot[
tick align=outside,
tick pos=left,
x grid style={white!69.0196078431373!black},
xmin=0, xmax=1100,
xtick={1,200,400,600,800,1000},
xticklabels={\large $1$,$200$,$400$,$600$,$800$,$1000$},
xtick style={color=black},
xlabel={\Huge Time $t$},
y grid style={white!69.0196078431373!black},
ymin=0, ymax=5,
ytick={0,2,4},
ytick style={color=black},
]
\addplot [ultra thick, color3]
table {%
1 0.544651065937235
13 0.531555836882391
25 0.476680472920917
37 0.55577735364553
49 0.500382788288728
61 0.461667279435683
73 0.632733221142643
85 0.531621167921799
97 0.494599920223151
109 0.53755009988658
121 0.467454447953725
133 0.601478025365718
145 0.591269801649268
157 0.663823096320972
169 0.521917957405479
181 0.577930405021633
193 0.561736201752358
205 0.411828244499488
217 0.394505453744963
229 0.450247662383622
241 0.520701428706772
253 0.445005114286794
265 0.524635243789294
277 0.650040808068332
289 0.606343745894527
301 0.609082000774771
313 0.379491025154989
325 0.397580669827098
337 0.478487899404503
349 0.462233404957643
361 0.517949279162777
373 0.480453958813702
385 0.420913671046746
397 0.55140007062012
409 0.632316541382085
421 0.461650140279777
433 0.598596532694089
445 0.504709476733914
457 0.687215999575411
469 0.404650316410481
481 0.480086235456324
493 0.535836500713077
505 0.563065485253879
517 0.618529827233648
529 0.592004756061919
541 0.607648233054169
553 0.658147567089117
565 0.55488266814344
577 0.580769919361799
589 0.536839243083186
601 0.560613089568122
613 0.453106838713304
625 0.420068568039192
637 0.525719800266664
649 0.522213498251748
661 0.519693070718194
673 0.53053316233125
685 0.622647279645961
697 0.436113520060323
709 0.58431169002471
721 0.474987267212281
733 0.468915356076155
745 0.37440140564503
757 0.662634751737368
769 0.581768933552738
781 0.540830193460916
793 0.601280398878465
805 1.91602340976668
817 2.92253518527529
829 3.31282271543123
841 3.78780816456466
853 2.91114248457947
865 2.79527991854258
877 2.21172681338058
889 3.29490008444582
901 1.21068203107749
913 2.67445935054281
925 3.18264771164211
937 2.10139133899847
949 3.15796583266196
961 3.42335176547177
973 2.4421536828801
985 1.68168317519056
997 2.44154701395637
1009 2.49056477122408
1021 2.96916469534675
1033 4.23405449620274
1045 2.13392874635912
1057 2.46812698270655
1069 4.68410265815653
1081 2.35570643321224
1093 2.11060781161422
1105 1.75811237985326
1117 5.12870376497511
1129 2.52902515190384
1141 2.83373069623737
1153 3.13215814144418
1165 3.72004682410703
1177 2.37579061012762
1189 3.14380631685534
};
\addplot [ultra thick, red]
table {%
800 -0.0277123396832821
800 6.45731780517138
};
\end{groupplot}
\end{tikzpicture}

%% file: stock.tex
\begin{tikzpicture}

\definecolor{color0}{rgb}{1,0.647058823529412,0}

\begin{axis}[
tick align=outside,
tick pos=left,
x grid style={white!69.0196078431373!black},
xmin=-0.38, xmax=4.68,
xtick style={color=black},
xtick={0.15,1.15,2.15,3.15,4.15},
xticklabels={Alg.~\ref{alg:em},SpecTemp,~Kalofolias,GLMM,SC},
extra x ticks={1.18,2.19,3.13},
extra x tick labels={\cite{segarra2017network},\cite{kalofolias2016learn},\cite{maretic2020graph}},
extra x tick style = {grid=none,yshift=-10pt},
y grid style={white!69.0196078431373!black},
ymin=0, ymax=0.75,
ylabel={\Large Average Correlation Score},
ytick style={color=black},
width=9cm,height=7cm,
]
\draw[draw=red,very thick,fill=red!50!white!100] (axis cs:-0.15,0) rectangle (axis cs:0.15,0.63);
\draw[draw=red,very thick,fill=red!50!white!100] (axis cs:0.85,0) rectangle (axis cs:1.15,0.54);
\draw[draw=red,very thick,fill=red!50!white!100] (axis cs:1.85,0) rectangle (axis cs:2.15,0.62);
\draw[draw=blue,very thick,fill=blue!50!white!100] (axis cs:2.85,0) rectangle (axis cs:3.15,0.25);
\draw[draw=orange,very thick,fill=color0!50!white!100] (axis cs:3.85,0) rectangle (axis cs:4.15,0.58);
\draw[draw=red,very thick,fill=red!50!white!100] (axis cs:0.15,0) rectangle (axis cs:0.45,0.69);
\draw[draw=red,very thick,fill=red!50!white!100] (axis cs:1.15,0) rectangle (axis cs:1.45,0.48);
\draw[draw=red,very thick,fill=red!50!white!100] (axis cs:2.15,0) rectangle (axis cs:2.45,0.66);
\draw[draw=blue,very thick,fill=blue!50!white!100] (axis cs:3.15,0) rectangle (axis cs:3.45,0.25);
\draw[draw=orange,very thick,fill=color0!50!white!100] (axis cs:4.15,0) rectangle (axis cs:4.45,0.57);
\draw (axis cs:-0.25,0.64) node[
  scale=0.5,
  anchor=base west,
  text=black,
  rotate=0.0
]{\LARGE \bf 0.63};
\draw (axis cs:0.8,0.55) node[
  scale=0.5,
  anchor=base west,
  text=black,
  rotate=0.0
]{\LARGE \bf 0.54};
\draw (axis cs:1.77,0.63) node[
  scale=0.5,
  anchor=base west,
  text=black,
  rotate=0.0
]{\LARGE \bf 0.62};
\draw (axis cs:2.8,0.26) node[
  scale=0.5,
  anchor=base west,
  text=black,
  rotate=0.0
]{\LARGE \bf 0.25};
\draw (axis cs:3.8,0.59) node[
  scale=0.5,
  anchor=base west,
  text=black,
  rotate=0.0
]{\LARGE \bf 0.58};
\draw (axis cs:0.16,0.7) node[
  scale=0.5,
  anchor=base west,
  text=black,
  rotate=0.0
]{\LARGE \bf 0.69};
\draw (axis cs:1.16,0.49) node[
  scale=0.5,
  anchor=base west,
  text=black,
  rotate=0.0
]{\LARGE \bf 0.48};
\draw (axis cs:2.14,0.67) node[
  scale=0.5,
  anchor=base west,
  text=black,
  rotate=0.0
]{\LARGE \bf 0.66};
\draw (axis cs:3.16,0.26) node[
  scale=0.5,
  anchor=base west,
  text=black,
  rotate=0.0
]{\LARGE \bf 0.25};
\draw (axis cs:4.16,0.58) node[
  scale=0.5,
  anchor=base west,
  text=black,
  rotate=0.0
]{\LARGE \bf 0.57};
\end{axis}

\end{tikzpicture}